\documentclass[10pt,twocolumn,letterpaper]{article}

\usepackage[pagenumbers]{cvpr} 

\makeatletter
\@namedef{ver@everyshi.sty}{}
\makeatother
\usepackage{tikz}
\usepackage{booktabs}
\usepackage{times}
\usepackage{epsfig}
\usepackage{graphicx}
\usepackage{amsmath}
\usepackage{amssymb}
\usepackage{csquotes}
\usepackage{color}
\usepackage{colortbl}
\usepackage{enumitem}
\usepackage{multirow}
\usepackage{dsfont}

\usepackage{adjustbox}
\usepackage[nice]{nicefrac}
\usepackage[weather,clock]{ifsym}
\usepackage[nointegrals]{wasysym} 
\usepackage{pifont}

\usepackage{array} 
\newcolumntype{M}[1]{>{\centering\arraybackslash}m{#1}}

\newcommand{\darkleftbox}[3]{%
\begin{tikzpicture}\node[inner sep=0pt] (p00) {\includegraphics[width=#1]{#2}};\node[rectangle, inner sep=1pt, above right, fill=black, fill opacity=0.6, text opacity=1.] at (p00.south west) {\footnotesize \textcolor{white}{#3}}; \end{tikzpicture}%
}

\newcommand{\darkleftupbox}[3]{%
\begin{tikzpicture}\node[inner sep=0pt] (p00) {\includegraphics[width=#1]{#2}};\node[rectangle, inner sep=1pt, below right, fill=black, fill opacity=0.6, text opacity=1.] at (p00.north west) {\footnotesize \textcolor{white}{#3}}; \end{tikzpicture}%
}

\newcommand{\darkleftrightbox}[4]{%
\begin{tikzpicture}
\node[inner sep=0pt] (p00) {\includegraphics[width=#1]{#2}};
\node[rectangle, inner sep=1pt, above right, fill=black, fill opacity=0.6, text opacity=1.] at (p00.south west) {\footnotesize \textcolor{white}{#3}};
\node[rectangle, inner sep=1pt, above right, fill=black, fill opacity=0.6, text opacity=1., anchor=north east] at (p00.north east) {\footnotesize \textcolor{white}{#4}};
\end{tikzpicture}%
}

\newcommand{\darkleftboxb}[3]{%
\begin{tikzpicture}\node[inner sep=0pt] (p00) {\includegraphics[width=#1]{#2}};\node[rectangle, inner sep=1pt, above right, fill=black, fill opacity=0.6, text opacity=1.] at (p00.south west) {\footnotesize \textcolor{white}{#3}}; \end{tikzpicture}%
}

\newcommand{\darkleftrightboxb}[4]{%
\begin{tikzpicture}
\node[inner sep=0pt] (p00) {\includegraphics[width=#1]{#2}};
\node[rectangle, inner sep=1pt, above right, fill=black, fill opacity=0.6, text opacity=1.] at (p00.south west) {\footnotesize \textcolor{white}{#3}};
\node[rectangle, inner sep=1pt, above right, fill=black, fill opacity=0.6, text opacity=1., anchor=north west] at (p00.north west) {\footnotesize \textcolor{white}{#4}};
\end{tikzpicture}%
}

\newcommand{\lightrightbox}[3]{%
\begin{tikzpicture}\node[inner sep=0pt] (p00) {\includegraphics[width=#1]{#2}};\node[rectangle, inner sep=1pt, above right, fill=white, fill opacity=0.6, text opacity=1., anchor=north east] at (p00.north east) {\footnotesize \textcolor{black}{#3}}; \end{tikzpicture}%
}

\newcommand{\rain}{\scalebox{0.8}{\RainCloud}}
\newcommand{\snow}{\scalebox{0.8}{\Snow}}
\newcommand{\fog}{\scalebox{0.8}{\Fog}}
\newcommand{\weath}[4]{\notblank{#1}{\rain}{\leavevmode\phantom{\rain}}
                       \notblank{#2}{\fog}{\leavevmode\phantom{\fog}}
                       \notblank{#3}{\snow}{\leavevmode\phantom{\snow}}}
\newcommand{\wc}{\Circle}
\newcommand{\bc}{\CIRCLE}
\newcommand{\hcl}{\LEFTcircle}
\newcommand{\threedots}[1]{\ifnumgreater{#1}{10}{\bc}{\ifnumequal{#1}{10}{\bc}{\ifnumgreater{#1}{4}{\hcl}{\wc}}}%
                       \ifnumgreater{#1}{20}{\bc}{\ifnumequal{#1}{20}{\bc}{\ifnumgreater{#1}{14}{\hcl}{\wc}}}%
                       \ifnumgreater{#1}{24}{\bc}{\ifnumgreater{#1}{23}{\hcl}{\wc}}}
\newcommand{\no}{-}
\newcommand{\yes}{\ding{51}}

\newcommand{\targ}{f^T}
\newcommand{\fadv}{\check{f}}
\newcommand{\flow}{f}
\newcommand{\loss}{\mathcal{L}}

\newcommand{\eins}{1}
\newcommand{\zwei}{2}
\newcommand{\de}{\delta_{\pos_{\eins}}}
\newcommand{\dt}{\delta_{\pos_{\zwei}}}
\newcommand{\dmot}{\delta_{\pos_{\eins},\pos_{\zwei}}}

\newcommand{\transp}{\theta}
\newcommand{\dtransp}{\delta_{\transp}}
\newcommand{\col}{\gamma}
\newcommand{\dcol}{\delta_{\col}}
\newcommand{\dhue}{\delta_{\col,\transp}}
\newcommand{\dall}{\delta_{\pos_{\eins},\pos_{\zwei},\col,\transp}}
\newcommand{\pos}{p}
\newcommand{\poszd}{p^{I}}
\newcommand{\mot}{m}
\newcommand{\flake}{B}

\newcommand{\Rd}{\mathds{R}}
\newcommand{\imgdims}{\Rd^{H \times W \times 3}}
\newcommand{\depthmapdims}{\Rd^{H \times W}}
\newcommand{\flakedims}{\Rd^{h \times w}}
\newcommand{\twod}{\Rd^{2}}
\newcommand{\threed}{\Rd^{3}}
\newcommand{\oned}{\Rd{}}

\usepackage[breaklinks=true,bookmarks=false]{hyperref}
\usepackage{orcidlink}

\def\cvprWorkshopName{Preprint. Paper to appear in IEEE/CVF International Conference on Computer Vision (ICCV) 2023.}

\begin{document}

\title{Distracting Downpour:\\Adversarial Weather Attacks for Motion Estimation}

\author{Jenny Schmalfuss \orcidlink{0000-0001-8507-927X} \hspace*{8mm} Lukas Mehl \orcidlink{0009-0001-0548-728X} \hspace*{8mm} Andrés Bruhn \orcidlink{0000-0003-0423-7411}\\
Institute for Visualization and Interactive Systems, University of Stuttgart\\
{\tt\small firstname.lastname@vis.uni-stuttgart.de}
}

\maketitle

\begin{abstract}
Current adversarial attacks on motion estimation, or optical flow, optimize small per-pixel perturbations, which are unlikely to appear in the real world.
In contrast, adverse weather conditions constitute a much more realistic threat scenario.
Hence, in this work, we present a novel attack on motion estimation that exploits adversarially optimized particles to mimic weather effects like snowflakes, rain streaks or fog clouds.
At the core of our attack framework is a differentiable particle rendering system that integrates particles (i) consistently over multiple time steps (ii) into the 3D space (iii) with a photo-realistic appearance.
Through optimization, we obtain adversarial weather that significantly impacts the motion estimation.
Surprisingly, methods that previously showed good robustness towards small per-pixel perturbations are particularly vulnerable to adversarial weather.
At the same time, augmenting the training with non-optimized weather increases a method's robustness towards weather effects and improves generalizability at almost no additional cost.
Our code is available at 
\href{https://github.com/cv-stuttgart/DistractingDownpour}{https://github.com/cv-stuttgart/DistractingDownpour}.\footnotemark[1]
\end{abstract}

\footnotetext[1]{This work is a direct extension of our extended abstract from~\cite{Schmalfuss2022AttackingMotionEstimation}}
\section{Introduction}
\label{sec:intro}

Adversarial attacks that pose a severe threat to neural networks have recently been introduced in the context of optical flow.
There, the goal is to compute the pixel-wise 2D motion $f$ between two consecutive frames $I_{\eins}$ and $I_{\zwei}$ of an image sequence over time.
Current attacks on optical flow~\cite{Agnihotri2023CospgdUnifiedWhite, Koren2021ConsistentSemanticAttacks, Ranjan2019AttackingOpticalFlow, Schmalfuss2022PerturbationConstrainedAdversarial,Schrodi2022TowardsUnderstandingAdversarial} modify these two frames in the 2D space and consequently ignore the actual 3D geometry of the scene as well as the objects moving within.
Moreover, when modifying pixels, they impose bounds on the perturbation's L$_p$ norm rather than imposing visual constraints, which yields attacked images that lack naturalism.
Therefore, robustness analyses with these attacks might not necessarily reflect the robustness of optical flow methods in the real world -- where perturbations are more likely to appear in the form of weather phenomena.

\begin{figure}
\centering
\setlength{\fboxrule}{0.1pt}%
\setlength{\fboxsep}{0pt}%
\begin{tabular}{@{}M{27.2mm}@{\ }M{27.2mm}@{\ }M{27.2mm}@{}}
    \darkleftrightbox{27.2mm}{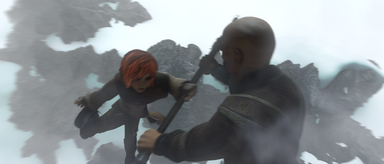}{Original}{Frame $I_{\eins}$} &
    \darkleftrightbox{27.2mm}{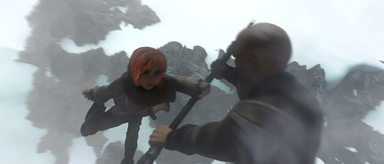}{}{Frame $I_{\zwei}$} &
    \fcolorbox{gray!50}{white}{\lightrightbox{27.2mm}{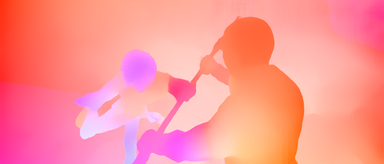}{Optical flow$\vphantom{+}$}}
    \\[-1pt]
    \darkleftbox{27.2mm}{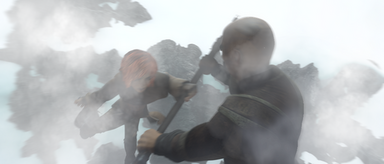}{Adversarial fog} &
    \includegraphics[width=27.2mm]{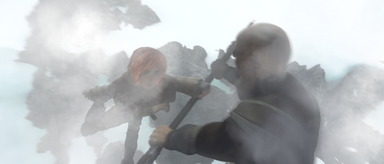} &
    \fcolorbox{gray!50}{white}{\includegraphics[width=27.2mm]{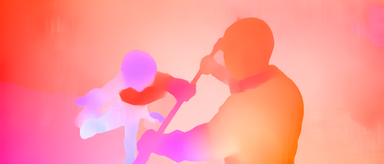}}
    \\[-1pt]
    \darkleftbox{27.2mm}{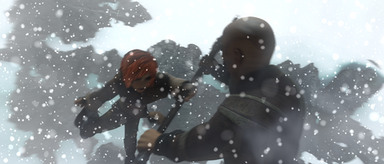}{Adversarial snow} &
    \includegraphics[width=27.2mm]{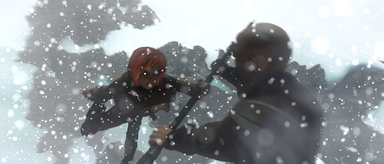} &
    \fcolorbox{gray!50}{white}{\includegraphics[width=27.2mm]{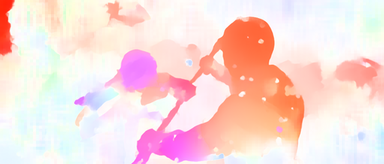}}
    \\[-1pt]
    \darkleftbox{27.2mm}{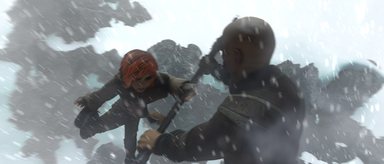}{Adversarial rain} &
    \includegraphics[width=27.2mm]{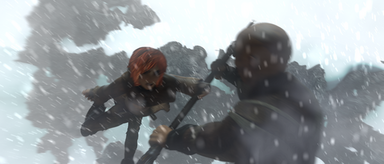} &
    \fcolorbox{gray!50}{white}{\includegraphics[width=27.2mm]{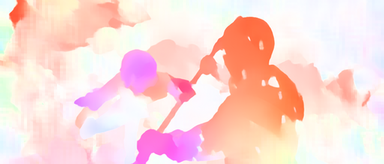}}
    \\[-1pt]
    \darkleftbox{27.2mm}{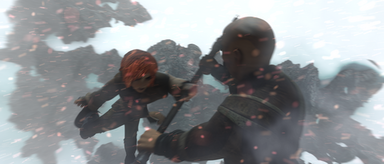}{Adversarial sparks} &
    \includegraphics[width=27.2mm]{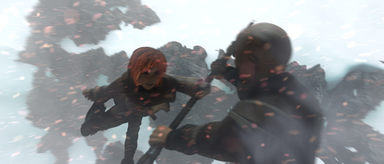} &
    \fcolorbox{gray!50}{white}{\includegraphics[width=27.2mm]{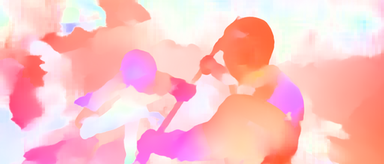}}    
\end{tabular}
\vspace{-0.5mm}
\caption{Weather attacks with \emph{adversarial fog, snow, rain} and \emph{sparks} to perturb optical flow estimation with GMA~\cite{Jiang2021LearningEstimateHidden}. Our weather attacks obey the 3D geometry and camera motion, which is visible in the dynamic motion blur.}
\label{fig:attackweatheroverview}
\vspace{-1mm}
\end{figure}

This work investigates whether naturally occurring weather effects like snow, rain or fog can be manipulated to serve as adversarial samples for motion estimation.
However, simulating weather in this context requires special care:
First, the motion of weather elements should be consistent with the 3D geometry of the scene.
Snowflakes should disappear behind objects and their falling distance should appear larger when closer to the camera.
Second, their motion should be coherent in time. 
A raindrop should fall from top to bottom over the first and second frame, and a fog cloud between two objects should remain there -- even if the camera moved or rotated.

Taking all this into account, we propose an adversarial attack framework that augments images with particle-based weather effects that feature a high degree of realism: 
We create weather particles with a view-consistent 3D motion over time, insert them into the 3D scene in a depth-aware manner, and ensure photo-realism through visual effects.
This enables us to generate adversarially manipulated weather that significantly deteriorates optical flow predictions, while still satisfying the spatiotemporal and visual constraints of naturalistic weather.
Our proposed augmentation and attack procedure can generate a wide range of particle effects, where single particles or super-particles move independently of the remaining scene content.
Fig.~\ref{fig:attackweatheroverview} shows examples of adversarial snowflakes, rain streaks, fire sparks and fog clouds, that differ in size, speed or motion blur, color and transparency.

\medskip \noindent
\textbf{Contributions.} 
Our contributions are three-fold:
\begin{enumerate}[label=(\roman*),topsep=0.25\baselineskip]
\setlength{\itemsep}{.25\baselineskip}%
\setlength{\parskip}{0pt}%
\setlength{\topsep}{0pt}
\setlength{\labelindent}{0pt}%
\item We present a differentiable particle-to-scene rendering framework that generates realistically moving particles in the 3D scene over multiple time steps. It supports a multitude of particle effects ranging from rain and snow over sparks to mist and fog.
\item Based on this differentiable rendering framework, we devise the first adversarial weather attacks for optical flow. They optimize 3D spatial positions and color properties of particles in the scene rather than 2D per-pixel perturbations, resulting in highly realistic images with regard to particle motion and appearance.
\item While being visually indistinguishable from benign weather augmentations, our adversarial weather a\-chieves significant degradations of optical flow predictions.
Interestingly, this particularly holds for methods with high robustness towards small L$_p$ perturbations.
\end{enumerate}

\section{Related work}

Tab.~\ref{tab:literature} provides an overview of weather attacks or spatiotemporal weather augmentations, without direct links to motion estimation and optical flow.
Before we discuss these methods in more detail, we review attacks and robustness towards weather for motion estimation with optical flow.

\medskip \noindent
\textbf{Optical flow attacks and robustness to weather.}
Current optical flow methods based on neural networks are susceptible to adversarially modified input images, which dramatically alter the attacked flow prediction.
Existing adversarial attacks on optical flow methods generate either perturbations with small L$_p$ norms~\cite{Schrodi2022TowardsUnderstandingAdversarial,Schmalfuss2022PerturbationConstrainedAdversarial,Koren2021ConsistentSemanticAttacks,Agnihotri2023CospgdUnifiedWhite} or adversarial patches~\cite{Ranjan2019AttackingOpticalFlow}.
Koren \etal~\cite{Koren2021ConsistentSemanticAttacks} add a constraint to modify semantically coherent pixels only, but none of the attacks introduces geometrical constraints for plausible motion in the 3D space or over time.
Regarding the robustness of optical flow towards weather conditions, few methods explicitly consider rain~\cite{Li2019RainflowOpticalFlow,Li2018RobustOpticalFlow}, snow~\cite{Sakaino2009FallingSnowMotion} or fog~\cite{Yan2020OpticalFlowDense,Yan2022OpticalFlowEstimation}.
However, adversarial attacks have not yet been used to assess the robustness of optical flow methods towards weather effects.

\begin{table}[tb]
\small
\begin{center}
\begin{tabular}{@{\ }l@{\ }c@{\ \ \ }c@{\ \ \ }c@{\ \ \ }c@{\ \ }c@{}}
\toprule
\rotatebox{0}{Method} & \rotatebox{45}{Weather} & \rotatebox{45}{Realism}  & \rotatebox{45}{Attack} & \rotatebox{45}{3D} & \rotatebox{45}{Tempor.} \\
\midrule
\multicolumn{6}{c}{Adversarial weather attacks}\\
\midrule
Sava \etal \cite{Sava2022AssessingImpactTransformations}                                & \weath{r}{}{}\   & \threedots{20} & \yes & \no  & \no  \\
Zhai \etal \cite{Zhai2020AdversarialRainAttack}                                         & \weath{r}{}{}\   & \threedots{25} & \yes & \no  & \no  \\
Marchisio \etal \cite{Marchisio2022FakeweatherAdversarialAttacks}                       & \weath{r}{}{s}\  & \threedots{0}  & \yes & \no  & \no  \\
Gao \etal \cite{Gao2022ScaleFreeTask}                                                   & \weath{r}{}{s}\  & \threedots{10} & \yes & \no  & \no  \\
Wang \etal \cite{Wang2021WhenHumanPose}                                                 & \weath{}{}{s}\   & \threedots{20} & \yes & \no  & \no  \\
Kang \etal \cite{Kang2019TestingRobustnessUnforeseen}                                   & \weath{}{f}{s}\  & \threedots{20} & \yes & \no  & \no  \\
Machiraju \etal \cite{Machiraju_2020_WACV}                                              & \weath{}{f}{}\   & \threedots{20} & \yes & \no  & \no  \\
Gao \etal \cite{Gao2021AdvhazeAdversarialHaze}                                          & \weath{}{f}{}\   & \threedots{25} & \yes & \yes & \no  \\
\midrule
\multicolumn{6}{c}{Realistic weather augmentations}\\
\midrule
Rousseau \etal \cite{Rousseau2006RealisticRealTime}                                     & \weath{r}{}{}\   & \threedots{10} & \no  & \yes & \no  \\
Starik \& Werman \cite{Starik2003SimulationRainVideos}                                  & \weath{r}{}{}\   & \threedots{20} & \no  & \yes & \no  \\
Volk \etal \cite{Volk2019TowardsRobustCnn}                                              & \weath{r}{}{}\   & \threedots{25} & \no  & \yes & \yes \\
Garg \& Nayar \cite{Garg2006PhotorealisticRenderingRain}                                & \weath{r}{}{}\   & \threedots{30} & \no  & \yes & \yes \\
Halder \etal \cite{Halder2019PhysicsBasedRendering}                                     & \weath{r}{f}{}\  & \threedots{20} & \no  & \yes & \yes \\
Tremblay \etal \cite{Tremblay2021RainRenderingEvaluating}                               & \weath{r}{f}{}\  & \threedots{25} & \no  & \yes & \yes \\
von Bernuth \etal \cite{Bernuth2019SimulatingPhotoRealistic}                                  & \weath{}{f}{s}\  & \threedots{25} & \no  & \yes & \yes \\
Wiesemann \& Jiang \cite{Wiesemann2016FogAugmentationRoad}                              & \weath{}{f}{}\   & \threedots{25} & \no  & \yes & \no  \\
\midrule
\midrule
Ours                                                                                    & \weath{r}{f}{s}\ & \threedots{25} & \yes & \yes & \yes \\
\bottomrule
\end{tabular}
\vspace{2.5mm}
\caption{Generating rain~\rain{}, fog~\fog{} and snow~\snow{} in images. The \emph{methods} may support adversarial \emph{attacks}, respect the scene's \emph{3D geometry} or ensure \emph{temporal} consistency over frames.}
\label{tab:literature}
\vspace{-2mm}
\end{center}
\end{table}

\medskip \noindent
\textbf{Adversarial weather attacks.}
In contrast, adversarial attacks that imitate weather effects have been investiga\-ted for classification~\cite{Kang2019TestingRobustnessUnforeseen,Marchisio2022FakeweatherAdversarialAttacks,Gao2022ScaleFreeTask,Gao2021AdvhazeAdversarialHaze,Zhong2022ShadowsCanBe}, object detection~\cite{Sava2022AssessingImpactTransformations,Zhai2020AdversarialRainAttack,Gao2022ScaleFreeTask}, instance segmentation~\cite{Gao2022ScaleFreeTask}, human pose estimation \cite{Wang2021WhenHumanPose} or autonomous steering~\cite{Machiraju_2020_WACV}.
They range from rain~\cite{Sava2022AssessingImpactTransformations,Zhai2020AdversarialRainAttack,Marchisio2022FakeweatherAdversarialAttacks,Gao2022ScaleFreeTask} over snow~\cite{Marchisio2022FakeweatherAdversarialAttacks,Gao2022ScaleFreeTask,Kang2019TestingRobustnessUnforeseen} to fog~\cite{Kang2019TestingRobustnessUnforeseen,Machiraju_2020_WACV,Gao2021AdvhazeAdversarialHaze} and shadows~\cite{Zhong2022ShadowsCanBe}.
As these weather attacks have only been applied to single images rather than sequences, they do not consider temporal consistency.
With exception of \cite{Gao2021AdvhazeAdversarialHaze}, they also neglect the 3D scene geometry.
Both shortcomings prevent their application to realistic motion estimation scenarios.
Moreover, the visual results of weather attacks are often only moderately convincing~\cite{Kang2019TestingRobustnessUnforeseen,Marchisio2022FakeweatherAdversarialAttacks} compared to conventional, non-differentiable rendering of weather effects~\cite{Bernuth2019SimulatingPhotoRealistic,Tremblay2021RainRenderingEvaluating,Garg2006PhotorealisticRenderingRain}.
Wea\-ther effects and attack capabilities are summarized in Tab.~\ref{tab:literature}.

\medskip \noindent
\textbf{Realistic weather augmentations.}
Before applying any vision-based method in the real world, testing its performance under non-perfect weather conditions is crucial.
As a result, there are numerous augmentations to trans\-form clean images into their bad-weather counterparts, \eg via modeled distributions~\cite{Hendrycks2018BenchmarkingNeuralNetwork,Michaelis2019BenchmarkingRobustnessObject}, generative networks~\cite{Ren2020DeepSnowSynthesizing,Wang2021RainGenerationRain,Ye2021ClosingLoopJoint,Wei2021DeraincycleganRainAttentive,Ni2021ControllingRainRemoval,Li2021WeatherGanMulti}, or classical rendering techniques~\cite{Tremblay2021RainRenderingEvaluating,Bernuth2019SimulatingPhotoRealistic}.

However, only a few augmentations respect the 3D geometry of the scene and, ideally, create time-consistent effects for realistic motion of weather across multiple frames and camera perspectives, see Tab.~\ref{tab:literature}.
All such augmentations use classical rendering because generative models~\cite{Ren2020DeepSnowSynthesizing,Wei2021DeraincycleganRainAttentive,Li2021WeatherGanMulti} cannot ensure the spatiotemporal consistency of their generated effects.
Augmentations that respect both, 3D geometry and temporal consistency were proposed for rain~\cite{Volk2019TowardsRobustCnn,Garg2006PhotorealisticRenderingRain}, rain \& fog~\cite{Halder2019PhysicsBasedRendering,Tremblay2021RainRenderingEvaluating}, or fog \& snow~\cite{Bernuth2019SimulatingPhotoRealistic}.
Augmentations that respect only the 3D geometry but not the temporal consistency exist for rain~\cite{Rousseau2006RealisticRealTime,Starik2003SimulationRainVideos} and fog~\cite{Wiesemann2016FogAugmentationRoad}.
To ensure a realistic 3D motion in time, our attack explicitly models the trajectory of weather particles, which is close in spirit to the augmentation of Halder \etal~\cite{Halder2019PhysicsBasedRendering}, and its extension by Tremblay \etal~\cite{Tremblay2021RainRenderingEvaluating}.
However, unlike all discussed rendering approaches, our augmentation is differentiable and thus can readily be used for adversarial attacks.

\section{Adversarial weather for motion estimation}
\label{sec:advsnow}

To study the robustness of optical flow methods towards weather effects, we design an adversarial attack framework that augments image sequences with particle-based weather.
There, we augment an image sequence with parametrized particles to simulate snowflakes, rain streaks or fog clouds of realistic appearance and motion.
Then, we optimize the particle parameters to cause wrong flow predictions with these snowy, rainy or foggy images.

\subsection{Particle-based weather augmentation}

The generation of spatiotemporally consistent and visually appealing weather imposes several constraints on the particles:
Because motion estimation detects moving objects in a 3D scene, a simple 2D animation of the weather particles in the image plane is not realistic enough.
Instead, we model their 3D motion, which also respects object depth and camera motion.
Moreover, expanding our pursuit of realism to the appearance of the weather effects, the particles are integrated with appropriate visual effects.
These include an occlusion-aware depth placement as well as out-of-focus and motion blur.
Finally, the parametrized particles need to be rendered in a differentiable manner to allow their adversarial optimization.

\begin{figure}
\centering
\begin{tikzpicture}
\draw (0, 0) node[inner sep=0,anchor=north west] (img) {
\includegraphics[width=.8\linewidth]{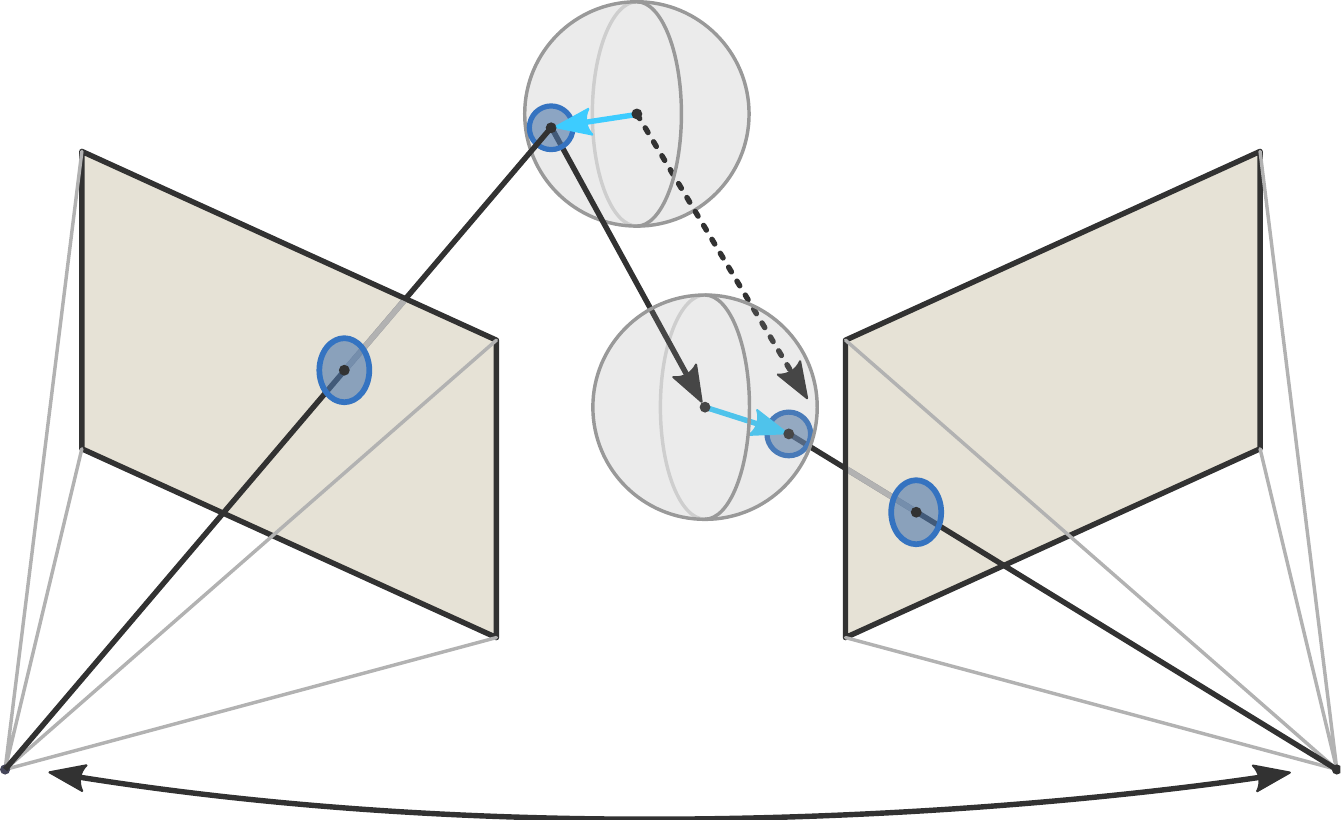}
};
\draw (2.99, -0.35) node {\footnotesize \textcolor{cyan}{$\de$}};
\draw (3.5, -2.2) node {\footnotesize \textcolor{cyan}{$\dt$}};
\draw (2.1, -1) node {\footnotesize $d_{\eins}$};
\draw (5.4, -3.26) node {\footnotesize $d_{\zwei}$};
\draw (2.9, -1.4) node {\footnotesize $\mot$};
\draw (3.5, -0.6) node {\footnotesize $\pos_{\eins}$};
\draw (0.65, -1.1) node {\footnotesize $I_{\eins}$};
\draw (6.05, -1.15) node {\footnotesize $I_{\zwei}$};
\draw (-0.2, -3.9) node {\footnotesize $T_1$};
\draw (6.9, -3.9) node {\footnotesize $T_2$};
\end{tikzpicture}
\vspace{2.0mm}
\caption{Model for particle motion in the 3D space.}
\label{fig:pointsvis}
\end{figure}
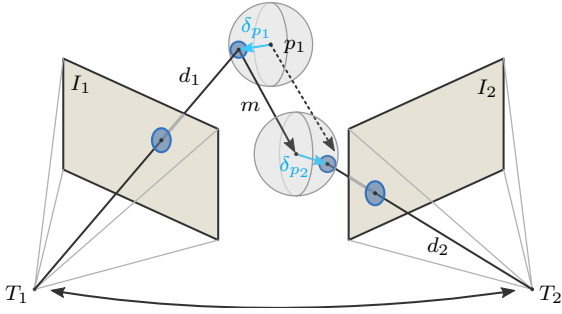
To create weather-augmented 2D images~$I_{\eins}, I_{\zwei}$, we initialize 3D particles and then render them into the images.
During the initialization, we generate a fixed set of particles $\mathcal{P}$ in the 3D scene and equip them with properties: initial 3D positions~$\pos_{\eins}$, 3D motion~$\mot$, 3D offsets~$\de$ before and~$\dt$ after the motion, shapes, scaling, color~$\col$ and transparencies~$\transp$ (see Fig.~\ref{fig:pointsvis} for the motion model).
Here, $\pos_{\eins}$, $\mot$, $\de$, $\dt \in \threed$ are vectors and $\col, \transp \in \oned$ scalars.
For the differentiable rendering of particles in both frames, we make use of the 3D scene information and assume that a depth map of the scene $D\in\depthmapdims$, camera poses $T_1, T_2 \in SE(3)$ and a camera projection matrix $P$ are given.
Below, we describe initialization and rendering in more detail.

\begin{figure}
\centering
\setlength{\fboxrule}{0.1pt}%
\setlength{\fboxsep}{0pt}%
\begin{tabular}{@{}M{41mm}@{\ }M{41mm}@{}}
    \darkleftrightbox{41mm}{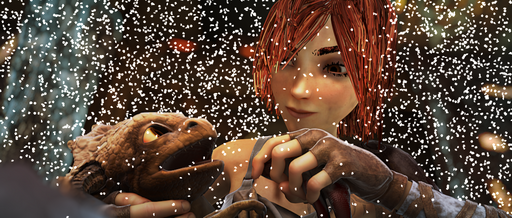}{Flake initialization}{Frame $I_{\eins}$} & \darkleftrightbox{41mm}{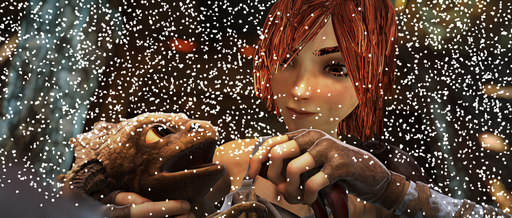}{}{Frame $I_{\zwei}$} \\[-1pt]
    \darkleftbox{41mm}{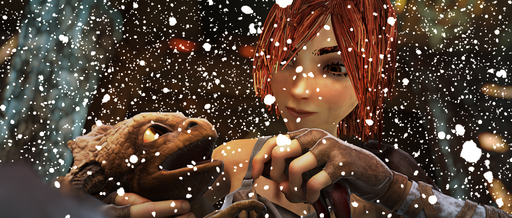}{+ Scaling} & \includegraphics[width=41mm]{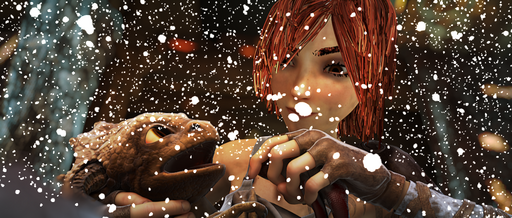}{} \\[-1pt]
    \darkleftbox{41mm}{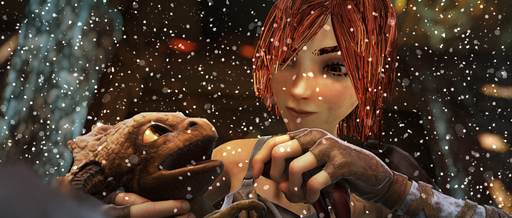}{+ Transparency} & \includegraphics[width=41mm]{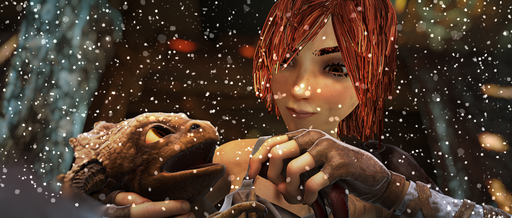}{} \\[-1pt]
    \darkleftbox{41mm}{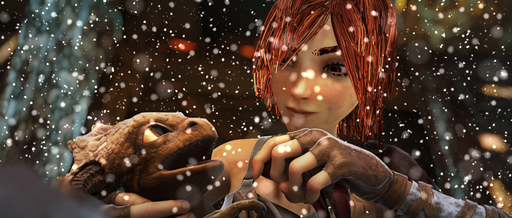}{+ Out-of-focus blur} & \includegraphics[width=41mm]{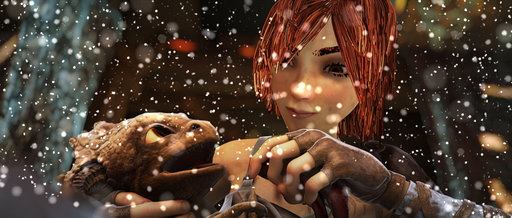}{} \\[-1pt]
    \darkleftbox{41mm}{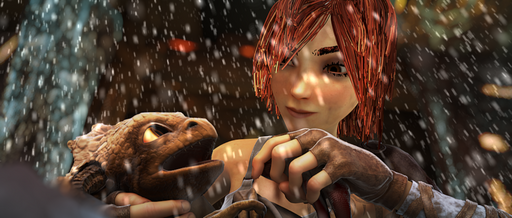}{+ Motion blur \& Occlusions} & \includegraphics[width=41mm]{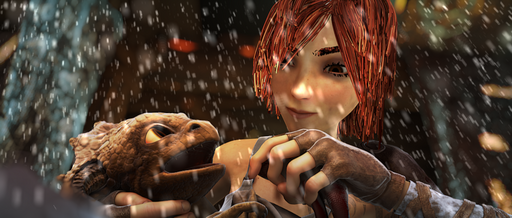}{} \\[-1pt]
    \darkleftbox{41mm}{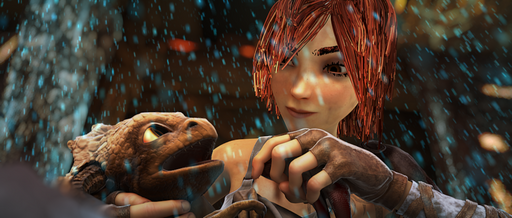}{+ Color additive} & \darkleftbox{41mm}{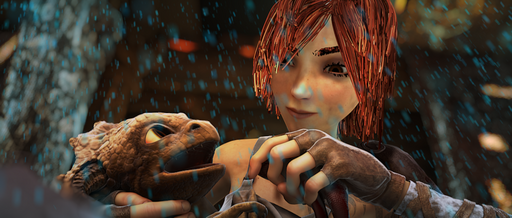}{+ Color alpha blending}
\end{tabular}
\vspace{-0.5mm}
\caption{Breakdown of our realistic snow rendering.}
\label{fig:snowAugmentation}
\vspace{-1.0mm}
\end{figure}

\medskip \noindent
\textbf{Weather particle initialization.}
To initialize the particle positions, we uniformly sample a fixed number of points~$\pos_{\eins}$ from the 3D scene that is visible in the first frame $I_{\eins}$ or -- after adding the 3D motion $\mot$ -- in the second frame $I_{\zwei}$.
Every particle is assigned a 2D gray-scale particle template~$\flake \in \flakedims$ (billboard), randomly sampled from a template library and rotated by a random angle (Fig.~\ref{fig:snowAugmentation}, row~1).
Then, each particle template is scaled by its particle's inverse depth (row~2), and the particle transparency $\transp$ is set to a depth-dependent value (rows~3).
Finally, we generate realistic out-of-focus blur by convolving the particle template with a disk-shaped point spread function (row~4).

\medskip \noindent
\textbf{Weather particle rendering.}
We render the particles with their associated motion blur, 3D positions, colors and transparencies in the given input frames in four steps, detailed below:
We initially add motion blur particles, then project all particle templates onto the image plane, subsequently handle occlusions and finally update the pixels with the colored particle template.

First, if motion blur is added, each initial particle is replaced by $K$ particles.
These are evenly spaced along the 3D motion vector and their transparency is reduced to $\frac{1}{K}$; Otherwise, the rendering proceeds as described below.
In contrast to simple 2D approximations, this true motion blur respects 3D motion and camera motion (Fig.~\ref{fig:snowAugmentation}, row 5).

Second, for each particle, we compute the 2D points in both frames from its position in 3D.
This yields the center positions $\poszd_{\eins}$, $\poszd_{\zwei} \in \twod$ of the 2D particle templates in the 2D images $I_{\eins}$, $I_{\zwei} \in \imgdims$.
Using the camera projec\-tion matrix $P$ and the relative transformation matrix $T_\text{rel} = T_2 T_1^{-1}$, we project the 3D points and their motion-displaced positions into the first and second frame, respectively:
\begin{align}
\poszd_{\eins} &= P \big(\pos_{\eins}\!+\!\de\big),\\
\poszd_{\zwei} &= P \big(T_\text{rel} (\pos_{\eins}\!+\!\de) + (\mot\!+\!\dt)\big).
\end{align}%
Because this maps the template to subpixel locations, interpolating the 2D particle templates at the true pixel locations becomes necessary.
Using bilinear interpolation enables differentiation \wrt the 3D particle positions.

Third, we handle occlusions by multiplying the particle template with a visibility map.
The visibility map $V \in \flakedims$ uses a scene depth map $D \in \flakedims$, cropped to the location of $\flake$, and the particle depth $d \in \oned$ per camera:
\begin{equation}
V_t = (1+e^{\beta  (d_t-D_t)})^{-1} \quad \text{for } t=\eins,\zwei.
\end{equation}
This sigmoid function is $\approx$1 (full visibility) for particles whose depth is smaller than the scene depth, and $\approx$0 (full occlusion) otherwise.
When the depths are similar, it creates a smooth transition to allow differentiation.
We use sharper transitions $\beta=250$ for pure rendering and smoother ones $\beta=30$ for differentiation.
Overall, this yields a realistic, occlusion-aware scene integration (Fig.~\ref{fig:snowAugmentation}, row 5).

Fourth and last, the particle color templates can be applied to the previously computed pixel positions.
Our rendering framework supports two color modes for this: additive color blending and alpha blending.
Additive blending creates a brightening effect (Fig.~\ref{fig:snowAugmentation}, last row left), similar to colored light sources, by updating the pixel color as
\begin{equation}
I_c = I_c + \sum_{j\in\mathcal{P}}  \col^j_c \transp^j \flake^j \quad \text{for }c=\text{R,G,B}.
\end{equation}
For each particle, $\col_c$ is the color per channel, $\transp$ the transparency scaling and $\flake$ the template, which itself is a transparency map.
In contrast, alpha blending creates more \enquote{solid} particles (Fig.~\ref{fig:snowAugmentation}, last row right) by weighting background and particle color according to particle transparency.
We use Meshkin's method~\cite{Meshkin2007SortIndependentAlpha} for an order-independent alpha blending that can process all particles in parallel:
\begin{equation}
I_c = I_c \Big(1-\sum_{j\in\mathcal{P}} \transp^j \flake^j \Big) + \sum_{j\in\mathcal{P}} \col^j_c \transp^j \flake^j \ \  \text{for }c=\text{R,G,B}.
\end{equation}

\subsection{Adversarial weather optimization}

After the particles $\mathcal{P}$ are initialized and rendered, we adversarially optimize certain weather parameters to change the output $\fadv$ of optical flow networks towards a desired target flow $\targ$.
In this context, we consider the particle motion offsets $\de$ before and $\dt$ after the motion as well as transparency $\dtransp$ and color $\dcol$ offsets.
Other parameters like initial 3D positions, 3D motion and 2D template are fixed.
To ensure a valid range of color $\col$ and transparency $\transp$ values after the optimization, we transform these bounded variables to unbounded ones $\eta_\col$, $\eta_\transp$ via an atanh-transformation~\cite{Carlini2017TowardsEvaluatingRobustness}
\begin{equation}
\eta_\xi = \operatorname{atanh}( 2 \xi - 1 ), \quad \xi = \transp, \col
\end{equation}
and optimize $\eta_\col+\dcol$ and $\eta_\transp+\dtransp$ in this domain.
Then, our loss function measures the difference between initial and attacked flow via the average endpoint error (AEE)~\cite{Schmalfuss2022PerturbationConstrainedAdversarial}:
\begin{equation}
\loss(\fadv, \targ, \mathcal{P}) = \text{AEE}(\fadv, \targ) +\! \sum_{t  \in \eins,\zwei} \frac{\alpha_t}{|\mathcal{P}|} \sum_{j\in\mathcal{P}} \frac{\|\delta_{\pos_t}^j\|_2^2}{d_t^j}.
\end{equation}
Additionally, this loss restricts the magnitude of the motion offset via an $\alpha$-balanced MSE-like term, where $|\mathcal{P}|$ is the number of particles.
It allows larger offsets $\de$, $\dt$ for distant snowflakes, as the same 3D motion in the background yields smaller 2D offsets than in the foreground.
Hence, we encourage similar motion offsets in the rendered 2D images by scaling the offsets with the inverse particle depth $d$.

\section{Experiments}
\label{sec:experiments}
In several experiments,
(i) we demonstrate our augmentation framework and identify weather that strongly impacts optical flow methods, 
(ii) we attack optical flow methods with adversarially optimized particles to evaluate their sensitivity and
(iii) we augment training data with snow to improve quality and robustness towards weather.
A full list of parameters for the experiments is given in the supplement.
Our PyTorch framework is available at \href{https://github.com/cv-stuttgart/DistractingDownpour}{https://github.com/cv-stuttgart/DistractingDownpour}.

In the experiments, we augment frames from Sintel~\cite{Butler2012NaturalisticOpenSource}, a standard dataset for optical flow that provides depth and camera information.
We calculate the adversarial robustness AEE$(\flow,\fadv)$ from~\cite{Schmalfuss2022PerturbationConstrainedAdversarial}, which measures how the benign optical flow $\flow$ differs from $\fadv$ on weather-augmented images.
For robust methods, the output should only change proportional to input.
This is formalized by the Lipschitz constant (the concept underlying adversarial robustness), which allows robustness comparisons for input changes of similar magnitude, independent of the ground truth optical flow.
Following~\cite{Schmalfuss2022PerturbationConstrainedAdversarial}, we select RAFT~\cite{Teed2020RaftRecurrentAll} \& GMA~\cite{Jiang2021LearningEstimateHidden}, FlowNet2 (FN2)~\cite{Ilg2017Flownet2Evolution} and SpyNet~\cite{Ranjan2017OpticalFlowEstimation} as approaches with either high quality \& low robustness, medium quality and robustness or low quality \& high robustness, respectively.
Additionally, we consider FlowFormer (FF)~\cite{Huang2022FlowformerTransformerArchitecture} for its transformer architecture and top results, and FlowNetCRobust (FNCR)~\cite{Schrodi2022TowardsUnderstandingAdversarial} for its robustness-enhancing design.

\subsection{Weather augmentations}

\begin{table}
\small
\begin{center}
\begin{tabular}{@{\ }l@{\quad}l@{\ }rrrrrr@{\ }}
\toprule
\multicolumn{2}{l}{\rotatebox{0}{Weather}} & \multicolumn{1}{c}{\rotatebox{45}{FN2}} & \multicolumn{1}{c}{\rotatebox{45}{FNCR}} & \multicolumn{1}{c}{\rotatebox{45}{SpyNet}} & \multicolumn{1}{c}{\rotatebox{45}{RAFT}} & \multicolumn{1}{c}{\rotatebox{45}{GMA}} & \multicolumn{1}{c}{\rotatebox{45}{FF}} \\
\midrule
\multirow{5}{*}{\rotatebox{90}{Particles}} &
1000       &          3.94  &          5.28  &         3.55  &         1.39  &         1.16  &         0.83  \\
&
2000       &          7.58  &          7.94  &         5.33  &         2.97  &         2.51  &         1.86  \\
& 
\cellcolor{gray!20}3000       & \cellcolor{gray!20}11.95  & \cellcolor{gray!20}10.29  & \cellcolor{gray!20}6.75  & \cellcolor{gray!20}5.03  & \cellcolor{gray!20}4.14  & \cellcolor{gray!20}3.27  \\
&
4000       &         17.01  &         12.35  &         7.75  &         7.40  &         5.91  &         4.42  \\
&
5000       & \textbf{23.42} & \textbf{14.62} & \textbf{8.67} & \textbf{9.81} & \textbf{7.91} & \textbf{5.53} \\
\midrule
\multirow{5}{*}{\rotatebox{90}{Motion blur}} &
0.0\phantom{000}     &         11.95  &         10.29  & \textbf{6.75} & \textbf{5.03} & \textbf{4.14} & \textbf{3.27} \\
&
0.0375               & \textbf{15.60} &         12.95  &         6.44  &         4.04  &         3.22  &         3.17  \\
&
0.075\phantom{0}     &         15.01  & \textbf{13.35} &         5.78  &         3.90  &         3.04  &         3.22  \\
&
0.1125               &         13.27  &         12.97  &         5.30  &         3.78  &         2.76  &         2.73  \\
&
\cellcolor{gray!20}0.15\phantom{00}     &\cellcolor{gray!20}10.86  &\cellcolor{gray!20}11.52  &\cellcolor{gray!20}4.64  &\cellcolor{gray!20}3.50  &\cellcolor{gray!20}2.49  &\cellcolor{gray!20}2.05  \\
\midrule
\multirow{5}{*}{\rotatebox{90}{Color additive}}
&
white      & \textbf{14.05} & \textbf{14.68} & \textbf{6.47} & \textbf{5.49} & \textbf{4.63} & \textbf{4.68} \\
&
\cellcolor{gray!20}red        &\cellcolor{gray!20}12.57  &\cellcolor{gray!20}12.07  &\cellcolor{gray!20}4.21  &\cellcolor{gray!20}3.74  &\cellcolor{gray!20}2.95  &\cellcolor{gray!20}3.03  \\
&
green      &          9.02  &          9.64  &         3.68  &         3.16  &         2.76  &         2.56  \\
&
blue       &          7.84  &         10.47  &         3.37  &         3.52  &         3.31  &         2.30  \\
&
color      &         11.17  &         11.50  &         4.36  &         4.11  &         3.75  &         3.18  \\
\midrule
\multirow{4}{*}{\rotatebox{90}{Size}} &
small      & \textbf{5.48} & \textbf{5.52} &         4.41  & \textbf{4.58} & \textbf{4.41} & \textbf{4.47} \\
&
medium     &         4.45  &         4.47  &         5.63  &         3.04  &         3.03  &         2.50  \\
&
large      &         2.23  &         2.91  &         3.51  &         1.17  &         1.16  &         0.92  \\
&
\cellcolor{gray!20}fog & \cellcolor{gray!20}4.72  & \cellcolor{gray!20}5.25  & \cellcolor{gray!20}\textbf{5.87} & \cellcolor{gray!20}3.59  & \cellcolor{gray!20}3.66  & \cellcolor{gray!20}3.24  \\
\bottomrule
\end{tabular}
\vspace{2mm}
\caption{Robustness AEE$(\flow,\fadv)$ $\downarrow$~\cite{Schmalfuss2022PerturbationConstrainedAdversarial} of particle-based weather augmentations for \emph{optical flow methods} on Sintel train, worst robustness is bold. The main augmentations snow, rain, sparks and fog are highlighted in grey and visualized in Fig.~\ref{fig:augmentations}.}
\label{table:augmentations}
\vspace{-1.5mm}
\end{center}
\end{table}

\begin{figure}[tb]
\centering
\setlength{\fboxrule}{0.1pt}%
\setlength{\fboxsep}{0pt}%
\begin{tabular}{@{}M{40mm}@{\ \ }M{40mm}@{}M{40mm}@{\ }M{40mm}@{}}
    \darkleftupbox{40mm}{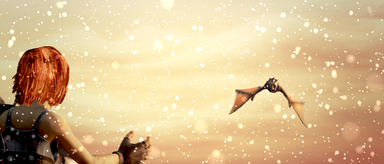}{Snow} &
    \darkleftupbox{40mm}{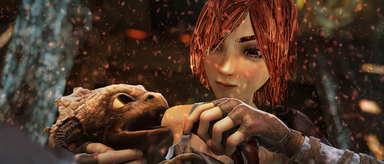}{Sparks}
    \\[-3pt]
    \includegraphics[width=40mm]{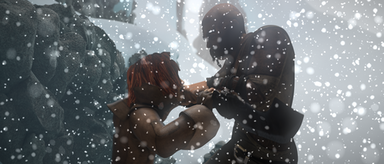} &
    \includegraphics[width=40mm]{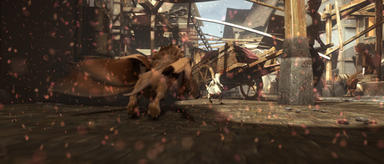}
    \\[2pt]
    \darkleftupbox{40mm}{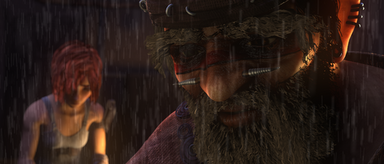}{Rain} &
    \darkleftupbox{40mm}{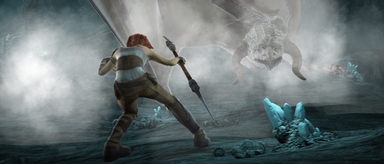}{Fog}
    \\[-3pt]
    \includegraphics[width=40mm]{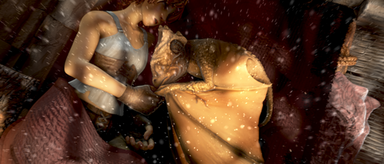} &
    \includegraphics[width=40mm]{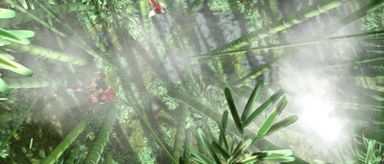}    
\end{tabular}
\vspace{1mm}
\caption{Visual examples for \emph{snow, rain, sparks} and \emph{fog} augmentations on a single frame for highlighted effects from Tab.~\ref{table:augmentations}. Note the realistic motion blur (birds-eye view) in column 1 row 4.}
\label{fig:augmentations}
\vspace{-1mm}
\end{figure}

To observe how \emph{random augmentations} change the predictions of optical flow methods, we first investigate the impact of various particle effects on Sintel~\cite{Butler2012NaturalisticOpenSource} data and select default configurations for snow, rain, sparks and fog.
Then, we illustrate our flexible rendering on further datasets.

\medskip \noindent
\textbf{Particle parameters for weather creation.}
Here, we create diverse weather effects through complex hyper-parameter combinations in our rendering framework to test their effect on flow predictions.
The hyperparameters listed in Tab.~\ref{table:augmentations} are the most prominent ones that were altered, \ie the number of particles, motion blur length, color and size (full parameter list in supplement).
Our baseline weather (\emph{particles: 3000}) uses 3000 small additive white particles without motion blur, \ie 0.0 fraction of motion magnitude.

Tab.~\ref{table:augmentations} summarizes the robustness of optical flow methods on particle-augmented Sintel training data \emph{without} adversarial optimization.
Visualizations are in the supplement.
All methods are most sensitive to the number of particles and change their prediction strongest when many particles are present.
The sensitivity also increases for non-transparent effects, \eg for \emph{motion blur: 0.0} or \emph{particle size: small}.
Also, large color offsets on multiple channels are strongly perturbing, \ie most for white or random colors, and additive blending perturbs more than alpha blending, see supplement.
To summarize, optical flow methods change their predictions significantly in the presence of many small, bright particles, which do not exist in the standard training datasets~\cite{Menze2015ObjectSceneFlow,Butler2012NaturalisticOpenSource,Dosovitskiy2015FlownetLearningOptical,Mayer2016LargeDatasetTrain}.
However, we find that accurate methods like FlowFormer, RAFT or GMA are more robust, already hinting at an improved particle recognition that is discussed in the next subsection.

For further analyses, we select defaults for snow (\emph{particles: 3000}), rain (\emph{motion blur: 0.15}), sparks (\emph{color: red}) and fog (\emph{size: fog}), highlighted gray in Tab.~\ref{table:augmentations} and illustrated in Fig.~\ref{fig:augmentations}.
Because the most effective configurations, \eg \emph{color: white} or \emph{size: small} all basically represent snow, we opted for weather configurations with greater visual diversity.
For snow, \emph{particles: 3000} is computationally more efficient than the most effective configuration \emph{particles: 5000}.

\medskip \noindent
\textbf{Augmenting different datasets.}
\begin{figure}[tb]
\centering
\setlength{\fboxrule}{0.1pt}%
\setlength{\fboxsep}{0pt}%
\begin{tabular}{@{}M{40mm}@{\ \ }M{40mm}@{}}
    \darkleftupbox{40mm}{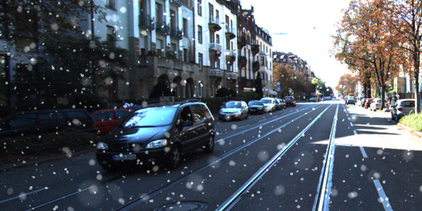}{KITTI} &
    \darkleftupbox{40mm}{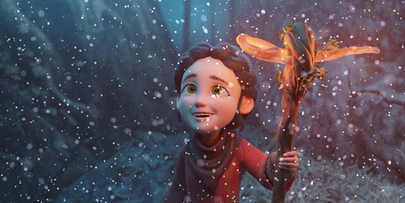}{Spring}
    \\[-2pt]
    \includegraphics[width=40mm]{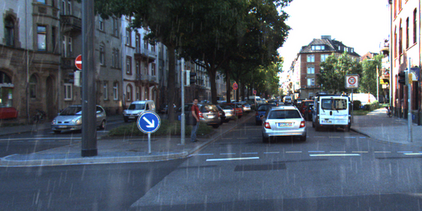} &
    \includegraphics[width=40mm]{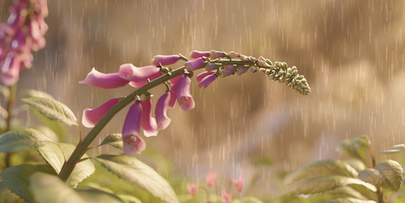}
    \\
\end{tabular}
\vspace{1mm}
\caption{Example augmentations for \emph{KITTI}~\cite{Menze2015ObjectSceneFlow} and \emph{Spring}~\cite{Mehl2023SpringHighResolution} datasets, with snow (top) and rain (bottom).}
\label{fig:KittiSpring}
\vspace{-1mm}
\end{figure}
Even though we focus on Sintel, our rendering approach also permits the augmentation of other datasets.
Augmented samples from KITTI~\cite{Menze2015ObjectSceneFlow} and Spring~\cite{Mehl2023SpringHighResolution} are shown in Fig.~\ref{fig:KittiSpring}.
For KITTI, we use interpolated depth maps and estimate camera poses from the 3D motion in rigid parts of the scene~\cite{Arun1987LeastSquaresFitting}.
A full robustness evaluation on augmented KITTI data is in the supplement.

\subsection{Adversarial weather attacks}

With our framework to generate natural weather effects, we now evaluate the \emph{attack capabilities} of this differentiable weather.
First, we investigate the sensitivity of optical flow methods towards optimizing different particle parameters. 
Second, we attack them with snow, rain, sparks and fog from the previous section.
Third and last, we compare the effectiveness of a non-L$_p$ snow attack to previous L$_p$ attacks on optical flow.
All attacks use $\alpha_{\eins}\!=\!\alpha_{\zwei}\!=\!1000$ in the loss, Adam with learning rate 1e-5 and, following~\cite{Schmalfuss2022PerturbationConstrainedAdversarial}, a zero-flow target $\targ=0$ which yields a white flow visualization.

\medskip \noindent
\textbf{Investigation of weather attack parameters.}
\begin{table}[tb]
\small
\begin{center}
\begin{tabular}{@{\ }l@{\ \ \ }r@{\ \ \ }r@{\ \ }r@{\ \ \ }r@{\ \ \ }r@{\ \ \ }r@{\ }}
\toprule
Parameters & \multicolumn{1}{c}{\rotatebox{45}{FN2}} & \multicolumn{1}{c}{\rotatebox{45}{FNCR}} & \multicolumn{1}{c}{\rotatebox{45}{SpyNet}} & \multicolumn{1}{c}{\rotatebox{45}{RAFT}} & \multicolumn{1}{c}{\rotatebox{45}{GMA}} & \multicolumn{1}{c}{\rotatebox{45}{FF}} \\
\midrule
Initial                     &            10.23  &            10.68  &            4.42  &            3.80  &            3.77  &            2.56  \\
\midrule
$\de$                       &            13.54  &            15.65  &            7.08  &            7.39  &            8.64  &            5.33  \\
$\dt$                       &            11.99  &            14.21  &            5.64  &            5.83  &            6.69  &            4.04  \\
$\dcol$                     &            12.86  &            15.95  &            7.52  &            6.00  &            7.58  &            4.74  \\
$\dtransp$                  &            11.70  &            14.45  &            6.75  &            5.29  &            6.24  &            3.56  \\
\midrule
$\dmot$                     & \underline{14.08} &            15.87  &            7.71  & \underline{8.27} & \underline{9.42} &            5.49  \\
$\dhue$                     &            14.06  &    \textbf{16.71} &    \textbf{8.94} &            7.39  &            8.99  &    \textbf{5.84} \\
\midrule
$\dall$                     &    \textbf{14.23} & \underline{16.01} & \underline{7.78} &    \textbf{8.32} &    \textbf{9.50} & \underline{5.71} \\
\bottomrule
\end{tabular}
\vspace{2mm}
\caption{Adversarial robustness AEE$(\flow,\fadv)$ $\downarrow$~\cite{Schmalfuss2022PerturbationConstrainedAdversarial} of adversarial particles, optimized for combinations of \emph{particle parameters} $\de$, $\dt$, $\dcol$ and $\dtransp$ on Sintel-tr115. \emph{Initial} measures the robustness of randomly initialized particles. The most vulnerable setup is bold.}
\label{table:optimvariables}
\vspace{-1.5mm}
\end{center}
\end{table}
\begin{figure*}
\centering
\setlength{\fboxrule}{0.1pt}%
\setlength{\fboxsep}{0pt}%
\begin{tabular}{@{}M{19.1mm}@{}M{19.1mm}@{}M{19.1mm}@{\ }M{19.1mm}@{}M{19.1mm}@{}M{19.1mm}@{\ }M{19.1mm}@{}M{19.1mm}@{}M{19.1mm}@{}}
    \multicolumn{3}{c}{FlowNet2} & \multicolumn{3}{c}{FlowNetCRobust} & \multicolumn{3}{c}{SpyNet}
    \\
    \darkleftbox{19.1mm}{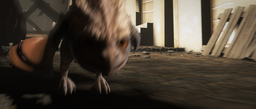}{\scriptsize Orig.} &
    \includegraphics[width=19.1mm]{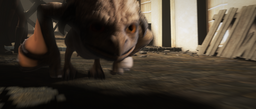} &
    \fcolorbox{gray!50}{white}{\includegraphics[width=19.1mm]{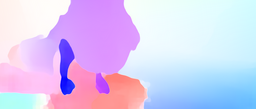}} &
    \darkleftbox{19.1mm}{attacks_smaller/market_5/frame_0001.png}{\scriptsize Orig.} &
    \includegraphics[width=19.1mm]{attacks_smaller/market_5/frame_0002.png} &
    \fcolorbox{gray!50}{white}{\includegraphics[width=19.1mm]{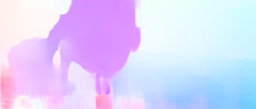}} &
    \darkleftbox{19.1mm}{attacks_smaller/market_5/frame_0001.png}{\scriptsize Orig.} &
    \includegraphics[width=19.1mm]{attacks_smaller/market_5/frame_0002.png} &
    \fcolorbox{gray!50}{white}{\includegraphics[width=19.1mm]{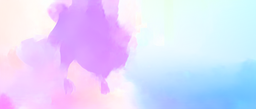}}
    \\[-3.5pt]
    \darkleftbox{19.1mm}{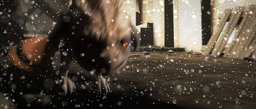}{\scriptsize Snow rand.} &
    \includegraphics[width=19.1mm]{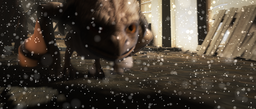} &
    \fcolorbox{gray!50}{white}{\includegraphics[width=19.1mm]{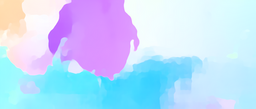}} &
    \darkleftbox{19.1mm}{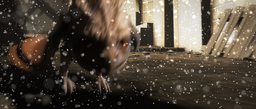}{\scriptsize Snow rand.} &
    \includegraphics[width=19.1mm]{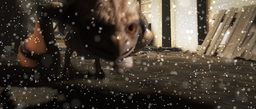} &
    \fcolorbox{gray!50}{white}{\includegraphics[width=19.1mm]{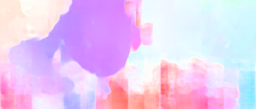}} &
    \darkleftbox{19.1mm}{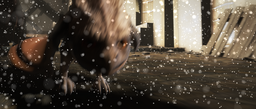}{\scriptsize Snow rand.} &
    \includegraphics[width=19.1mm]{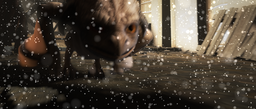} &
    \fcolorbox{gray!50}{white}{\includegraphics[width=19.1mm]{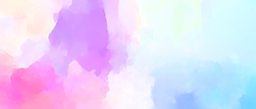}}
    \\[-3.5pt]
    \darkleftbox{19.1mm}{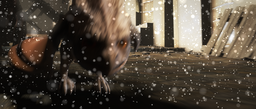}{\scriptsize Snow adv.} &
    \includegraphics[width=19.1mm]{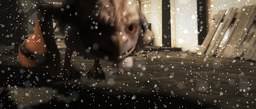} &
    \fcolorbox{gray!50}{white}{\includegraphics[width=19.1mm]{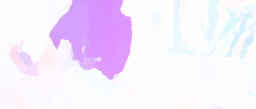}} &
    \darkleftbox{19.1mm}{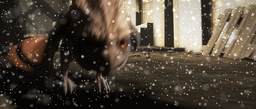}{\scriptsize Snow adv.} &
    \includegraphics[width=19.1mm]{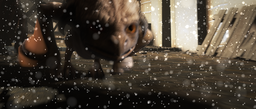} &
    \fcolorbox{gray!50}{white}{\includegraphics[width=19.1mm]{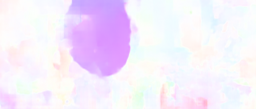}} &
    \darkleftbox{19.1mm}{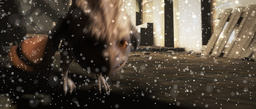}{\scriptsize Snow adv.} &
    \includegraphics[width=19.1mm]{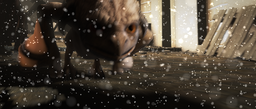} &
    \fcolorbox{gray!50}{white}{\includegraphics[width=19.1mm]{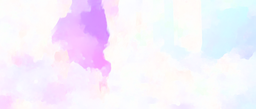}}
    \\
    \darkleftbox{19.1mm}{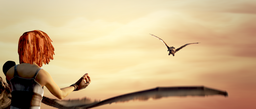}{\scriptsize Orig.} &
    \includegraphics[width=19.1mm]{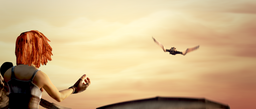} &
    \fcolorbox{gray!50}{white}{\includegraphics[width=19.1mm]{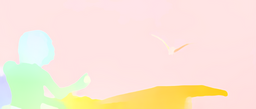}} &
    \darkleftbox{19.1mm}{attacks_smaller/temple_3/frame_0003.png}{\scriptsize Orig.} &
    \includegraphics[width=19.1mm]{attacks_smaller/temple_3/frame_0004.png} &
    \fcolorbox{gray!50}{white}{\includegraphics[width=19.1mm]{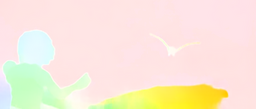}} &
    \darkleftbox{19.1mm}{attacks_smaller/temple_3/frame_0003.png}{\scriptsize Orig.} &
    \includegraphics[width=19.1mm]{attacks_smaller/temple_3/frame_0004.png} &
    \fcolorbox{gray!50}{white}{\includegraphics[width=19.1mm]{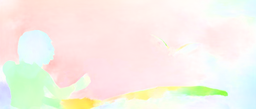}}
    \\[-3.5pt]
    \darkleftbox{19.1mm}{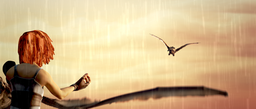}{\scriptsize Rain rand.} &
    \includegraphics[width=19.1mm]{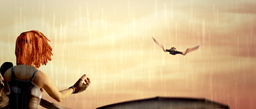} &
    \fcolorbox{gray!50}{white}{\includegraphics[width=19.1mm]{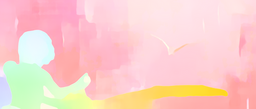}} &
    \darkleftbox{19.1mm}{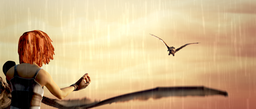}{\scriptsize Rain rand.} &
    \includegraphics[width=19.1mm]{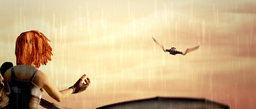} &
    \fcolorbox{gray!50}{white}{\includegraphics[width=19.1mm]{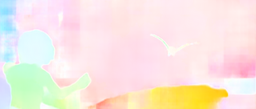}} &
    \darkleftbox{19.1mm}{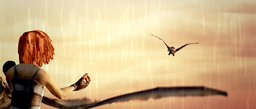}{\scriptsize Rain rand.} &
    \includegraphics[width=19.1mm]{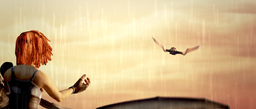} &
    \fcolorbox{gray!50}{white}{\includegraphics[width=19.1mm]{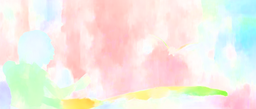}}
    \\[-3.5pt]
    \darkleftbox{19.1mm}{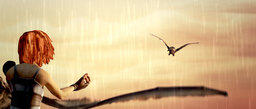}{\scriptsize Rain adv.} &
    \includegraphics[width=19.1mm]{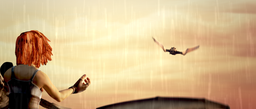} &
    \fcolorbox{gray!50}{white}{\includegraphics[width=19.1mm]{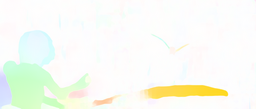}} &
    \darkleftbox{19.1mm}{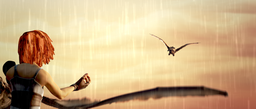}{\scriptsize Rain adv.} &
    \includegraphics[width=19.1mm]{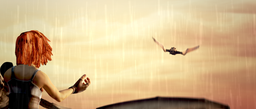} &
    \fcolorbox{gray!50}{white}{\includegraphics[width=19.1mm]{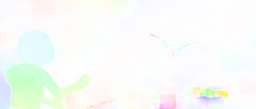}} &
    \darkleftbox{19.1mm}{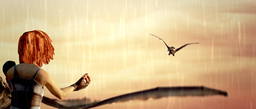}{\scriptsize Rain adv.} &
    \includegraphics[width=19.1mm]{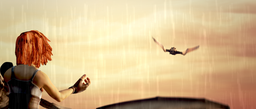} &
    \fcolorbox{gray!50}{white}{\includegraphics[width=19.1mm]{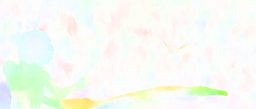}}
    \\
    \darkleftbox{19.1mm}{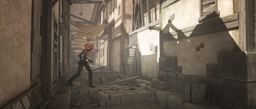}{\scriptsize Orig.} &
    \includegraphics[width=19.1mm]{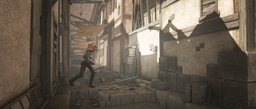} &
    \fcolorbox{gray!50}{white}{\includegraphics[width=19.1mm]{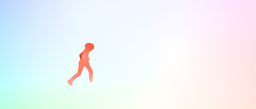}} &
    \darkleftbox{19.1mm}{attacks_smaller/alley_2/frame_0002.png}{\scriptsize Orig.} &
    \includegraphics[width=19.1mm]{attacks_smaller/alley_2/frame_0003.png} &
    \fcolorbox{gray!50}{white}{\includegraphics[width=19.1mm]{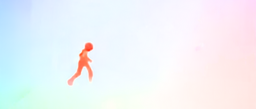}} &
    \darkleftbox{19.1mm}{attacks_smaller/alley_2/frame_0002.png}{\scriptsize Orig.} &
    \includegraphics[width=19.1mm]{attacks_smaller/alley_2/frame_0003.png} &
    \fcolorbox{gray!50}{white}{\includegraphics[width=19.1mm]{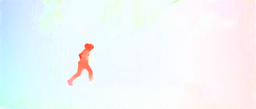}}
    \\[-3.5pt]
    \darkleftbox{19.1mm}{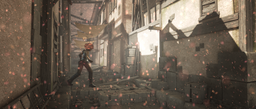}{\scriptsize Sparks rand.} &
    \includegraphics[width=19.1mm]{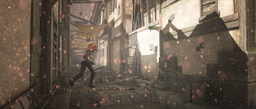} &
    \fcolorbox{gray!50}{white}{\includegraphics[width=19.1mm]{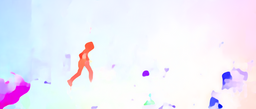}} &
    \darkleftbox{19.1mm}{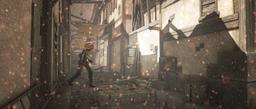}{\scriptsize Sparks rand.} &
    \includegraphics[width=19.1mm]{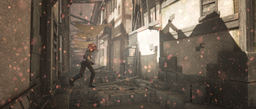} &
    \fcolorbox{gray!50}{white}{\includegraphics[width=19.1mm]{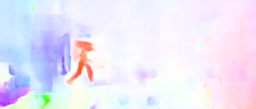}} &
    \darkleftbox{19.1mm}{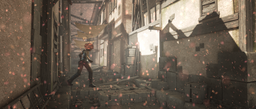}{\scriptsize Sparks rand.} &
    \includegraphics[width=19.1mm]{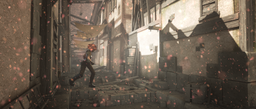} &
    \fcolorbox{gray!50}{white}{\includegraphics[width=19.1mm]{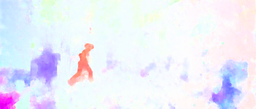}}
    \\[-3.5pt]
    \darkleftbox{19.1mm}{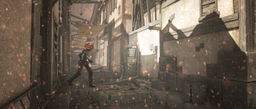}{\scriptsize Sparks adv.} &
    \includegraphics[width=19.1mm]{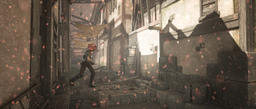} &
    \fcolorbox{gray!50}{white}{\includegraphics[width=19.1mm]{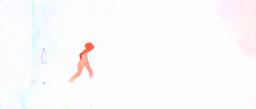}} &
    \darkleftbox{19.1mm}{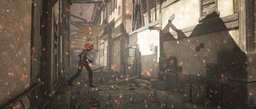}{\scriptsize Sparks adv.} &
    \includegraphics[width=19.1mm]{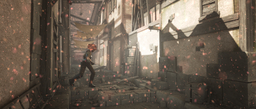} &
    \fcolorbox{gray!50}{white}{\includegraphics[width=19.1mm]{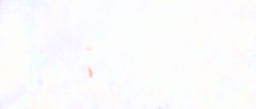}} &
    \darkleftbox{19.1mm}{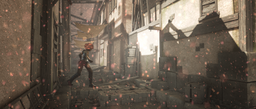}{\scriptsize Sparks adv.} &
    \includegraphics[width=19.1mm]{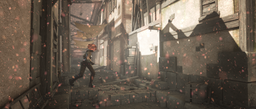} &
    \fcolorbox{gray!50}{white}{\includegraphics[width=19.1mm]{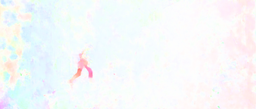}}
    \\
    \multicolumn{3}{c}{RAFT} & \multicolumn{3}{c}{GMA} & \multicolumn{3}{c}{FlowFormer}
    \\
    \darkleftbox{19.1mm}{attacks_smaller/market_5/frame_0001.png}{\scriptsize Orig.} &
    \includegraphics[width=19.1mm]{attacks_smaller/market_5/frame_0002.png} &
    \fcolorbox{gray!50}{white}{\includegraphics[width=19.1mm]{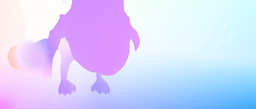}} &
    \darkleftbox{19.1mm}{attacks_smaller/market_5/frame_0001.png}{\scriptsize Orig.} &
    \includegraphics[width=19.1mm]{attacks_smaller/market_5/frame_0002.png} &
    \fcolorbox{gray!50}{white}{\includegraphics[width=19.1mm]{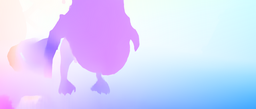}} &
    \darkleftbox{19.1mm}{attacks_smaller/market_5/frame_0001.png}{\scriptsize Orig.} &
    \includegraphics[width=19.1mm]{attacks_smaller/market_5/frame_0002.png} &
    \fcolorbox{gray!50}{white}{\includegraphics[width=19.1mm]{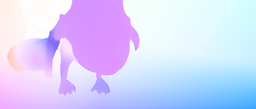}}
    \\[-3.5pt]
    \darkleftbox{19.1mm}{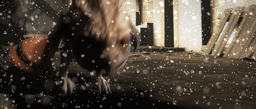}{\scriptsize Snow rand.} &
    \includegraphics[width=19.1mm]{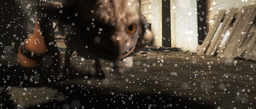} &
    \fcolorbox{gray!50}{white}{\includegraphics[width=19.1mm]{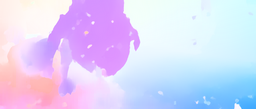}} &
    \darkleftbox{19.1mm}{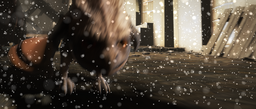}{\scriptsize Snow rand.} &
    \includegraphics[width=19.1mm]{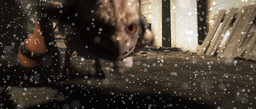} &
    \fcolorbox{gray!50}{white}{\includegraphics[width=19.1mm]{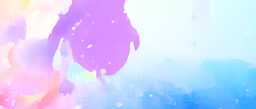}} &
    \darkleftbox{19.1mm}{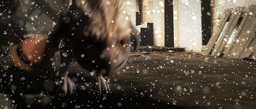}{\scriptsize Snow rand.} &
    \includegraphics[width=19.1mm]{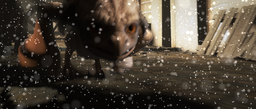} &
    \fcolorbox{gray!50}{white}{\includegraphics[width=19.1mm]{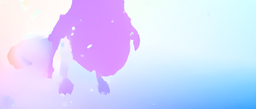}}
    \\[-3.5pt]
    \darkleftbox{19.1mm}{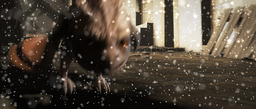}{\scriptsize Snow adv.} &
    \includegraphics[width=19.1mm]{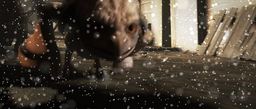} &
    \fcolorbox{gray!50}{white}{\includegraphics[width=19.1mm]{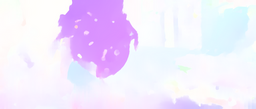}} &
    \darkleftbox{19.1mm}{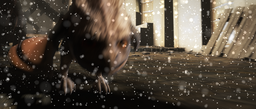}{\scriptsize Snow adv.} &
    \includegraphics[width=19.1mm]{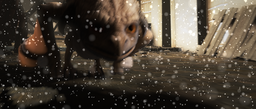} &
    \fcolorbox{gray!50}{white}{\includegraphics[width=19.1mm]{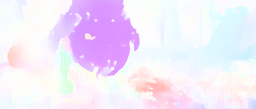}} &
    \darkleftbox{19.1mm}{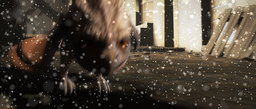}{\scriptsize Snow adv.} &
    \includegraphics[width=19.1mm]{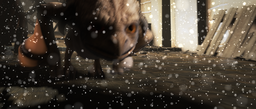} &
    \fcolorbox{gray!50}{white}{\includegraphics[width=19.1mm]{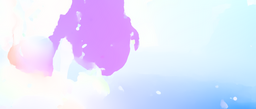}}
    \\
    \darkleftbox{19.1mm}{attacks_smaller/temple_3/frame_0003.png}{\scriptsize Orig.} &
    \includegraphics[width=19.1mm]{attacks_smaller/temple_3/frame_0004.png} &
    \fcolorbox{gray!50}{white}{\includegraphics[width=19.1mm]{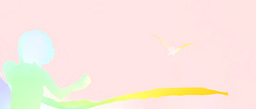}} &
    \darkleftbox{19.1mm}{attacks_smaller/temple_3/frame_0003.png}{\scriptsize Orig.} &
    \includegraphics[width=19.1mm]{attacks_smaller/temple_3/frame_0004.png} &
    \fcolorbox{gray!50}{white}{\includegraphics[width=19.1mm]{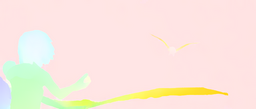}} &
    \darkleftbox{19.1mm}{attacks_smaller/temple_3/frame_0003.png}{\scriptsize Orig.} &
    \includegraphics[width=19.1mm]{attacks_smaller/temple_3/frame_0004.png} &
    \fcolorbox{gray!50}{white}{\includegraphics[width=19.1mm]{attacks_smaller/0112_flow_init_RAFT_rain_15.png}}
    \\[-3.5pt]
    \darkleftbox{19.1mm}{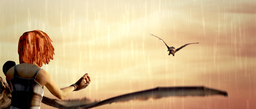}{\scriptsize Rain rand.} &
    \includegraphics[width=19.1mm]{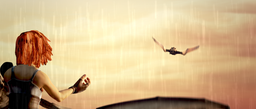} &
    \fcolorbox{gray!50}{white}{\includegraphics[width=19.1mm]{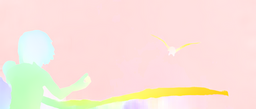}} &
    \darkleftbox{19.1mm}{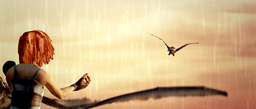}{\scriptsize Rain rand.} &
    \includegraphics[width=19.1mm]{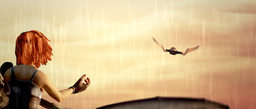} &
    \fcolorbox{gray!50}{white}{\includegraphics[width=19.1mm]{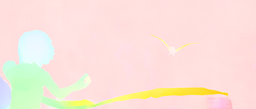}} &
    \darkleftbox{19.1mm}{attacks_smaller/0112_img1_init_RAFT_rain_15.png}{\scriptsize Rain rand.} &
    \includegraphics[width=19.1mm]{attacks_smaller/0112_img2_init_RAFT_rain_15.png} &
    \fcolorbox{gray!50}{white}{\includegraphics[width=19.1mm]{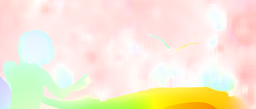}}
    \\[-3.5pt]
    \darkleftbox{19.1mm}{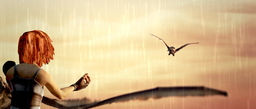}{\scriptsize Rain adv.} &
    \includegraphics[width=19.1mm]{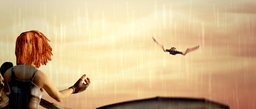} &
    \fcolorbox{gray!50}{white}{\includegraphics[width=19.1mm]{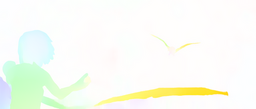}} &
    \darkleftbox{19.1mm}{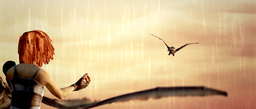}{\scriptsize Rain adv.} &
    \includegraphics[width=19.1mm]{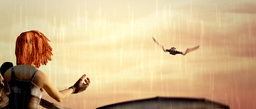} &
    \fcolorbox{gray!50}{white}{\includegraphics[width=19.1mm]{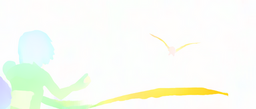}} &
    \darkleftbox{19.1mm}{attacks_smaller/0112_img1_best_RAFT_rain_15.png}{\scriptsize Rain adv.} &
    \includegraphics[width=19.1mm]{attacks_smaller/0112_img2_best_RAFT_rain_15.png} &
    \fcolorbox{gray!50}{white}{\includegraphics[width=19.1mm]{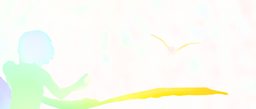}}
    \\
    \darkleftbox{19.1mm}{attacks_smaller/alley_2/frame_0002.png}{\scriptsize Orig.} &
    \includegraphics[width=19.1mm]{attacks_smaller/alley_2/frame_0003.png} &
    \fcolorbox{gray!50}{white}{\includegraphics[width=19.1mm]{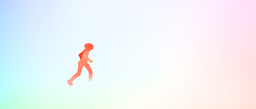}} &
    \darkleftbox{19.1mm}{attacks_smaller/alley_2/frame_0002.png}{\scriptsize Orig.} &
    \includegraphics[width=19.1mm]{attacks_smaller/alley_2/frame_0003.png} &
    \fcolorbox{gray!50}{white}{\includegraphics[width=19.1mm]{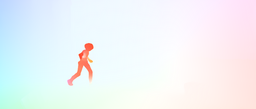}} &
    \darkleftbox{19.1mm}{attacks_smaller/alley_2/frame_0002.png}{\scriptsize Orig.} &
    \includegraphics[width=19.1mm]{attacks_smaller/alley_2/frame_0003.png} &
    \fcolorbox{gray!50}{white}{\includegraphics[width=19.1mm]{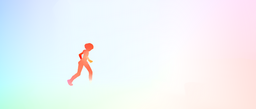}}
    \\[-3.5pt]
    \darkleftbox{19.1mm}{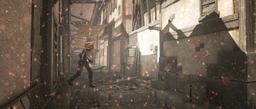}{\scriptsize Sparks rand.} &
    \includegraphics[width=19.1mm]{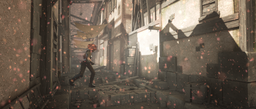} &
    \fcolorbox{gray!50}{white}{\includegraphics[width=19.1mm]{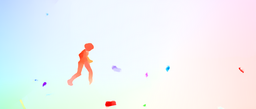}} &
    \darkleftbox{19.1mm}{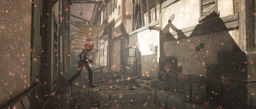}{\scriptsize Sparks rand.} &
    \includegraphics[width=19.1mm]{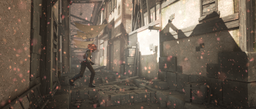} &
    \fcolorbox{gray!50}{white}{\includegraphics[width=19.1mm]{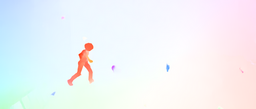}} &
    \darkleftbox{19.1mm}{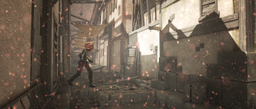}{\scriptsize Sparks rand.} &
    \includegraphics[width=19.1mm]{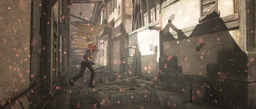} &
    \fcolorbox{gray!50}{white}{\includegraphics[width=19.1mm]{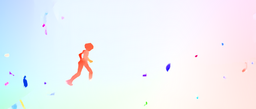}}
    \\[-3.5pt]
    \darkleftbox{19.1mm}{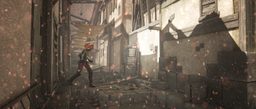}{\scriptsize Sparks adv.} &
    \includegraphics[width=19.1mm]{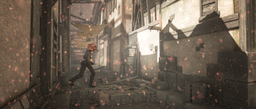} &
    \fcolorbox{gray!50}{white}{\includegraphics[width=19.1mm]{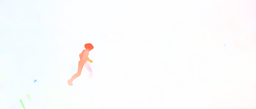}} &
    \darkleftbox{19.1mm}{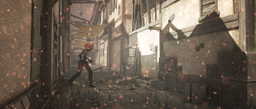}{\scriptsize Sparks adv.} &
    \includegraphics[width=19.1mm]{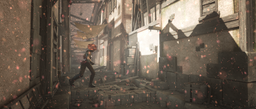} &
    \fcolorbox{gray!50}{white}{\includegraphics[width=19.1mm]{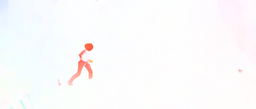}} &
    \darkleftbox{19.1mm}{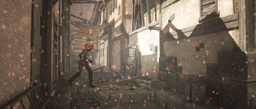}{\scriptsize Sparks adv.} &
    \includegraphics[width=19.1mm]{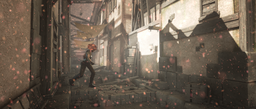} &
    \fcolorbox{gray!50}{white}{\includegraphics[width=19.1mm]{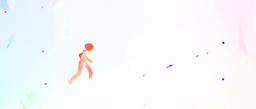}}    
\end{tabular}
\vspace{1mm}
\caption{Qualitative results for weather attacks on optical flow predictions for FlowNet2~\cite{Ilg2017Flownet2Evolution}, FlowNetCRobust~\cite{Schrodi2022TowardsUnderstandingAdversarial}, SpyNet~\cite{Ranjan2017OpticalFlowEstimation}, RAFT~\cite{Teed2020RaftRecurrentAll}, GMA~\cite{Jiang2021LearningEstimateHidden} and FlowFormer~\cite{Huang2022FlowformerTransformerArchitecture} (\emph{top left to bottom right}). Images from the Sintel final dataset, with \emph{random} initialization and after \emph{adversarial} weather optimization towards zero-target (white flow). See supplement for more visualizations.}
\label{fig:strongattack}
\vspace{-1mm}
\end{figure*}
To understand the impact of adversarial particles on optical flow, we consider artificial weather, initialized with 3000 gray particles that fall down without motion blur.
Per initial particle ($\delta\!=\!0$), we then adversarially optimize offsets to their positions before $\de$ and after $\dt$ the motion, colors $\dcol$ and transparencies $\dtransp$.
For optimizing $\dcol$, $\dtransp$ and $\dhue$, we set the learning rate to 1e-3 and use a sub\-set \enquote{Sintel-tr115} with 115 frame pairs (the first five per scene) of Sintel-train.

Tab.~\ref{table:optimvariables} summarizes the adversarial robustness for the different optimization parameters on all tested optical flow methods.
Considering single parameters, the particle offset $\de$ before the motion has the strongest influence.
That motion offsets have the strongest influence on the motion estimation is intuitively plausible.
However, our motion model favors the first motion offset over the second, as $\de$ affects both frames while $\dt$ affects only the second one.
Jointly optimizing all parameters generally leads to the worst degradation of optical flow estimates.
Yet, focusing on motion parameters $\dmot$ or hue parameters $\dhue$ alone also strongly degrades performance.
Interestingly, the tested flow methods show either a high sensitivity towards motion, for RAFT and GMA, or a high sensitivity towards hues, for FlowNetCRobust, SpyNet and Flow\-Former.
For the latter, optimizing hues even yields the strongest degradation overall.
This insight is valuable for color-reduced environments, \eg night scenes, where a greater independence of the color representation may be wanted.

\medskip \noindent
\textbf{Robustness against snow, rain, sparks and fog.}
\begin{table}[tb]
\small
\begin{center}
\begin{tabular}{lrrrrrr}
\toprule
Attack  &  \multicolumn{1}{c}{\rotatebox{45}{FN2}} & \multicolumn{1}{c}{\rotatebox{45}{FNCR}} & \multicolumn{1}{c}{\rotatebox{45}{SpyNet}} & \multicolumn{1}{c}{\rotatebox{45}{RAFT}} & \multicolumn{1}{c}{\rotatebox{45}{GMA}} & \multicolumn{1}{c}{\rotatebox{45}{FF}} \\
\midrule
snow                &          21.37  &          18.23  &   \textbf{9.99} &   \textbf{11.20} &     \textbf{10.90} &       \textbf{7.22} \\
rain                &          21.95  &  \textbf{19.85} &           8.37  &            9.53  &              8.22  &               5.82  \\
sparks              &  \textbf{22.76} &          19.54  &           8.25  &            8.72  &              9.39  &               6.41  \\
fog                 &           2.32  &           3.37  &           2.28  &            0.92  &              0.97  &               0.73  \\
\bottomrule
\end{tabular}
\vspace{1.5mm}
\caption{Adversarial robustness AEE$(\flow,\fadv)$ $\downarrow$~\cite{Schmalfuss2022PerturbationConstrainedAdversarial} for \emph{adversarial snow, rain, sparks} and \emph{fog} on Sintel-tr115. Worst robustness bold.}
\label{table:weatherattacks}
\vspace{-1mm}
\end{center}
\end{table}
\begin{table}
\small
\begin{center}
\begin{tabular}{@{\ }l@{\ }r@{\ }r@{}r@{\ \ }r@{\ \ }r@{\ \ }r@{\ }}
\toprule
Attack & \multicolumn{1}{c}{\rotatebox{45}{FN2}} & \multicolumn{1}{c}{\rotatebox{45}{FNCR}} & \multicolumn{1}{c}{\rotatebox{45}{SpyNet}} & \multicolumn{1}{c}{\rotatebox{45}{RAFT}} & \multicolumn{1}{c}{\rotatebox{45}{GMA}} & \multicolumn{1}{c}{\rotatebox{45}{FF}} \\
\midrule
PCFA~\cite{Schmalfuss2022PerturbationConstrainedAdversarial}    &           11.77  &          13.82  &          7.83  &  \textbf{12.96} & \textbf{12.83} &   \textbf{14.68}  \\
I-FGSM~\cite{Schrodi2022TowardsUnderstandingAdversarial}        &            7.58  &          13.69  &          5.07  &          11.07  &         11.40  &         12.35  \\
Snow (ours)                                                     &   \textbf{16.83} &  \textbf{16.28} &  \textbf{9.94} &          10.32  &          9.85  &          7.10  \\
\bottomrule
\end{tabular}
\vspace{0.5mm}
\caption{Adversarial robustness AEE$(\flow,\fadv)$ $\downarrow$~\cite{Schmalfuss2022PerturbationConstrainedAdversarial} for \emph{different attacks} on Sintel train, the worst robustness per method is bold.}
\label{table:of_attacks}
\vspace{0mm}
\end{center}
\end{table}
Next, we transition to more natural attacks with snow, rain, sparks and fog.
We optimize all parameters for snow, rain and sparks, but do not optimize $\dt$ for fog, keeping it static in the scene.
Tab.~\ref{table:weatherattacks} summarizes the optical flow robustness against adversarial weather, again on Sintel-tr115.
For adversarial weather, the methods rank similar to pure augmentation, \cf Tab.~\ref{table:augmentations}, but the optimization amplifies optical flow changes.
For every weather, lower-quality methods, \eg FlowNet2, are very vulnerable while high-quality methods, \eg FlowFormer, are comparatively robust against any weather.
For GMA, Fig.~\ref{fig:attackweatheroverview} visualizes the attacked weather and resulting flows.
Remarkably, moving particles eradicate the estimated motion despite their constant movement due to falling and camera motion.
When we compare randomly initialized particles to their adversarial counterparts in Fig.~\ref{fig:strongattack} their positions hardly differ, making the adversarial sample indistinguishable from random weather to human observers.
As adversarial snow greatly affects all optical flow methods, we select it for further analysis.

\medskip \noindent
\textbf{Comparison to L$_p$ attacks.}
\begin{table*}
\small
\begin{center}
\begin{tabular}{@{\ }lcc|c|cccc|cccc@{\ }}
\toprule
{} & \multicolumn{2}{c}{Sintel EPE $\downarrow$ (te.)}  & \multicolumn{1}{c}{KITTI $\downarrow$ (tr.)} & \multicolumn{4}{c}{Augmentation robustness $\downarrow$} &  \multicolumn{4}{c}{Attack robustness $\downarrow$} \\
\cmidrule(l{0.5em}r{0.5em}){2-3} \cmidrule(l{0.5em}r{0.5em}){4-4} \cmidrule(l{0.5em}r{0.5em}){5-8} \cmidrule(l{0.5em}r{0.5em}){9-12}
   Snow & clean & final & F1-all & snow & rain & sparks & fog & snow & rain & sparks & fog \\
\midrule
\phantom{10}0\%    &         1.642  &         3.167  &             5.65  &         4.19  &         3.60  &         3.64  &         3.54  &         9.93   &         8.02   &         8.47   & \textbf{0.87}  \\
\phantom{1}50\%    &         1.589  & \textbf{3.155} &     \textbf{5.54} &         0.91  &         1.66  & \textbf{1.29} & \textbf{3.52} &         3.76   &         5.96   &         5.68   &         0.93   \\
          100\%   & \textbf{1.551} &         3.384   &             5.69  & \textbf{0.83} & \textbf{1.37} &         1.32  &         3.57  & \textbf{3.48}  & \textbf{5.61}  & \textbf{5.49}  &         1.04   \\
\bottomrule
\end{tabular}
\vspace{2mm}
\caption{Training RAFT~\cite{Teed2020RaftRecurrentAll} with \emph{0, 50} or \emph{100\% snowy} Sintel-final frames during the Sintel/KITTI (S/K) training phase~\cite{Teed2020RaftRecurrentAll} The \emph{quality} is measured on Sintel test and KITTI train, robustness values for \emph{weather augmentations} on Sintel test and \emph{weather attacks} on Sintel-tr115.}
\label{table:raft_ckpt}
\vspace{-1.5mm}
\end{center}
\end{table*}
To conclude our attack evaluation, we compare our adversarial snow attack to previous attacks on optical flow and analyze the performance of optical flow methods in detail.
Tab.~\ref{table:of_attacks} compares the robustness of optical flow methods under two L$_p$ attacks to our non-L$_p$ attack with adversarial snow on the full Sintel training set.
The L$_2$ attack PCFA~\cite{Schmalfuss2022PerturbationConstrainedAdversarial} is the strongest adversarial attack in the literature, while I-FGSM~\cite{Schrodi2022TowardsUnderstandingAdversarial} is a weaker L$_\infty$ attack.
Despite being much more constrained by its physically plausible motion, our adversarial snow can compete with PCFA in terms of induced flow perturbation.

Surprisingly, high-quality methods like RAFT, GMA or FlowFormer that suffer most from L$_p$ attacks~\cite{Schmalfuss2022PerturbationConstrainedAdversarial} offer the best robustness towards adversarial snow.
Instead, lower-quality methods like FlowNet2 and SpyNet that are most robust towards L$_p$ attacks alter their predictions disproportionately to the added snow particles -- or any other particle-based weather (\cf Tab.~\ref{table:weatherattacks}).
We ascribe the better weather robustness to the more detailed flow estimations of high-quality methods, which detect the localized motion of single particles (\cf Fig.~\ref{fig:strongattack}, snow and sparks on RAFT and FlowFormer, where circular particles are visible).
The less accurate methods FlowNet2, FlowNetCRobust and SpyNet instead propagate the detected particle motion over larger areas, rather than attributing it to small moving objects (\cf Fig.~\ref{fig:strongattack}, rows 2/3, where flow predictions have few details).
Notably, the robustness of FlowNetCRobust against patch attacks as reported in~\cite{Schrodi2022TowardsUnderstandingAdversarial} does not transfer, making it one of the most vulnerable methods irrespective of the attack.

\subsection{Training with weather}
\label{sec:TrainingWithWeather}

As all optical flow methods change their predictions significantly in the presence of weather, we end our experiments by presenting a robustifying training strategy.
Here, we choose RAFT~\cite{Teed2020RaftRecurrentAll}, which is the baseline architecture for GMA and FlowFormer.
We retrain RAFT from the author-provided C+T checkpoint according to their training protocol~\cite{Teed2020RaftRecurrentAll} but augment 0\%, 50\% or 100\% of the Sintel final training data with \emph{random} snow.
We evaluate the quality, and the robustness towards random augmentations as well as optimized weather attacks, \cf Tab.~\ref{table:augmentations} and Tab.~\ref{table:weatherattacks}.

Tab.~\ref{table:raft_ckpt} summarizes the results.
Compared to standard training, augmenting any percentage of Sintel-final frames with snow clearly improves the robustness.
Furthermore, augmenting half of Sintel clean improves the quality on all datasets and snows a better generalization.
It is remarkable that training with random snow has such a positive effect on robustness and quality~\cite{Teed2020RaftRecurrentAll}, because training with L$_p$ perturbations does not generally improve the robustness towards adversarial perturbations.
For example, FlowFormer~\cite{Huang2022FlowformerTransformerArchitecture} augments its training with random noise, but is highly vulnerable against L$_p$ attacks, \cf Tab.~\ref{table:of_attacks}.
Therefore, adversarial training~\cite{Madry2018TowardsDeepLearning} is commonly used to improve the robustness against L$_p$ attacks.
However, it (i) significantly increases the training time because adversarial samples are continuously included, leading to a slowly-converging training and (ii) often lowers the quality for non-attacked samples.
Both drawbacks are not observed for training with snow augmentations.
This makes it particularly noteworthy that simple augmentation with 50\% non-L$_p$ snow improves robustness, quality and generalization at the same time.

\section{Limitations}
\label{sec:Limitations}
Although we focus on realism, our attack does not aim at threatening optical flow methods in the real world, 
where manipulating weather is clearly impossible.
While this hol\-ds for most optical flow attacks~\cite{Koren2021ConsistentSemanticAttacks,Schmalfuss2022PerturbationConstrainedAdversarial,Schrodi2022TowardsUnderstandingAdversarial,Agnihotri2023CospgdUnifiedWhite} our adversarial weather assesses methods under worst-case weather conditions, which is a more realistic scenario that even allows sig\-ni\-ficant alterations without being noticeably adversarial.
Further\-more, optimizing snow on Sintel-test may take several days on a Nvidia A100 GPU, but these higher computational costs are tolerable in an offline benchmarking setting.

\section{Conclusion}
\label{sec:conclusion}

In this paper, we developed a novel framework for adversarial attacks on motion estimation with realistic weather.
We proposed a differentiable particle renderer that can be used to generate adversarial weather with a strong impact on optical flow methods.
With its realistic appearance, our adversarial weather is hard to notice; yet it lets optical flow networks predict zero-flow although the particles undergo both individual and camera motion.
Surprisingly, accurate methods that are very vulnerable to L$_p$ attacks appear to be more robust towards adversarially optimized weather, as they detect the motion of single particles rather than propagating it into the wider image.
Additionally, we find that augmenting a network's training with unoptimized weather not only improves the robustness towards weather augmentations and attacks but also increases generalization across datasets at a much lower cost than adversarial training.
Finally, our weather attacks could easily be extended to problems that also require 3D-awareness or temporal motion consistency, like monocular depth estimation~\cite{Yamanaka2021SimultaneousAttackCnn,Hu2019AnalysisDeepNetworks}, stereo reconstruction~\cite{Wong2021StereopagnosiaFoolingStereo,Berger2022StereoscopicUniversalPerturbations} or scene flow computation.

\medskip \noindent
\textbf{Acknowledgments.}
Funded by the Deutsche For\-schungs\-gemein\-schaft (DFG, German Research Foundation) -- Project-ID 251654672 -- TRR 161 (B04).
Jenny Schmalfuss is supported by the International Max Planck Research School for Intelligent Systems (IMPRS-IS).

{\small
\bibliographystyle{ieee_fullname}
\bibliography{egbib}

\begin{thebibliography}{10}\itemsep=-1pt

\bibitem{Agnihotri2023CospgdUnifiedWhite}
Shashank Agnihotri and Margret Keuper.
\newblock {CosPGD}: A unified white-box adversarial attack for pixel-wise
  prediction tasks.
\newblock In {\em arXiv preprint 2302.02213}. arXiv, 2023.

\bibitem{Arun1987LeastSquaresFitting}
K.~Somani Arun, Thomas~S. Huang, and Steven~D. Blostein.
\newblock Least-squares fitting of two 3-d point sets.
\newblock {\em IEEE Transactions on Pattern Analysis and Machine Intelligence
  (TPAMI)}, 9(5):698--700, 1987.

\bibitem{Berger2022StereoscopicUniversalPerturbations}
Zachary Berger, Parth Agrawal, Tyan~Yu Liu, Stefano Soatto, and Alex Wong.
\newblock Stereoscopic universal perturbations across different architectures
  and datasets.
\newblock In {\em Proc. IEEE/CVF Conference on Computer Vision and Pattern
  Recognition (CVPR)}, pages 15180--15190, 2022.

\bibitem{Butler2012NaturalisticOpenSource}
Daniel Butler, Jonas Wulff, Garrett Stanley, and Michael~J. Black.
\newblock A naturalistic open source movie for optical flow evaluation.
\newblock In {\em Proc. European Conference on Computer Vision (ECCV)}, pages
  611--625, 2012.

\bibitem{Carlini2017TowardsEvaluatingRobustness}
Nicholas Carlini and David Wagner.
\newblock Towards evaluating the robustness of neural networks.
\newblock In {\em IEEE Symposium on Security and Privacy (SP)}, pages 39--57,
  2017.

\bibitem{Dosovitskiy2015FlownetLearningOptical}
Alexey Dosovitskiy, Philipp Fischer, Eddy Ilg, Philip Hausser, Caner Hazirbas,
  Vladimir Golkov, Patrick van~der Smagt, Daniel Cremers, and Thomas Brox.
\newblock {FlowNet}: Learning optical flow with convolutional networks.
\newblock In {\em Proc. IEEE/CVF International Conference on Computer Vision
  (ICCV)}, pages 2758--2766, 2015.

\bibitem{Gao2021AdvhazeAdversarialHaze}
Ruijun Gao, Qing Guo, Felix Juefei-Xu, Hongkai Yu, and Wei Feng.
\newblock {AdvHaze}: Adversarial haze attack.
\newblock In {\em arXiv preprint 2104.13673}. arXiv, 2021.

\bibitem{Gao2022ScaleFreeTask}
Xiangbo Gao, Cheng Luo, Qinliang Lin, Weicheng Xie, Minmin Liu, Linlin Shen,
  Keerthy Kusumam, and Siyang Song.
\newblock Scale-free and task-agnostic attack: Generating photo-realistic
  adversarial patterns with patch quilting generator.
\newblock In {\em arXiv preprint 2208.06222}. arXiv, 2022.

\bibitem{Garg2006PhotorealisticRenderingRain}
Kshitiz Garg and Shree~K. Nayar.
\newblock Photorealistic rendering of rain streaks.
\newblock {\em ACM Transactions on Graphics (TOG)}, 25(3):996--1002, 2006.

\bibitem{Halder2019PhysicsBasedRendering}
Shirsendu Halder, Jean-Francois Lalonde, and Raoul de Charette.
\newblock Physics-based rendering for improving robustness to rain.
\newblock In {\em Proc. IEEE/CVF International Conference on Computer Vision
  (ICCV)}, pages 10202--10211, 2019.

\bibitem{Hendrycks2018BenchmarkingNeuralNetwork}
Dan Hendrycks and Thomas Dietterich.
\newblock Benchmarking neural network robustness to common corruptions and
  perturbations.
\newblock In {\em Proc. International Conference on Learning Representations
  (ICLR)}, pages 1--16, 2019.

\bibitem{Hu2019AnalysisDeepNetworks}
Junjie Hu and Takayuki Okatani.
\newblock Analysis of deep networks for monocular depth estimation through
  adversarial attacks with proposal of a defense method.
\newblock In {\em arXiv preprint 1911.08790}. arXiv, 2019.

\bibitem{Huang2022FlowformerTransformerArchitecture}
Zhaoyang Huang, Xiaoyu Shi, Chao Zhang, Qiang Wang, Ka~Chun Cheung, Hongwei
  Qin, Jifeng Dai, and Hongsheng Li.
\newblock {FlowFormer}: A transformer architecture for optical flow.
\newblock In {\em Proc. European Conference on Computer Vision (ECCV)}, pages
  668--685, 2022.

\bibitem{Ilg2017Flownet2Evolution}
Eddy Ilg, Nikolaus Mayer, Tonmoy Saikia, Margret Keuper, Alexey Dosovitskiy,
  and Thomas Brox.
\newblock {FlowNet} 2.0: Evolution of optical flow estimation with deep
  networks.
\newblock In {\em Proc. IEEE/CVF Conference on Computer Vision and Pattern
  Recognition (CVPR)}, pages 2462--2470, 2017.

\bibitem{Jiang2021LearningEstimateHidden}
Shihao Jiang, Dylan Campbell, Yao Lu, Hongdong Li, and Richard Hartley.
\newblock Learning to estimate hidden motions with global motion aggregation.
\newblock In {\em Proc. IEEE/CVF International Conference on Computer Vision
  (ICCV)}, pages 9772--9781, 2021.

\bibitem{Kang2019TestingRobustnessUnforeseen}
Daniel Kang, Yi Sun, Dan Hendrycks, Tom Brown, and Jacob Steinhardt.
\newblock Testing robustness against unforeseen adversaries.
\newblock In {\em arXiv preprint 1908.08016}. arXiv, 2019.

\bibitem{Koren2021ConsistentSemanticAttacks}
Tom Koren, Lior Talker, Michael Dinerstein, and Ran Vitek.
\newblock Consistent semantic attacks on optical flow.
\newblock In {\em Proc. Asian Conference on Computer Vision (ACCV)}, pages
  1658--1674, 2022.

\bibitem{Li2018RobustOpticalFlow}
Ruoteng Li, Robby~T. Tan, and Loong-Fah Cheong.
\newblock Robust optical flow in rainy scenes.
\newblock In {\em Proc. European Conference on Computer Vision (ECCV)}, pages
  288--304, 2018.

\bibitem{Li2019RainflowOpticalFlow}
Ruoteng Li, Robby~T. Tan, Loong-Fah Cheong, Angelica~I. Aviles-Rivero, Qingnan
  Fan, and Carola-Bibiane Schonlieb.
\newblock {RainFlow}: Optical flow under rain streaks and rain veiling effect.
\newblock In {\em Proc. IEEE/CVF International Conference on Computer Vision
  (ICCV)}, pages 7304--7313, 2019.

\bibitem{Li2021WeatherGanMulti}
Xuelong Li, Kai Kou, and Bin Zhao.
\newblock Weather {GAN}: Multi-domain weather translation using generative
  adversarial networks.
\newblock In {\em arXiv preprint 2103.05422}. arXiv, 2021.

\bibitem{Machiraju_2020_WACV}
Harshitha Machiraju and Vineeth~N Balasubramanian.
\newblock A little fog for a large turn.
\newblock In {\em Proc. IEEE/CVF Winter Conference on Applications of Computer
  Vision (WACV)}, pages 2891--2900, 2020.

\bibitem{Madry2018TowardsDeepLearning}
Aleksander Madry, Aleksandar Makelov, Ludwig Schmidt, Dimitris Tsipras, and
  Adrian Vladu.
\newblock Towards deep learning models resistant to adversarial attacks.
\newblock In {\em Proc. International Conference on Learning Representations
  (ICML)}, pages 1--10, 2018.

\bibitem{Marchisio2022FakeweatherAdversarialAttacks}
Alberto Marchisio, Giovanni Caramia, Maurizio Martina, and Muhammad Shafique.
\newblock {fakeWeather}: Adversarial attacks for deep neural networks emulating
  weather conditions on the camera lens of autonomous systems.
\newblock In {\em Proc. International Joint Conference on Neural Networks
  (IJCNN)}, pages 1--9, 2022.

\bibitem{Mayer2016LargeDatasetTrain}
Nikolaus Mayer, Eddy Ilg, Philip Hausser, Philipp Fischer, Daniel Cremers,
  Alexey Dosovitskiy, and Thomas Brox.
\newblock A large dataset to train convolutional networks for disparity,
  optical flow, and scene flow estimation.
\newblock In {\em Proc. IEEE/CVF Conference on Computer Vision and Pattern
  Recognition (CVPR)}, pages 4040--4048, 2016.

\bibitem{Mehl2023SpringHighResolution}
Lukas Mehl, Jenny Schmalfuss, Azin Jahedi, Yaroslava Nalivayko, and Andr\'es
  Bruhn.
\newblock Spring: A high-resolution high-detail dataset and benchmark for scene
  flow, optical flow and stereo.
\newblock In {\em arXiv preprint 2303.01943}. arXiv, 2023.

\bibitem{Menze2015ObjectSceneFlow}
Moritz Menze and Andreas Geiger.
\newblock Object scene flow for autonomous vehicles.
\newblock In {\em Proc. IEEE/CVF Conference on Computer Vision and Pattern
  Recognition (CVPR)}, pages 3061--3070, 2015.

\bibitem{Meshkin2007SortIndependentAlpha}
Houman Meshkin.
\newblock Sort-independent alpha blending.
\newblock In {\em Game Developers Conference}, 2007.

\bibitem{Michaelis2019BenchmarkingRobustnessObject}
Claudio Michaelis, Benjamin Mitzkus, Robert Geirhos, Evgenia Rusak, Oliver
  Bringmann, Akexander~S. Ecker, Matthias Bethge, and Wieland Brendel.
\newblock Benchmarking robustness in object detection: Autonomous driving when
  winter is coming.
\newblock In {\em Proc. Conference on Neural Information Processing Systems
  Workshops (NeurIPSW)}, 2019.

\bibitem{Ni2021ControllingRainRemoval}
Siqi Ni, Xueyun Cao, Tao Yue, and Xuemei Hu.
\newblock Controlling the rain: From removal to rendering.
\newblock In {\em Proc. IEEE/CVF Conference on Computer Vision and Pattern
  Recognition (CVPR)}, pages 6328--6337, 2021.

\bibitem{Niklaus2018PytorchSpyNet}
Simon Niklaus.
\newblock A reimplementation of {SPyNet} using {PyTorch}, 2018.

\bibitem{Ranjan2017OpticalFlowEstimation}
Anurag Ranjan and Michael~J. Black.
\newblock Optical flow estimation using a spatial pyramid network.
\newblock In {\em Proc. IEEE/CVF Conference on Computer Vision and Pattern
  Recognition (CVPR)}, pages 4161--4170, 2017.

\bibitem{Ranjan2019AttackingOpticalFlow}
Anurag Ranjan, Joel Janai, Andreas Geiger, and Michael~J. Black.
\newblock Attacking optical flow.
\newblock In {\em Proc. IEEE/CVF International Conference on Computer Vision
  (ICCV)}, pages 2004--2013, 2019.

\bibitem{Reda2017Flownet2PytorchPytorch}
Fitsum Reda, Robert Pottorff, Jon Barker, and Bryan Catanzaro.
\newblock flownet2-pytorch: Pytorch implementation of {FlowNet} 2.0: Evolution
  of optical flow estimation with deep networks, 2017.

\bibitem{Ren2020DeepSnowSynthesizing}
Christopher~X. Ren, Amanda Ziemann, James Theiler, and Alice M.~S. Durieux.
\newblock Deep snow: Synthesizing remote sensing imagery with generative
  adversarial nets.
\newblock In {\em Proc. SPIE Defense + Commercial Sensing}, pages 196--205,
  2020.

\bibitem{Rousseau2006RealisticRealTime}
Pierre Rousseau, Vincent Jolivet, and Djamchid Ghazanfarpour.
\newblock Realistic real-time rain rendering.
\newblock {\em Computers \& Graphics}, pages 507--518, 2006.

\bibitem{Sakaino2009FallingSnowMotion}
Hidetomo Sakaino, Yang Shen, Yuanhang Pang, and Lizhuang Ma.
\newblock Falling snow motion estimation based on a semi-transparent and
  particle trajectory model.
\newblock In {\em Proc. IEEE International Conference on Image Processing
  (ICIP)}, pages 1609--1612, 2009.

\bibitem{Sava2022AssessingImpactTransformations}
Paul~Andrei Sava, Jan-Philipp Schulze, Philip Sperl, and Konstantin
  B\"{o}ttinger.
\newblock Assessing the impact of transformations on physical adversarial
  attacks.
\newblock In {\em Proc. ACM Workshop on Artificial Intelligence and Security
  (AiSec)}, pages 79--90, 2022.

\bibitem{Schmalfuss2022AttackingMotionEstimation}
Jenny Schmalfuss, Lukas Mehl, and Andr\'{e}s Bruhn.
\newblock Attacking motion estimation with adversarial snow.
\newblock {\em ECCV 2022 Workshop on Adversarial Robustness in the Real World
  (ECCV-AROW)}, 2022.

\bibitem{Schmalfuss2022PerturbationConstrainedAdversarial}
Jenny Schmalfuss, Philipp Scholze, and Andr\'es Bruhn.
\newblock A perturbation-constrained adversarial attack for evaluating the
  robustness of optical flow.
\newblock In {\em Proc. European Conference on Computer Vision (ECCV)}, pages
  183--200, 2022.

\bibitem{Schrodi2022TowardsUnderstandingAdversarial}
Simon Schrodi, Tonmoy Saikia, and Thomas Brox.
\newblock Towards understanding adversarial robustness of optical flow
  networks.
\newblock In {\em Proc. IEEE/CVF Conference on Computer Vision and Pattern
  Recognition (CVPR)}, pages 8916--8924, 2022.

\bibitem{Starik2003SimulationRainVideos}
Sonia Starik and Michael Werman.
\newblock Simulation of rain in videos.
\newblock In {\em Proc. IEEE/CVF International Conference on Computer Vision
  Workshops (ICCVW)}, pages 406--409, 2003.

\bibitem{Teed2020RaftRecurrentAll}
Zachary Teed and Jia Deng.
\newblock {RAFT}: Recurrent all-pairs field transforms for optical flow.
\newblock In {\em Proc. European Conference on Computer Vision (ECCV)}, pages
  402--419, 2020.

\bibitem{Tremblay2021RainRenderingEvaluating}
Maxime Tremblay, Shirsendu~Sukanta Halder, Raoul De~Charette, and
  Jean-Fran{\c{c}}ois Lalonde.
\newblock Rain rendering for evaluating and improving robustness to bad
  weather.
\newblock {\em International Journal of Computer Vision (IJCV)},
  129(2):341--360, 2021.

\bibitem{Volk2019TowardsRobustCnn}
Georg Volk, Stefan Müller, Alexander von Bernuth, Dennis Hospach, and Oliver
  Bringmann.
\newblock Towards robust {CNN}-based object detection through augmentation with
  synthetic rain variations.
\newblock In {\em Proc. IEEE Intelligent Transportation Systems Conference
  (ITSC)}, pages 285--292, 2019.

\bibitem{Bernuth2019SimulatingPhotoRealistic}
Alexander von Bernuth, Georg Volk, and Oliver Bringmann.
\newblock Simulating photo-realistic snow and fog on existing images for
  enhanced {CNN} training and evaluation.
\newblock In {\em Proc. IEEE Intelligent Transportation Systems Conference
  (ITSC)}, pages 41--46, 2019.

\bibitem{Wang2021RainGenerationRain}
Hong Wang, Zongsheng Yue, Qi Xie, Qian Zhao, Yefeng Zheng, and Deyu Meng.
\newblock From rain generation to rain removal.
\newblock In {\em Proc. IEEE/CVF Conference on Computer Vision and Pattern
  Recognition (CVPR)}, pages 14791--14801, 2021.

\bibitem{Wang2021WhenHumanPose}
Jiahang Wang, Sheng Jin, Wentao Liu, Weizhong Liu, Chen Qian, and Ping Luo.
\newblock When human pose estimation meets robustness: Adversarial algorithms
  and benchmarks.
\newblock In {\em Proc. IEEE/CVF Conference on Computer Vision and Pattern
  Recognition (CVPR)}, pages 11855--11864, 2021.

\bibitem{Wei2021DeraincycleganRainAttentive}
Yanyan Wei, Zhao Zhang, Yang Wang, Mingliang Xu, Yi Yang, Shuicheng Yan, and
  Meng Wang.
\newblock {DerainCycleGAN}: Rain attentive {CycleGAN} for single image
  deraining and rainmaking.
\newblock {\em IEEE Transaction on Image Processing (TIP)}, 30:4788--4801,
  2021.

\bibitem{Wiesemann2016FogAugmentationRoad}
Thomas Wiesemann and Xiaoyi Jiang.
\newblock Fog augmentation of road images for performance analysis of traffic
  sign detection algorithms.
\newblock In {\em Proc. International Conference on Advanced Concepts for
  Intelligent Vision Systems (ACVIS)}, pages 685--697, 2016.

\bibitem{Wong2021StereopagnosiaFoolingStereo}
Alex Wong, Mukund Mundhra, and Stefano Soatto.
\newblock Stereopagnosia: Fooling stereo networks with adversarial
  perturbations.
\newblock {\em Proc. AAAI Conference on Artificial Intelligence (AAAI)}, pages
  2879--2888, 2021.

\bibitem{Yamanaka2021SimultaneousAttackCnn}
Koichiro Yamanaka, Keita Takahashi, Toshiaki Fujii, and Ryuraroh Matsumoto.
\newblock Simultaneous attack on {CNN}-based monocular depth estimation and
  optical flow estimation.
\newblock {\em IEICE Transactions on Information and Systems}, pages 785--788,
  2021.

\bibitem{Yan2020OpticalFlowDense}
Wending Yan, Aashish Sharma, and Robby~T. Tan.
\newblock Optical flow in dense foggy scenes using semi-supervised learning.
\newblock In {\em Proc. IEEE/CVF Conference on Computer Vision and Pattern
  Recognition (CVPR)}, pages 13259--13268, 2020.

\bibitem{Yan2022OpticalFlowEstimation}
Wending Yan, Aashish Sharma, and Robby~T. Tan.
\newblock Optical flow estimation in dense foggy scenes with domain-adaptive
  networks.
\newblock {\em IEEE Transactions on Artificial Intelligence (AI)}, pages 1--12,
  2022.

\bibitem{Ye2021ClosingLoopJoint}
Yuntong Ye, Yi Chang, Hanyu Zhou, and Luxin Yan.
\newblock Closing the loop: Joint rain generation and removal via disentangled
  image translation.
\newblock In {\em Proc. IEEE/CVF Conference on Computer Vision and Pattern
  Recognition (CVPR)}, pages 2053--2062, 2021.

\bibitem{Zhai2020AdversarialRainAttack}
Liming Zhai, Felix Juefei-Xu, Qing Guo, Xiaofei Xie, Lei Ma, Wei Feng,
  Shengchao Qin, and Yang Liu.
\newblock Adversarial rain attack and defensive deraining for {DNN} perception.
\newblock In {\em arXiv preprint 2009.09205}. arXiv, 2020.

\bibitem{Zhong2022ShadowsCanBe}
Yiqi Zhong, Xianming Liu, Deming Zhai, Junjun Jiang, and Xiangyang Ji.
\newblock Shadows can be dangerous: Stealthy and effective physical-world
  adversarial attack by natural phenomenon.
\newblock In {\em Proc. IEEE/CVF Conference on Computer Vision and Pattern
  Recognition (CVPR)}, pages 15345--15354, 2022.

\end{thebibliography}
}

\clearpage

\maketitle
\appendix
\setcounter{table}{0}
\renewcommand{\thetable}{A\arabic{table}}
\setcounter{figure}{0}
\renewcommand{\thefigure}{A\arabic{figure}}

\twocolumn[\centering \Large\bf%
Distracting Downpour:\\Adversarial Weather Attacks for Motion Estimation\\-- Supplementary Material --\vspace*{1.5\baselineskip}
]

\section{Additional Material for Experiments}

We provide the code to generate the weather augmentations and run all adversarial weather attacks at \href{https://github.com/cv-stuttgart/DistractingDownpour}{https://github.com/cv-stuttgart/DistractingDownpour}.
The tested optical flow networks utilize the respective author-provided PyTorch implementations with Sintel-check\-points for FlowFormer~\cite{Huang2022FlowformerTransformerArchitecture}, GMA~\cite{Teed2020RaftRecurrentAll}, RAFT~\cite{Teed2020RaftRecurrentAll} and FlowNetCRobust~\cite{Schrodi2022TowardsUnderstandingAdversarial}.
For FlowNet2~\cite{Ilg2017Flownet2Evolution} and SpyNet~\cite{Ranjan2017OpticalFlowEstimation} we use the implementations from \cite{Reda2017Flownet2PytorchPytorch} and~\cite{Niklaus2018PytorchSpyNet}, respectively.

\subsection{Weather augmentations}

\subsubsection{Weather configurations and parameters}

In addition to the evaluation of particle-based augmentations in Main Tab.~\ref{table:augmentations}, we compare the two color-blending modes additive and alpha-blending in Tab.~\ref{table:augmentationalpha}.
While alpha-blending introduces larger deviations from the initial optical flow for all methods, the ranking across colors is the same for both color-blending methods.

\begin{figure}
\centering
\setlength{\fboxrule}{0.1pt}%
\setlength{\fboxsep}{0pt}%
\begin{tabular}{@{}M{41mm}@{\ }M{41mm}@{}}
    \darkleftrightboxb{41mm}{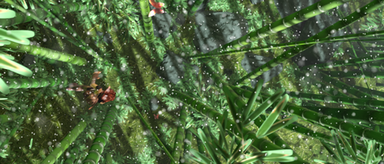}{small}{Size} &
    \darkleftboxb{41mm}{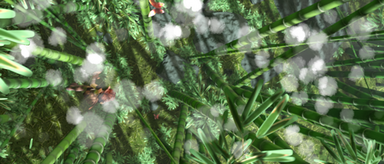}{medium}
    \\
    \darkleftboxb{41mm}{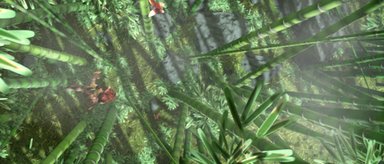}{large} &
    \darkleftboxb{41mm}{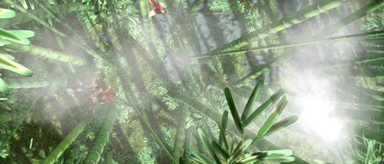}{fog}
    \\
\end{tabular}
\vspace{-0.5mm}
\caption{Augmentations for different \emph{particle sizes} and \emph{transparencies} from Main Tab.~\ref{table:augmentations} on an exemplary Sintel~\cite{Butler2012NaturalisticOpenSource} frame.
Augmentations for \emph{particle count}, \emph{motion blur} and \emph{color} are shown in Fig.~\ref{fig:augmentations_numraincol}, augmentation parameters are listed in Tab.~\ref{table:configsaug}.}
\label{fig:augmentationssize}
\vspace{0.5mm}
\end{figure}

\begin{figure*}
\centering
\setlength{\fboxrule}{0.1pt}%
\setlength{\fboxsep}{0pt}%
\begin{tabular}{@{}M{43mm}@{\ }M{43mm}@{\ }M{43mm}@{\ }M{43mm}@{\ }@{}}
    \darkleftrightboxb{43mm}{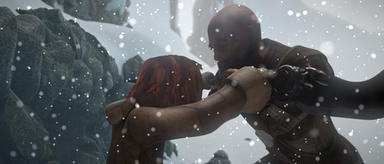}{1000}{Particle count} &
    \darkleftrightboxb{43mm}{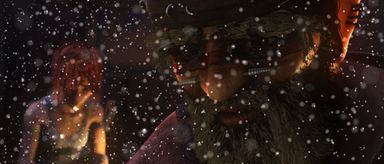}{0.0}{Motion blur} &
    \darkleftrightboxb{43mm}{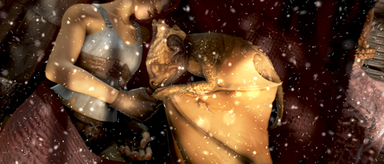}{white}{Additive blending} &
    \darkleftrightboxb{43mm}{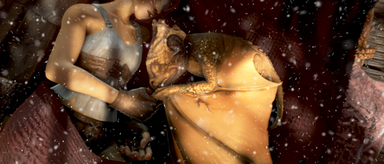}{white}{Alpha blending}
    \\[-2.5pt]
    \darkleftboxb{43mm}{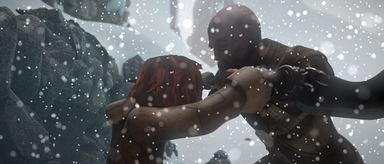}{2000} &
    \darkleftboxb{43mm}{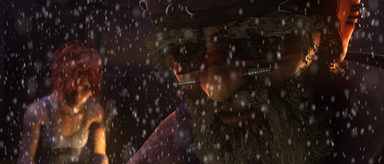}{0.0375} &
    \darkleftboxb{43mm}{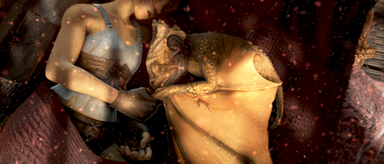}{red} &
    \darkleftboxb{43mm}{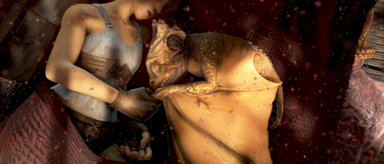}{red}
    \\[-2.5pt]
    \darkleftboxb{43mm}{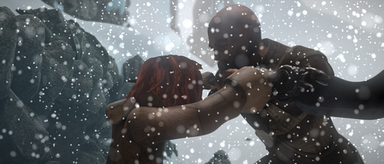}{3000} &
    \darkleftboxb{43mm}{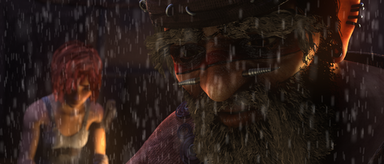}{0.75} &
    \darkleftboxb{43mm}{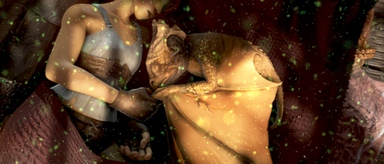}{green} &
    \darkleftboxb{43mm}{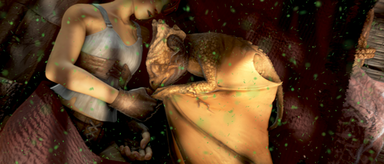}{green}
    \\[-2.5pt]
    \darkleftboxb{43mm}{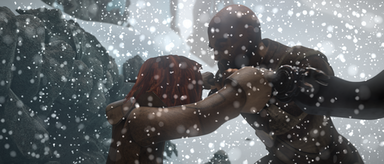}{4000} &
    \darkleftboxb{43mm}{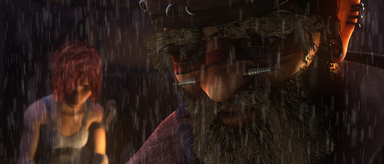}{0.1125} &
    \darkleftboxb{43mm}{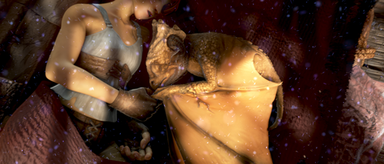}{blue} &
    \darkleftboxb{43mm}{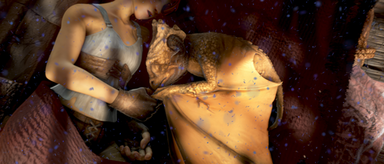}{blue}
    \\[-2.5pt]
    \darkleftboxb{43mm}{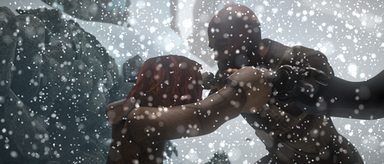}{5000} &
    \darkleftboxb{43mm}{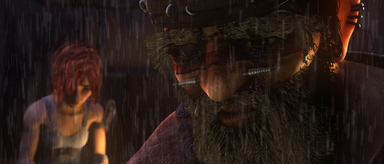}{0.15} &
    \darkleftboxb{43mm}{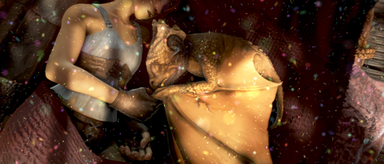}{color} &
    \darkleftboxb{43mm}{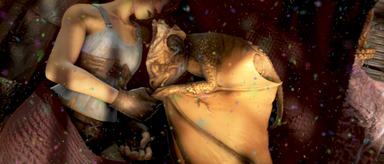}{color}
    \\
\end{tabular}
\caption{Augmentations for different \emph{particle count}, \emph{motion blur} and \emph{color} from Main Tab.~\ref{table:augmentations} and Tab.~\ref{table:augmentationalpha} on exemplary Sintel~\cite{Butler2012NaturalisticOpenSource} frames.
Augmentations for \emph{particle sizes} are shown in Fig.~\ref{fig:augmentationssize}, augmentation parameters are listed in Tab.~\ref{table:configsaug}.}
\label{fig:augmentations_numraincol}
\end{figure*}

In Tab.~\ref{table:configsaug} we give a full list of parameter configurations for the particle effects from Main Tab.~\ref{table:augmentations} and Tab.~\ref{table:augmentationalpha}.
In addition to the weather visualizations in Main Fig.~4, we visualize all \emph{Size}-variations for particles in Fig.~\ref{fig:augmentationssize}, and all \emph{Particle count}, \emph{Motion blur} and \emph{color} variations in Fig.~\ref{fig:augmentations_numraincol}.
From these figures it becomes clear that the configurations \emph{size: small}, \emph{motion blur: 0.0} and \emph{color: white} all visually correspond to snow.
Therefore, they were not chosen as four main weather effects.
Instead, we selected configurations that lead to more diverse visual appearance, even though these configurations were not necessarily the most effective ones to perturb the optical flow output in Main Tab.~\ref{table:augmentations}.

\begin{table}
\small
\begin{center}
\begin{tabular}{@{\ }l@{\ }l@{\ \ }lrrrrrr@{\ }}
\toprule
\multicolumn{3}{l}{\rotatebox{0}{Weather}} & \multicolumn{1}{c}{\rotatebox{60}{FN2}} & \multicolumn{1}{c}{\rotatebox{60}{FNCR}} & \multicolumn{1}{c}{\rotatebox{60}{SpyNet}} & \multicolumn{1}{c}{\rotatebox{60}{RAFT}} & \multicolumn{1}{c}{\rotatebox{60}{GMA}} & \multicolumn{1}{c}{\rotatebox{60}{FF}} \\
\midrule
\multirow{10}{*}{\rotatebox{90}{Color}} & \multirow{5}{*}{\rotatebox{90}{$\alpha$-Blending}} &
white      & \textbf{10.03} & \textbf{10.70} & \textbf{4.95} & \textbf{3.88} &\textbf{3.74} & \textbf{2.93} \\
& &
red        &          9.41  &          9.03  &         3.14  &         2.72  &        2.56  &         2.74  \\
& &
green      &          6.81  &          8.46  &         2.84  &         2.44  &        2.35  &         2.11  \\
& &
blue       &          6.32  &          8.18  &         2.67  &         2.69  &        2.76  &         1.85  \\
& &
color      &          8.20  &          8.14  &         3.22  &         2.91  &        3.17  &         2.39  \\
\cmidrule(l{0.5em}r{0.5em}){2-9}
& \multirow{5}{*}{\rotatebox{90}{Additive}} &
white      & \textbf{14.05} & \textbf{14.68} & \textbf{6.47} & \textbf{5.49} & \textbf{4.63} & \textbf{4.68} \\
& &
red        &         12.57  &         12.07  &         4.21  &         3.74  &         2.95  &         3.03  \\
& &
green      &          9.02  &          9.64  &         3.68  &         3.16  &         2.76  &         2.56  \\
& &
blue       &          7.84  &         10.47  &         3.37  &         3.52  &         3.31  &         2.30  \\
& &
color      &         11.17  &         11.50  &         4.36  &         4.11  &         3.75  &         3.18  \\
\bottomrule
\end{tabular}
\vspace{1.5mm}
\caption{\emph{Robustness} AEE$(\flow,\fadv)$ $\downarrow$~\cite{Schmalfuss2022PerturbationConstrainedAdversarial} of particle-based weather augmentations for optical \emph{flow methods} when varying the color rendering (additive / $\alpha$-blending) to supplement Main Tab.~\ref{table:augmentations}. Illustration provided in Fig.~\ref{fig:augmentations_numraincol}. Worst robustness is bold.}
\label{table:augmentationalpha}
\end{center}
\vspace{-1mm}
\end{table}

\begin{table*}
\small
\begin{center}
\scalebox{0.8}{%
\begin{tabular}{@{\ }l@{\ }l@{\ \ }lrrlcccclrcrclc}
\toprule
& & & \multicolumn{4}{c}{Particle base properties} & \multicolumn{4}{c}{Color properties} & \multicolumn{3}{c}{Motion properties} & \multicolumn{3}{c}{Motion blur} \\
\cmidrule(l{0.5em}r{0.5em}){4-7} \cmidrule(l{0.5em}r{0.5em}){8-11} \cmidrule(l{0.5em}r{0.5em}){12-14} \cmidrule(l{0.5em}r{0.5em}){15-17}
\multicolumn{3}{l}{\rotatebox{0}{Weather}} & \multicolumn{1}{c}{\rotatebox{0}{Count}} & \multicolumn{1}{c}{\rotatebox{0}{Size}} & \multicolumn{1}{c}{\rotatebox{0}{$d$-decay}} & \multicolumn{1}{c}{\rotatebox{0}{Templates}} & \multicolumn{1}{c}{\rotatebox{0}{(R,G,B)}} & \multicolumn{1}{c}{\rotatebox{0}{($\delta$H, $\delta$L, $\delta$S)}} & \multicolumn{1}{c}{\rotatebox{0}{Mode}} & \multicolumn{1}{c}{\rotatebox{0}{$\transp$}} & \multicolumn{1}{c}{\rotatebox{0}{$m_y$}} & \multicolumn{1}{c}{\rotatebox{0}{$\delta \angle m$}} & \multicolumn{1}{c}{\rotatebox{0}{$\delta \|m\|$}} & \multicolumn{1}{c}{\rotatebox{0}{Blur}} & \multicolumn{1}{c}{\rotatebox{0}{Length}} & \multicolumn{1}{c}{\rotatebox{0}{Particles}} \\
\midrule
\multirow{5}{*}{\rotatebox{90}{Particles}} & &
1000                 & 1000 &  71 &  9 & particles     &       (255,255,255) & (\phantom{00}0, 0.0, 0.0) &    additive & 0.75 &   0.2  &  0.0  &   0 &  \no  & 0.0    &  0 \\
& &
2000                 & 2000 &  71 &  9 & particles     &       (255,255,255) & (\phantom{00}0, 0.0, 0.0) &    additive & 0.75 &   0.2  &  0.0  &   0 &  \no  & 0.0    &  0 \\
& &
\cellcolor{gray!20}3000  (snow)         & \cellcolor{gray!20}3000 & \cellcolor{gray!20}71 & \cellcolor{gray!20}9 & \cellcolor{gray!20}particles     & \cellcolor{gray!20}(255,255,255) & \cellcolor{gray!20}(\phantom{00}0, 0.0, 0.0) & \cellcolor{gray!20}additive & \cellcolor{gray!20}0.75 & \cellcolor{gray!20}0.2  & \cellcolor{gray!20}0.0  & \cellcolor{gray!20}0 & \cellcolor{gray!20}\no  & \cellcolor{gray!20}0.0    & \cellcolor{gray!20}0 \\
& &
4000                 & 4000 &  71 &  9 & particles     &       (255,255,255) & (\phantom{00}0, 0.0, 0.0) &    additive & 0.75 &   0.2  &  0.0  &   0 &  \no  & 0.0    &  0 \\
& &
5000                 & 5000 &  71 &  9 & particles     &       (255,255,255) & (\phantom{00}0, 0.0, 0.0) &    additive & 0.75 &   0.2  &  0.0  &   0 &  \no  & 0.0    &  0 \\
\midrule
& &
\cellcolor{gray!20}grey                 & \cellcolor{gray!20}3000 & \cellcolor{gray!20}71 & \cellcolor{gray!20}9 & \cellcolor{gray!20}particles     & \cellcolor{gray!20}(127,127,127) & \cellcolor{gray!20}(\phantom{00}0, 0.0, 0.0) & \cellcolor{gray!20}Meshkin & \cellcolor{gray!20}0.75 & \cellcolor{gray!20}0.2  & \cellcolor{gray!20}0.0  & \cellcolor{gray!20}0 & \cellcolor{gray!20}\no  & \cellcolor{gray!20}0.0    & \cellcolor{gray!20}0 \\
\midrule
\multirow{5}{*}{\rotatebox{90}{Motion blur}} & &
0.0\phantom{000}     & 3000 &  71 &  9 & particles     &       (255,255,255) & (\phantom{00}0, 0.0, 0.0) &    additive & 0.75 &   0.2  &  0.0  &   0 &  \no  & 0.0    &  0 \\
& &
0.0375               & 3000 &  67 &  9 & particles     &       (255,255,255) & (\phantom{00}0, 0.0, 0.0) &    additive & 0.75 &   0.2  &  0.1  &   4 &  \yes & 0.0375 & 20 \\
& &
0.075\phantom{0}     & 3000 &  61 &  9 & particles     &       (255,255,255) & (\phantom{00}0, 0.0, 0.0) &    additive & 0.75 &   0.2  &  0.1  &   4 &  \yes & 0.075  & 20 \\
& &
0.1125               & 3000 &  57 &  9 & particles     &       (255,255,255) & (\phantom{00}0, 0.0, 0.0) &    additive & 0.75 &   0.2  &  0.1  &   4 &  \yes & 0.1125 & 20 \\
& &
\cellcolor{gray!20}0.15  (rain)         & \cellcolor{gray!20}3000 & \cellcolor{gray!20}51 & \cellcolor{gray!20}9 & \cellcolor{gray!20}particles     & \cellcolor{gray!20}(255,255,255) & \cellcolor{gray!20}(\phantom{00}0, 0.0, 0.0) & \cellcolor{gray!20}additive & \cellcolor{gray!20}0.75 & \cellcolor{gray!20}0.2  & \cellcolor{gray!20}0.1  & \cellcolor{gray!20}4 & \cellcolor{gray!20}\yes & \cellcolor{gray!20}0.15   & \cellcolor{gray!20}20 \\
\midrule
\multirow{10}{*}{\rotatebox{90}{Color}} & \multirow{5}{*}{\rotatebox{90}{Additive}} &
white                & 3000 &  41 &  9 & particles     &       (255,255,255) & (\phantom{00}0, 0.0, 0.0) &    additive & 1.5  &  -0.05 &  0.2  &  60 &  \yes & 0.3    & 10 \\
& &
\cellcolor{gray!20}red (sparks)         & \cellcolor{gray!20}3000 & \cellcolor{gray!20}41 & \cellcolor{gray!20}9 & \cellcolor{gray!20}particles     & \cellcolor{gray!20}(191,\phantom{0}79,\phantom{0}64) & \cellcolor{gray!20}(\phantom{0}15, 0.1, 0.1) & \cellcolor{gray!20}additive & \cellcolor{gray!20}1.5  & \cellcolor{gray!20}-0.05 & \cellcolor{gray!20}0.2  & \cellcolor{gray!20}60 & \cellcolor{gray!20}\yes & \cellcolor{gray!20}0.3    & \cellcolor{gray!20}10 \\
& &
green                & 3000 &  41 &  9 & particles     &       (\phantom{0}79,191,\phantom{0}64) & (\phantom{0}15, 0.1, 0.1) &    additive & 1.5  &  -0.05 &  0.2  &  60 &  \yes & 0.3    & 10 \\
& &
blue                 & 3000 &  41 &  9 & particles     &       (\phantom{0}64,\phantom{0}79,191)  & (\phantom{0}15, 0.1, 0.1) &    additive & 1.5  &  -0.05 &  0.2  &  60 &  \yes & 0.3    & 10 \\
& &
color                & 3000 &  41 &  9 & particles     &       (205,\phantom{0}80,\phantom{0}80) &           (180, 0.1, 0.1) &    additive & 1.5  &  -0.05 &  0.2  &  60 &  \yes & 0.3    & 10 \\
\cmidrule(l{0.5em}r{0.5em}){2-17}
 & \multirow{5}{*}{\rotatebox{90}{$\alpha$-Blending}} &
white                & 3000 &  41 &  9 & particles     &       (255,255,255) & (\phantom{00}0, 0.0, 0.0) &     Meshkin & 1.5  &  -0.05 &  0.2  &  60 &  \yes & 0.3    & 10 \\
& &
red                  & 3000 &  41 &  9 & particles     &       (191,\phantom{0}79,\phantom{0}64) & (\phantom{0}15, 0.1, 0.1) &     Meshkin & 1.5  &  -0.05 &  0.2  &  60 &  \yes & 0.3    & 10 \\
& &
green                & 3000 &  41 &  9 & particles     &       (\phantom{0}79,191,\phantom{0}64) & (\phantom{0}15, 0.1, 0.1) &     Meshkin & 1.5  &  -0.05 &  0.2  &  60 &  \yes & 0.3    & 10 \\
& &
blue                 & 3000 &  41 &  9 & particles     &       (\phantom{0}64,\phantom{0}79,191) & (\phantom{0}15, 0.1, 0.1) &     Meshkin & 1.5  &  -0.05 &  0.2  &  60 &  \yes & 0.3    & 10 \\
& &
color                & 3000 &  41 &  9 & particles     &       (205,\phantom{0}80,\phantom{0}80) &           (180, 0.1, 0.1) &     Meshkin & 1.5  &  -0.05 &  0.2  &  60 &  \yes & 0.3    & 10 \\
\midrule
\multirow{4}{*}{\rotatebox{90}{Size}} & &
small                & 3000 &  71 &  9 & dust          &       (255,255,255) & (\phantom{00}0, 0.0, 0.0) &     Meshkin & 1.0  &   0.0  &  0.0  &   0 &  \no  & 0.0    &  0 \\
& &
medium               &  141 & 161 &1.75& dust          &       (255,255,255) & (\phantom{00}0, 0.0, 0.0) &     Meshkin & 0.78 &   0.0  &  0.0  &   0 &  \no  & 0.0    &  0 \\
& &
large                &   60 & 451 &0.8 & dust          &       (255,255,255) & (\phantom{00}0, 0.0, 0.0) &     Meshkin & 0.1  &   0.0  &  0.0  &   0 &  \no  & 0.0    &  0 \\
& &
\cellcolor{gray!20}fog  (fog)           & \cellcolor{gray!20}60 & \cellcolor{gray!20}451 & \cellcolor{gray!20}0.8 & \cellcolor{gray!20}dust          & \cellcolor{gray!20}(255,255,255) & \cellcolor{gray!20}(\phantom{00}0, 0.0, 0.0) & \cellcolor{gray!20}Meshkin & \cellcolor{gray!20}0.25 & \cellcolor{gray!20}0.0  & \cellcolor{gray!20}0.0  & \cellcolor{gray!20} 0 & \cellcolor{gray!20}\no  & \cellcolor{gray!20}0.0    & \cellcolor{gray!20}0 \\
\bottomrule
\end{tabular}%
}
\vspace{2.5mm}
\caption{\emph{Particle configurations} for Sintel train dataset augmentations from Main Tab.~\ref{table:augmentations} and Tab.~\ref{table:augmentationalpha}. It additionally lists the \emph{grey} particle configuration used in Main Tab.~\ref{table:optimvariables}. $d$-Decay is a depth-decay parameter that affects the size, $\delta$H, $\delta$L and $\delta$S are random color variations in the HLS space around the initial RGB color configurations. While all effects use a depth-dependent transparency scaling, fog has a depth-constant transparency of 0.3. The motion $m$ is always along the $y$-axis (vertically, $m_x=m_z=0$), and may vary by a random angle $\delta \angle m$ or be scaled by a random factor that scales with a fraction of $\|m\|$. All configurations were created with 8 GPUs and a random seed of 0. To train RAFT on another snow dataset than the test set (Main Tab.~\ref{table:raft_ckpt}), the training set uses a random seed of 1234.}
\label{table:configsaug}
\vspace{-2.5mm}
\end{center}
\end{table*}

\subsubsection{Robustness on real-world KITTI data}

In addition to the KITTI samples from Main Fig.~\ref{fig:KittiSpring}, we evaluate the robustness values AEE$(f,\check{f})$ of all methods on randomly augmented real-world data from KITTI train in Tab.~\ref{tab:KITTIaugmentation}.
Compared to the Sintel augmentations in Main Tab.~\ref{table:augmentations}, snow, rain and sparks based on additive color-rendering behave similarly (i.e.\ snow is the most effective, sparks and rain have comparable strength).
Therefore, the results in the main paper on Sintel data can largely be transferred to real-world scenarios.

Due to the lighter colors in KITTI scenes, which are often captured in bright/sunny conditions, alpha-blending is slightly less effective as further brightening causes less color change.
However, real situations with snow or rain in bright sunshine are not very common.
In contrast, fog has a larger effectiveness, as it is based on additive-blending and additionally obfuscates more objects because KITTI has fewer foreground objects / more scene depth than Sintel.

\begin{table}
{
\small
\begin{center}
\begin{tabular}{lrrrrrr}
\toprule
Augment.  & \multicolumn{1}{c}{\rotatebox{60}{FN2}} & \multicolumn{1}{c}{\rotatebox{60}{FNCR}} & \multicolumn{1}{c}{\rotatebox{60}{SpyNet}} & \multicolumn{1}{c}{\rotatebox{60}{RAFT}} & \multicolumn{1}{c}{\rotatebox{60}{GMA}} & \multicolumn{1}{c}{\rotatebox{60}{FF}} \\
\midrule
snow                &   \textbf{8.32} &   \textbf{7.71} &           5.49  &    \textbf{3.08} &      \textbf{3.27} &               4.21  \\
rain                &           3.70  &           4.99  &           3.59  &            1.51  &              1.91  &               2.67  \\
sparks              &           4.16  &           4.28  &           3.15  &            1.43  &              1.91  &               2.73  \\
fog                 &           7.30  &           7.20  &  \textbf{11.48} &            2.98  &              3.25  &       \textbf{4.83} \\
\bottomrule
\end{tabular}
\end{center}
}
\caption{Robustness AEE$(f,\check{f})$~\cite{Schmalfuss2022PerturbationConstrainedAdversarial} of random particle-based weather augmentations for \emph{optical flow methods} on KITTI train~\cite{Menze2015ObjectSceneFlow}, worst robustness is bold. The table above provides additional results that correspond to the highlighted weather augmentations in Main Tab.~\ref{table:augmentations} on the Sintel dataset.}
\label{tab:KITTIaugmentation}
\end{table}

\subsection{Adversarial weather attacks}

\subsubsection{Attack Configurations}

With the provided code and the network implementations above, Tab.~\ref{table:configattacks} lists the configurations for all weather attacks that were used to create Tables~\ref{table:optimvariables},~\ref{table:weatherattacks} and~\ref{table:of_attacks} from the Main paper.
To compare to PCFA~\cite{Schmalfuss2022PerturbationConstrainedAdversarial} and I-FGSM~\cite{Schrodi2022TowardsUnderstandingAdversarial}, we use the implementation from ~\cite{Schmalfuss2022PerturbationConstrainedAdversarial} and the author-provided configurations for PCFA with $\varepsilon_2={}$5e-3, AEE loss, COV constraint and disjoint, non-universal perturbations.
For I-FGSM we use a perturbation bound of $\varepsilon_\infty={}$5e-3 and 25 optimization steps.
Both attacks are run on Sintel train.

\begin{table}
\small
\begin{center}
\scalebox{0.9}{%
\begin{tabular}{cccccccccc}
\toprule
Tab. & \multicolumn{1}{c}{\rotatebox{0}{Dataset}} & \multicolumn{1}{c}{\rotatebox{0}{Augment.}} & \multicolumn{1}{c}{\rotatebox{0}{LR}} & \multicolumn{1}{c}{\rotatebox{0}{$\de$}} & \multicolumn{1}{c}{\rotatebox{0}{$\dt$}} & \multicolumn{1}{c}{\rotatebox{0}{$\dcol$}} & \multicolumn{1}{c}{\rotatebox{0}{$\dtransp$}} \\
\midrule
\multirow{7}{*}{\rotatebox{90}{Table~\ref{table:optimvariables}}}
& Sintel-tr115 &     grey &  1e-5 & \yes & \no  & \no  & \no  \\
& Sintel-tr115 &     grey &  1e-5 & \no  & \yes & \no  & \no  \\
& Sintel-tr115 &     grey &  1e-3 & \no  & \no  & \yes & \no  \\
& Sintel-tr115 &     grey &  1e-3 & \no  & \no  & \no  & \yes \\
& Sintel-tr115 &     grey &  1e-5 & \yes & \yes & \no  & \no  \\
& Sintel-tr115 &     grey &  1e-3 & \no  & \no  & \yes & \yes \\
& Sintel-tr115 &     grey &  1e-5 & \yes & \yes & \yes & \yes \\
\midrule
\multirow{4}{*}{\rotatebox{90}{Table~\ref{table:weatherattacks}}}
& Sintel-tr115 &     snow &  1e-5 & \yes & \yes & \yes & \yes \\
& Sintel-tr115 &     rain &  1e-5 & \yes & \yes & \yes & \yes \\
& Sintel-tr115 &   sparks &  1e-5 & \yes & \yes & \yes & \yes \\
& Sintel-tr115 &      fog &  1e-5 & \yes & \no  & \yes & \yes \\
\midrule
\multirow{1}{*}{\rotatebox{90}{T.\ref{table:of_attacks}}}
& Sintel train &     snow &  1e-5 & \yes & \yes & \yes & \yes \\
\bottomrule
\end{tabular}%
}
\end{center}
\vspace{-0.5mm}
\caption{\emph{Weather attack configurations} for the experiments from Main Tables~\ref{table:optimvariables},~\ref{table:weatherattacks} and~\ref{table:of_attacks}. \emph{Augment} specifies the augmentation, \cf Tab.~\ref{table:configsaug}, \emph{LR} denotes the optimizer learning rate and the optimization variables $\de, \dt, \dcol$ and $\dtransp$ indicate which of them were optimized. Optimization with 750 steps of Adam using weights $\alpha_1 = \alpha_2 = 1000$ for the loss function.}
\label{table:configattacks}
\end{table}

\subsubsection{Additional configurations for training with snow}

Regarding the configurations for the snow-augmented training in Sec.~\ref{sec:TrainingWithWeather}, Main Tab.~\ref{table:raft_ckpt} uses snow, rain, sparks and fog augmentations as specified in Tab.~\ref{table:configsaug} and the respective attack configurations that were used for Main Tab.~\ref{table:weatherattacks}, which are listed in Tab.~\ref{table:configattacks}.

\subsubsection{Additional visualizations for weather attacks}

\begin{figure*}
\centering
\setlength{\fboxrule}{0.1pt}%
\setlength{\fboxsep}{0pt}%
\begin{tabular}{@{}M{19.1mm}@{}M{19.1mm}@{}M{19.1mm}@{\ }M{19.1mm}@{}M{19.1mm}@{}M{19.1mm}@{\ }M{19.1mm}@{}M{19.1mm}@{}M{19.1mm}@{}}
    \multicolumn{3}{c}{FlowNet2} & \multicolumn{3}{c}{FlowNetCRobust} & \multicolumn{3}{c}{SpyNet}
    \\
    \darkleftbox{19.1mm}{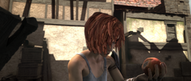}{\scriptsize Original} &
    \includegraphics[width=19.1mm]{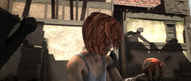} &
    \fcolorbox{gray!50}{white}{\includegraphics[width=19.1mm]{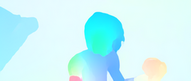}} &
    \darkleftbox{19.1mm}{attacks_half/alley_1/frame_0005.png}{\scriptsize Original} &
    \includegraphics[width=19.1mm]{attacks_half/alley_1/frame_0006.png} &
    \fcolorbox{gray!50}{white}{\includegraphics[width=19.1mm]{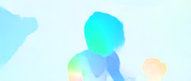}} &
    \darkleftbox{19.1mm}{attacks_half/alley_1/frame_0005.png}{\scriptsize Original} &
    \includegraphics[width=19.1mm]{attacks_half/alley_1/frame_0006.png} &
    \fcolorbox{gray!50}{white}{\includegraphics[width=19.1mm]{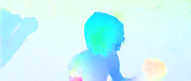}}
    \\[-3.5pt]
    \darkleftbox{19.1mm}{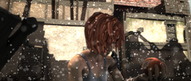}{\scriptsize Random} &
    \includegraphics[width=19.1mm]{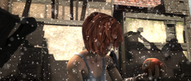} &
    \fcolorbox{gray!50}{white}{\includegraphics[width=19.1mm]{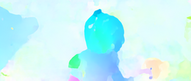}} &
    \darkleftbox{19.1mm}{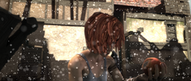}{\scriptsize Random} &
    \includegraphics[width=19.1mm]{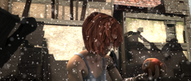} &
    \fcolorbox{gray!50}{white}{\includegraphics[width=19.1mm]{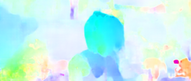}} &
    \darkleftbox{19.1mm}{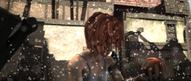}{\scriptsize Random} &
    \includegraphics[width=19.1mm]{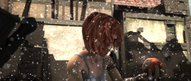} &
    \fcolorbox{gray!50}{white}{\includegraphics[width=19.1mm]{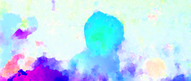}}
    \\[-3.5pt]
    \darkleftbox{19.1mm}{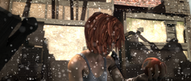}{\scriptsize Adversarial} &
    \includegraphics[width=19.1mm]{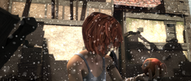} &
    \fcolorbox{gray!50}{white}{\includegraphics[width=19.1mm]{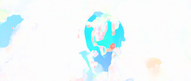}} &
    \darkleftbox{19.1mm}{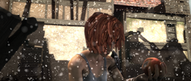}{\scriptsize Adversarial} &
    \includegraphics[width=19.1mm]{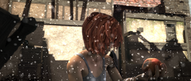} &
    \fcolorbox{gray!50}{white}{\includegraphics[width=19.1mm]{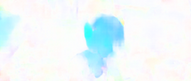}} &
    \darkleftbox{19.1mm}{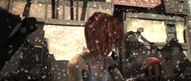}{\scriptsize Adversarial} &
    \includegraphics[width=19.1mm]{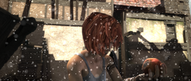} &
    \fcolorbox{gray!50}{white}{\includegraphics[width=19.1mm]{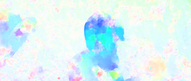}}
    \\
    \darkleftbox{19.1mm}{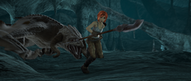}{\scriptsize Original} &
    \includegraphics[width=19.1mm]{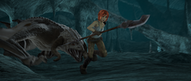} &
    \fcolorbox{gray!50}{white}{\includegraphics[width=19.1mm]{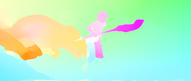}} &
    \darkleftbox{19.1mm}{attacks_half/cave_2/frame_0005.png}{\scriptsize Original} &
    \includegraphics[width=19.1mm]{attacks_half/cave_2/frame_0006.png} &
    \fcolorbox{gray!50}{white}{\includegraphics[width=19.1mm]{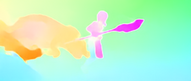}} &
    \darkleftbox{19.1mm}{attacks_half/cave_2/frame_0005.png}{\scriptsize Original} &
    \includegraphics[width=19.1mm]{attacks_half/cave_2/frame_0006.png} &
    \fcolorbox{gray!50}{white}{\includegraphics[width=19.1mm]{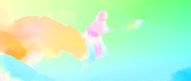}}
    \\[-3.5pt]
    \darkleftbox{19.1mm}{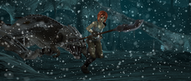}{\scriptsize Random} &
    \includegraphics[width=19.1mm]{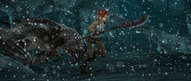} &
    \fcolorbox{gray!50}{white}{\includegraphics[width=19.1mm]{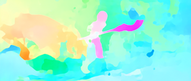}} &
    \darkleftbox{19.1mm}{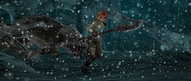}{\scriptsize Random} &
    \includegraphics[width=19.1mm]{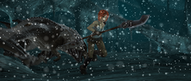} &
    \fcolorbox{gray!50}{white}{\includegraphics[width=19.1mm]{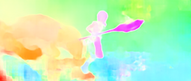}} &
    \darkleftbox{19.1mm}{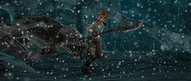}{\scriptsize Random} &
    \includegraphics[width=19.1mm]{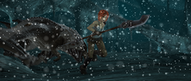} &
    \fcolorbox{gray!50}{white}{\includegraphics[width=19.1mm]{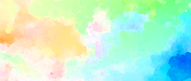}}
    \\[-3.5pt]
    \darkleftbox{19.1mm}{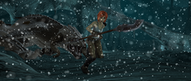}{\scriptsize Adversarial} &
    \includegraphics[width=19.1mm]{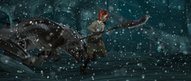} &
    \fcolorbox{gray!50}{white}{\includegraphics[width=19.1mm]{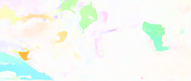}} &
    \darkleftbox{19.1mm}{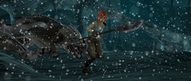}{\scriptsize Adversarial} &
    \includegraphics[width=19.1mm]{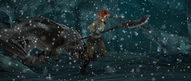} &
    \fcolorbox{gray!50}{white}{\includegraphics[width=19.1mm]{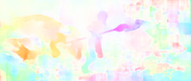}} &
    \darkleftbox{19.1mm}{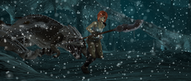}{\scriptsize Adversarial} &
    \includegraphics[width=19.1mm]{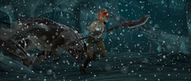} &
    \fcolorbox{gray!50}{white}{\includegraphics[width=19.1mm]{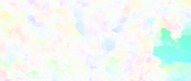}}
    \\
    \darkleftbox{19.1mm}{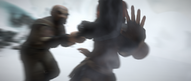}{\scriptsize Original} &
    \includegraphics[width=19.1mm]{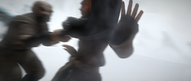} &
    \fcolorbox{gray!50}{white}{\includegraphics[width=19.1mm]{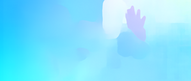}} &
    \darkleftbox{19.1mm}{attacks_half/ambush_2/frame_0003.png}{\scriptsize Original} &
    \includegraphics[width=19.1mm]{attacks_half/ambush_2/frame_0004.png} &
    \fcolorbox{gray!50}{white}{\includegraphics[width=19.1mm]{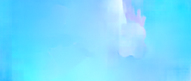}} &
    \darkleftbox{19.1mm}{attacks_half/ambush_2/frame_0003.png}{\scriptsize Original} &
    \includegraphics[width=19.1mm]{attacks_half/ambush_2/frame_0004.png} &
    \fcolorbox{gray!50}{white}{\includegraphics[width=19.1mm]{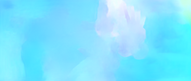}}
    \\[-3.5pt]
    \darkleftbox{19.1mm}{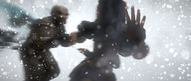}{\scriptsize Random} &
    \includegraphics[width=19.1mm]{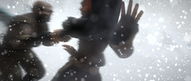} &
    \fcolorbox{gray!50}{white}{\includegraphics[width=19.1mm]{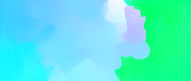}} &
    \darkleftbox{19.1mm}{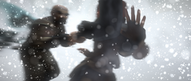}{\scriptsize Random} &
    \includegraphics[width=19.1mm]{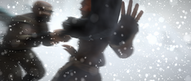} &
    \fcolorbox{gray!50}{white}{\includegraphics[width=19.1mm]{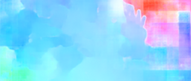}} &
    \darkleftbox{19.1mm}{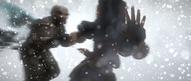}{\scriptsize Random} &
    \includegraphics[width=19.1mm]{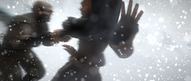} &
    \fcolorbox{gray!50}{white}{\includegraphics[width=19.1mm]{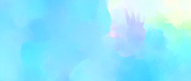}}
    \\[-3.5pt]
    \darkleftbox{19.1mm}{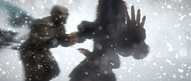}{\scriptsize Adversarial} &
    \includegraphics[width=19.1mm]{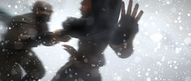} &
    \fcolorbox{gray!50}{white}{\includegraphics[width=19.1mm]{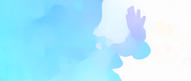}} &
    \darkleftbox{19.1mm}{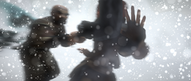}{\scriptsize Adversarial} &
    \includegraphics[width=19.1mm]{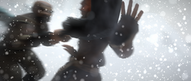} &
    \fcolorbox{gray!50}{white}{\includegraphics[width=19.1mm]{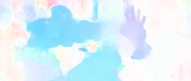}} &
    \darkleftbox{19.1mm}{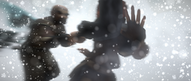}{\scriptsize Adversarial} &
    \includegraphics[width=19.1mm]{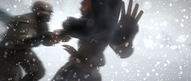} &
    \fcolorbox{gray!50}{white}{\includegraphics[width=19.1mm]{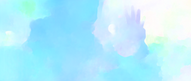}}
    \\
    \multicolumn{3}{c}{RAFT} & \multicolumn{3}{c}{GMA} & \multicolumn{3}{c}{FlowFormer}
    \\
    \darkleftbox{19.1mm}{attacks_half/alley_1/frame_0005.png}{\scriptsize Original} &
    \includegraphics[width=19.1mm]{attacks_half/alley_1/frame_0006.png} &
    \fcolorbox{gray!50}{white}{\includegraphics[width=19.1mm]{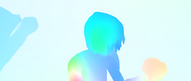}} &
    \darkleftbox{19.1mm}{attacks_half/alley_1/frame_0005.png}{\scriptsize Original} &
    \includegraphics[width=19.1mm]{attacks_half/alley_1/frame_0006.png} &
    \fcolorbox{gray!50}{white}{\includegraphics[width=19.1mm]{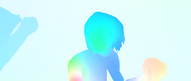}} &
    \darkleftbox{19.1mm}{attacks_half/alley_1/frame_0005.png}{\scriptsize Original} &
    \includegraphics[width=19.1mm]{attacks_half/alley_1/frame_0006.png} &
    \fcolorbox{gray!50}{white}{\includegraphics[width=19.1mm]{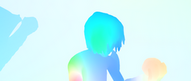}}
    \\[-3.5pt]
    \darkleftbox{19.1mm}{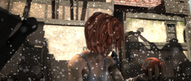}{\scriptsize Random} &
    \includegraphics[width=19.1mm]{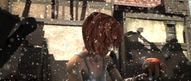} &
    \fcolorbox{gray!50}{white}{\includegraphics[width=19.1mm]{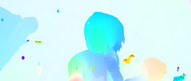}} &
    \darkleftbox{19.1mm}{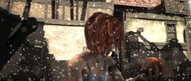}{\scriptsize Random} &
    \includegraphics[width=19.1mm]{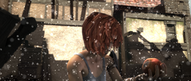} &
    \fcolorbox{gray!50}{white}{\includegraphics[width=19.1mm]{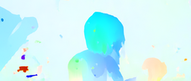}} &
    \darkleftbox{19.1mm}{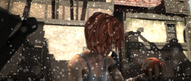}{\scriptsize Random} &
    \includegraphics[width=19.1mm]{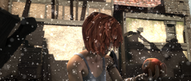} &
    \fcolorbox{gray!50}{white}{\includegraphics[width=19.1mm]{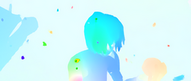}}
    \\[-3.5pt]
    \darkleftbox{19.1mm}{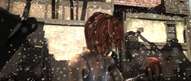}{\scriptsize Adversarial} &
    \includegraphics[width=19.1mm]{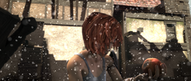} &
    \fcolorbox{gray!50}{white}{\includegraphics[width=19.1mm]{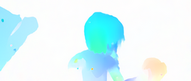}} &
    \darkleftbox{19.1mm}{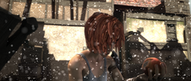}{\scriptsize Adversarial} &
    \includegraphics[width=19.1mm]{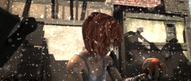} &
    \fcolorbox{gray!50}{white}{\includegraphics[width=19.1mm]{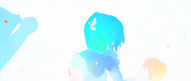}} &
    \darkleftbox{19.1mm}{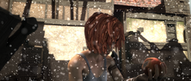}{\scriptsize Adversarial} &
    \includegraphics[width=19.1mm]{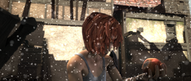} &
    \fcolorbox{gray!50}{white}{\includegraphics[width=19.1mm]{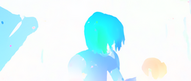}}
    \\
    \darkleftbox{19.1mm}{attacks_half/cave_2/frame_0005.png}{\scriptsize Original} &
    \includegraphics[width=19.1mm]{attacks_half/cave_2/frame_0006.png} &
    \fcolorbox{gray!50}{white}{\includegraphics[width=19.1mm]{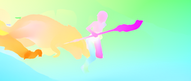}} &
    \darkleftbox{19.1mm}{attacks_half/cave_2/frame_0005.png}{\scriptsize Original} &
    \includegraphics[width=19.1mm]{attacks_half/cave_2/frame_0006.png} &
    \fcolorbox{gray!50}{white}{\includegraphics[width=19.1mm]{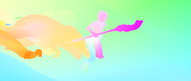}} &
    \darkleftbox{19.1mm}{attacks_half/cave_2/frame_0005.png}{\scriptsize Original} &
    \includegraphics[width=19.1mm]{attacks_half/cave_2/frame_0006.png} &
    \fcolorbox{gray!50}{white}{\includegraphics[width=19.1mm]{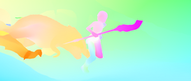}}
    \\[-3.5pt]
    \darkleftbox{19.1mm}{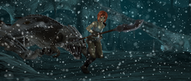}{\scriptsize Random} &
    \includegraphics[width=19.1mm]{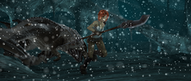} &
    \fcolorbox{gray!50}{white}{\includegraphics[width=19.1mm]{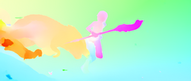}} &
    \darkleftbox{19.1mm}{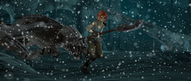}{\scriptsize Random} &
    \includegraphics[width=19.1mm]{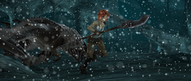} &
    \fcolorbox{gray!50}{white}{\includegraphics[width=19.1mm]{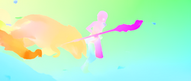}} &
    \darkleftbox{19.1mm}{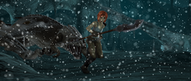}{\scriptsize Random} &
    \includegraphics[width=19.1mm]{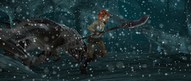} &
    \fcolorbox{gray!50}{white}{\includegraphics[width=19.1mm]{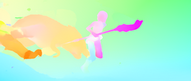}}
    \\[-3.5pt]
    \darkleftbox{19.1mm}{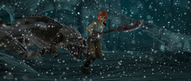}{\scriptsize Adversarial} &
    \includegraphics[width=19.1mm]{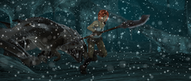} &
    \fcolorbox{gray!50}{white}{\includegraphics[width=19.1mm]{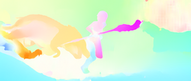}} &
    \darkleftbox{19.1mm}{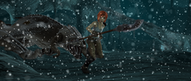}{\scriptsize Adversarial} &
    \includegraphics[width=19.1mm]{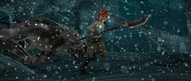} &
    \fcolorbox{gray!50}{white}{\includegraphics[width=19.1mm]{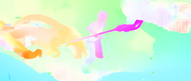}} &
    \darkleftbox{19.1mm}{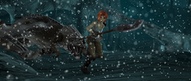}{\scriptsize Adversarial} &
    \includegraphics[width=19.1mm]{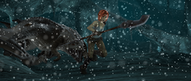} &
    \fcolorbox{gray!50}{white}{\includegraphics[width=19.1mm]{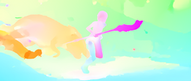}}
    \\
    \darkleftbox{19.1mm}{attacks_half/ambush_2/frame_0003.png}{\scriptsize Original} &
    \includegraphics[width=19.1mm]{attacks_half/ambush_2/frame_0004.png} &
    \fcolorbox{gray!50}{white}{\includegraphics[width=19.1mm]{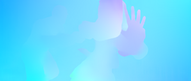}} &
    \darkleftbox{19.1mm}{attacks_half/ambush_2/frame_0003.png}{\scriptsize Original} &
    \includegraphics[width=19.1mm]{attacks_half/ambush_2/frame_0004.png} &
    \fcolorbox{gray!50}{white}{\includegraphics[width=19.1mm]{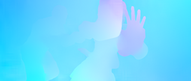}} &
    \darkleftbox{19.1mm}{attacks_half/ambush_2/frame_0003.png}{\scriptsize Original} &
    \includegraphics[width=19.1mm]{attacks_half/ambush_2/frame_0004.png} &
    \fcolorbox{gray!50}{white}{\includegraphics[width=19.1mm]{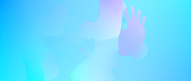}}
    \\[-3.5pt]
    \darkleftbox{19.1mm}{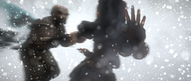}{\scriptsize Random} &
    \includegraphics[width=19.1mm]{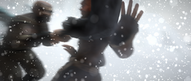} &
    \fcolorbox{gray!50}{white}{\includegraphics[width=19.1mm]{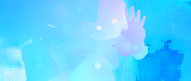}} &
    \darkleftbox{19.1mm}{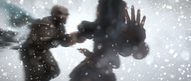}{\scriptsize Random} &
    \includegraphics[width=19.1mm]{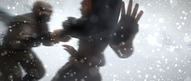} &
    \fcolorbox{gray!50}{white}{\includegraphics[width=19.1mm]{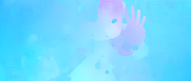}} &
    \darkleftbox{19.1mm}{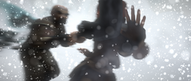}{\scriptsize Random} &
    \includegraphics[width=19.1mm]{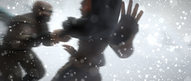} &
    \fcolorbox{gray!50}{white}{\includegraphics[width=19.1mm]{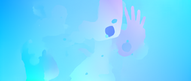}}
    \\[-3.5pt]
    \darkleftbox{19.1mm}{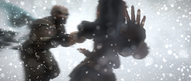}{\scriptsize Adversarial} &
    \includegraphics[width=19.1mm]{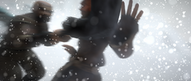} &
    \fcolorbox{gray!50}{white}{\includegraphics[width=19.1mm]{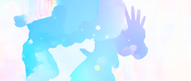}} &
    \darkleftbox{19.1mm}{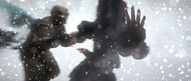}{\scriptsize Adversarial} &
    \includegraphics[width=19.1mm]{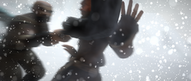} &
    \fcolorbox{gray!50}{white}{\includegraphics[width=19.1mm]{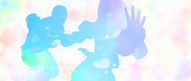}} &
    \darkleftbox{19.1mm}{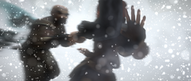}{\scriptsize Adversarial} &
    \includegraphics[width=19.1mm]{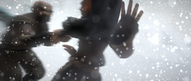} &
    \fcolorbox{gray!50}{white}{\includegraphics[width=19.1mm]{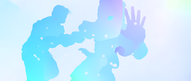}}
    \\
\end{tabular}
\vspace{1mm}
\caption{\emph{Snow}. Qualitative results for 3000 \emph{snowflakes} on images from the Sintel final dataset with \emph{random} initialization and \emph{adversarial} optimization with optical flow predictions for FlowNet2~\cite{Ilg2017Flownet2Evolution}, FlowNetCRobust~\cite{Schrodi2022TowardsUnderstandingAdversarial}, SpyNet~\cite{Ranjan2017OpticalFlowEstimation}, RAFT~\cite{Teed2020RaftRecurrentAll}, GMA~\cite{Jiang2021LearningEstimateHidden} and FlowFormer~\cite{Huang2022FlowformerTransformerArchitecture}, as extension to Main Fig.~\ref{fig:strongattack}. See also Figs.~\ref{fig:strongattackrain}, \ref{fig:strongattacksparks} and~\ref{fig:strongattackfog}.}
\label{fig:strongattacksnow}
\end{figure*}

\begin{figure*}
\centering
\setlength{\fboxrule}{0.1pt}%
\setlength{\fboxsep}{0pt}%
\begin{tabular}{@{}M{19.1mm}@{}M{19.1mm}@{}M{19.1mm}@{\ }M{19.1mm}@{}M{19.1mm}@{}M{19.1mm}@{\ }M{19.1mm}@{}M{19.1mm}@{}M{19.1mm}@{}}
    \multicolumn{3}{c}{FlowNet2} & \multicolumn{3}{c}{FlowNetCRobust} & \multicolumn{3}{c}{SpyNet}
    \\
    \darkleftbox{19.1mm}{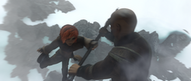}{\scriptsize Original} &
    \includegraphics[width=19.1mm]{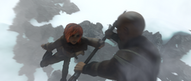} &
    \fcolorbox{gray!50}{white}{\includegraphics[width=19.1mm]{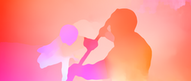}} &
    \darkleftbox{19.1mm}{attacks_half/ambush_6/frame_0001.png}{\scriptsize Original} &
    \includegraphics[width=19.1mm]{attacks_half/ambush_6/frame_0002.png} &
    \fcolorbox{gray!50}{white}{\includegraphics[width=19.1mm]{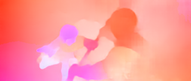}} &
    \darkleftbox{19.1mm}{attacks_half/ambush_6/frame_0001.png}{\scriptsize Original} &
    \includegraphics[width=19.1mm]{attacks_half/ambush_6/frame_0002.png} &
    \fcolorbox{gray!50}{white}{\includegraphics[width=19.1mm]{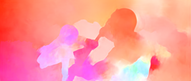}}
    \\[-3.5pt]
    \darkleftbox{19.1mm}{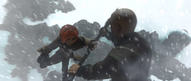}{\scriptsize Random} &
    \includegraphics[width=19.1mm]{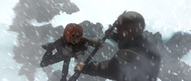} &
    \fcolorbox{gray!50}{white}{\includegraphics[width=19.1mm]{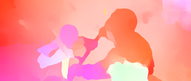}} &
    \darkleftbox{19.1mm}{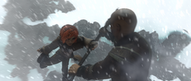}{\scriptsize Random} &
    \includegraphics[width=19.1mm]{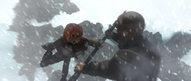} &
    \fcolorbox{gray!50}{white}{\includegraphics[width=19.1mm]{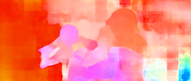}} &
    \darkleftbox{19.1mm}{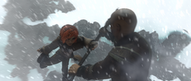}{\scriptsize Random} &
    \includegraphics[width=19.1mm]{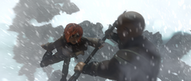} &
    \fcolorbox{gray!50}{white}{\includegraphics[width=19.1mm]{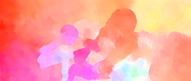}}
    \\[-3.5pt]
    \darkleftbox{19.1mm}{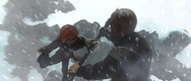}{\scriptsize Adversarial} &
    \includegraphics[width=19.1mm]{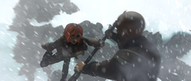} &
    \fcolorbox{gray!50}{white}{\includegraphics[width=19.1mm]{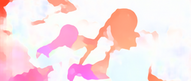}} &
    \darkleftbox{19.1mm}{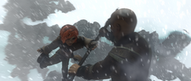}{\scriptsize Adversarial} &
    \includegraphics[width=19.1mm]{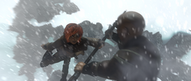} &
    \fcolorbox{gray!50}{white}{\includegraphics[width=19.1mm]{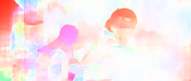}} &
    \darkleftbox{19.1mm}{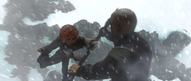}{\scriptsize Adversarial} &
    \includegraphics[width=19.1mm]{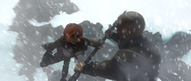} &
    \fcolorbox{gray!50}{white}{\includegraphics[width=19.1mm]{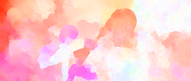}}
    \\
    \darkleftbox{19.1mm}{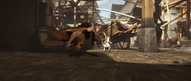}{\scriptsize Original} &
    \includegraphics[width=19.1mm]{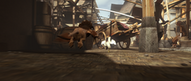} &
    \fcolorbox{gray!50}{white}{\includegraphics[width=19.1mm]{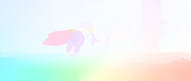}} &
    \darkleftbox{19.1mm}{attacks_half/market_6/frame_0003.png}{\scriptsize Original} &
    \includegraphics[width=19.1mm]{attacks_half/market_6/frame_0004.png} &
    \fcolorbox{gray!50}{white}{\includegraphics[width=19.1mm]{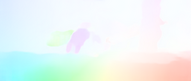}} &
    \darkleftbox{19.1mm}{attacks_half/market_6/frame_0003.png}{\scriptsize Original} &
    \includegraphics[width=19.1mm]{attacks_half/market_6/frame_0004.png} &
    \fcolorbox{gray!50}{white}{\includegraphics[width=19.1mm]{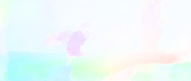}}
    \\[-3.5pt]
    \darkleftbox{19.1mm}{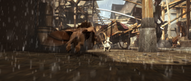}{\scriptsize Random} &
    \includegraphics[width=19.1mm]{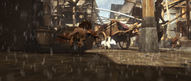} &
    \fcolorbox{gray!50}{white}{\includegraphics[width=19.1mm]{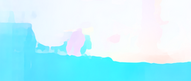}} &
    \darkleftbox{19.1mm}{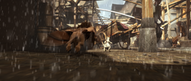}{\scriptsize Random} &
    \includegraphics[width=19.1mm]{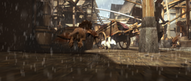} &
    \fcolorbox{gray!50}{white}{\includegraphics[width=19.1mm]{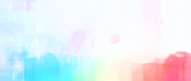}} &
    \darkleftbox{19.1mm}{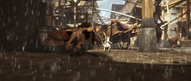}{\scriptsize Random} &
    \includegraphics[width=19.1mm]{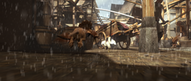} &
    \fcolorbox{gray!50}{white}{\includegraphics[width=19.1mm]{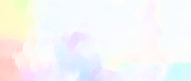}}
    \\[-3.5pt]
    \darkleftbox{19.1mm}{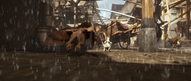}{\scriptsize Adversarial} &
    \includegraphics[width=19.1mm]{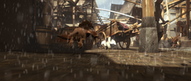} &
    \fcolorbox{gray!50}{white}{\includegraphics[width=19.1mm]{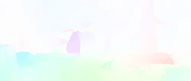}} &
    \darkleftbox{19.1mm}{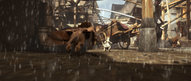}{\scriptsize Adversarial} &
    \includegraphics[width=19.1mm]{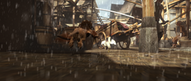} &
    \fcolorbox{gray!50}{white}{\includegraphics[width=19.1mm]{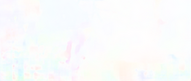}} &
    \darkleftbox{19.1mm}{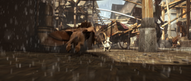}{\scriptsize Adversarial} &
    \includegraphics[width=19.1mm]{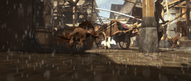} &
    \fcolorbox{gray!50}{white}{\includegraphics[width=19.1mm]{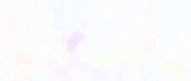}}
    \\
    \darkleftbox{19.1mm}{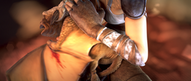}{\scriptsize Original} &
    \includegraphics[width=19.1mm]{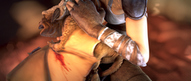} &
    \fcolorbox{gray!50}{white}{\includegraphics[width=19.1mm]{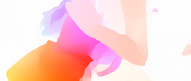}} &
    \darkleftbox{19.1mm}{attacks_half/bandage_1/frame_0003.png}{\scriptsize Original} &
    \includegraphics[width=19.1mm]{attacks_half/bandage_1/frame_0004.png} &
    \fcolorbox{gray!50}{white}{\includegraphics[width=19.1mm]{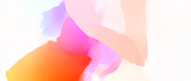}} &
    \darkleftbox{19.1mm}{attacks_half/bandage_1/frame_0003.png}{\scriptsize Original} &
    \includegraphics[width=19.1mm]{attacks_half/bandage_1/frame_0004.png} &
    \fcolorbox{gray!50}{white}{\includegraphics[width=19.1mm]{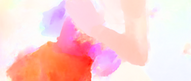}}
    \\[-3.5pt]
    \darkleftbox{19.1mm}{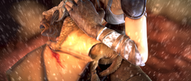}{\scriptsize Random} &
    \includegraphics[width=19.1mm]{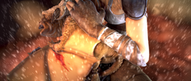} &
    \fcolorbox{gray!50}{white}{\includegraphics[width=19.1mm]{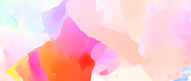}} &
    \darkleftbox{19.1mm}{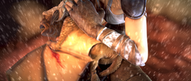}{\scriptsize Random} &
    \includegraphics[width=19.1mm]{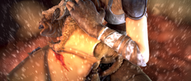} &
    \fcolorbox{gray!50}{white}{\includegraphics[width=19.1mm]{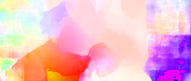}} &
    \darkleftbox{19.1mm}{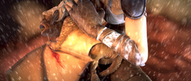}{\scriptsize Random} &
    \includegraphics[width=19.1mm]{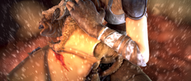} &
    \fcolorbox{gray!50}{white}{\includegraphics[width=19.1mm]{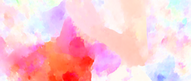}}
    \\[-3.5pt]
    \darkleftbox{19.1mm}{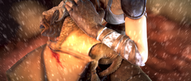}{\scriptsize Adversarial} &
    \includegraphics[width=19.1mm]{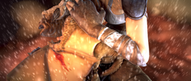} &
    \fcolorbox{gray!50}{white}{\includegraphics[width=19.1mm]{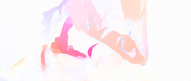}} &
    \darkleftbox{19.1mm}{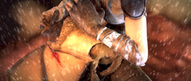}{\scriptsize Adversarial} &
    \includegraphics[width=19.1mm]{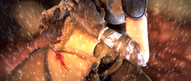} &
    \fcolorbox{gray!50}{white}{\includegraphics[width=19.1mm]{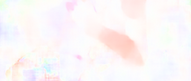}} &
    \darkleftbox{19.1mm}{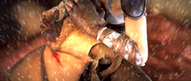}{\scriptsize Adversarial} &
    \includegraphics[width=19.1mm]{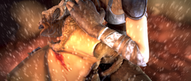} &
    \fcolorbox{gray!50}{white}{\includegraphics[width=19.1mm]{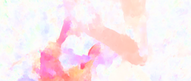}}
    \\
    \multicolumn{3}{c}{RAFT} & \multicolumn{3}{c}{GMA} & \multicolumn{3}{c}{FlowFormer}
    \\
    \darkleftbox{19.1mm}{attacks_half/ambush_6/frame_0001.png}{\scriptsize Original} &
    \includegraphics[width=19.1mm]{attacks_half/ambush_6/frame_0002.png} &
    \fcolorbox{gray!50}{white}{\includegraphics[width=19.1mm]{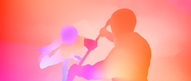}} &
    \darkleftbox{19.1mm}{attacks_half/ambush_6/frame_0001.png}{\scriptsize Original} &
    \includegraphics[width=19.1mm]{attacks_half/ambush_6/frame_0002.png} &
    \fcolorbox{gray!50}{white}{\includegraphics[width=19.1mm]{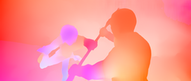}} &
    \darkleftbox{19.1mm}{attacks_half/ambush_6/frame_0001.png}{\scriptsize Original} &
    \includegraphics[width=19.1mm]{attacks_half/ambush_6/frame_0002.png} &
    \fcolorbox{gray!50}{white}{\includegraphics[width=19.1mm]{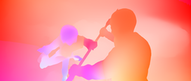}}
    \\[-3.5pt]
    \darkleftbox{19.1mm}{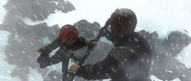}{\scriptsize Random} &
    \includegraphics[width=19.1mm]{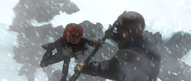} &
    \fcolorbox{gray!50}{white}{\includegraphics[width=19.1mm]{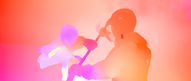}} &
    \darkleftbox{19.1mm}{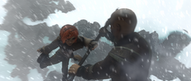}{\scriptsize Random} &
    \includegraphics[width=19.1mm]{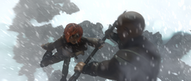} &
    \fcolorbox{gray!50}{white}{\includegraphics[width=19.1mm]{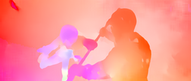}} &
    \darkleftbox{19.1mm}{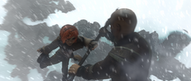}{\scriptsize Random} &
    \includegraphics[width=19.1mm]{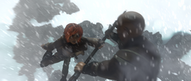} &
    \fcolorbox{gray!50}{white}{\includegraphics[width=19.1mm]{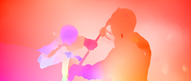}}
    \\[-3.5pt]
    \darkleftbox{19.1mm}{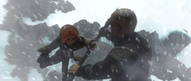}{\scriptsize Adversarial} &
    \includegraphics[width=19.1mm]{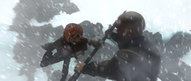} &
    \fcolorbox{gray!50}{white}{\includegraphics[width=19.1mm]{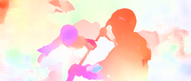}} &
    \darkleftbox{19.1mm}{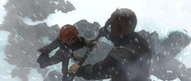}{\scriptsize Adversarial} &
    \includegraphics[width=19.1mm]{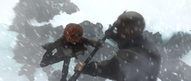} &
    \fcolorbox{gray!50}{white}{\includegraphics[width=19.1mm]{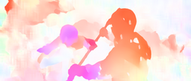}} &
    \darkleftbox{19.1mm}{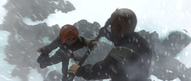}{\scriptsize Adversarial} &
    \includegraphics[width=19.1mm]{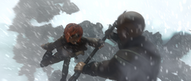} &
    \fcolorbox{gray!50}{white}{\includegraphics[width=19.1mm]{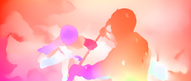}}
    \\
    \darkleftbox{19.1mm}{attacks_half/market_6/frame_0003.png}{\scriptsize Original} &
    \includegraphics[width=19.1mm]{attacks_half/market_6/frame_0004.png} &
    \fcolorbox{gray!50}{white}{\includegraphics[width=19.1mm]{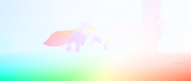}} &
    \darkleftbox{19.1mm}{attacks_half/market_6/frame_0003.png}{\scriptsize Original} &
    \includegraphics[width=19.1mm]{attacks_half/market_6/frame_0004.png} &
    \fcolorbox{gray!50}{white}{\includegraphics[width=19.1mm]{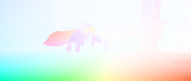}} &
    \darkleftbox{19.1mm}{attacks_half/market_6/frame_0003.png}{\scriptsize Original} &
    \includegraphics[width=19.1mm]{attacks_half/market_6/frame_0004.png} &
    \fcolorbox{gray!50}{white}{\includegraphics[width=19.1mm]{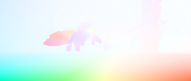}}
    \\[-3.5pt]
    \darkleftbox{19.1mm}{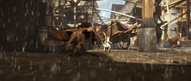}{\scriptsize Random} &
    \includegraphics[width=19.1mm]{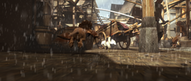} &
    \fcolorbox{gray!50}{white}{\includegraphics[width=19.1mm]{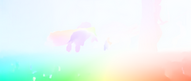}} &
    \darkleftbox{19.1mm}{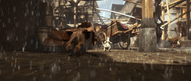}{\scriptsize Random} &
    \includegraphics[width=19.1mm]{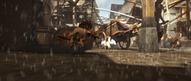} &
    \fcolorbox{gray!50}{white}{\includegraphics[width=19.1mm]{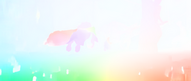}} &
    \darkleftbox{19.1mm}{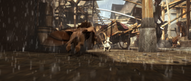}{\scriptsize Random} &
    \includegraphics[width=19.1mm]{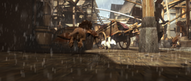} &
    \fcolorbox{gray!50}{white}{\includegraphics[width=19.1mm]{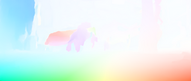}}
    \\[-3.5pt]
    \darkleftbox{19.1mm}{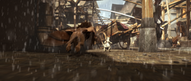}{\scriptsize Adversarial} &
    \includegraphics[width=19.1mm]{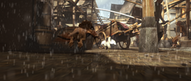} &
    \fcolorbox{gray!50}{white}{\includegraphics[width=19.1mm]{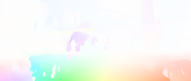}} &
    \darkleftbox{19.1mm}{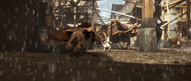}{\scriptsize Adversarial} &
    \includegraphics[width=19.1mm]{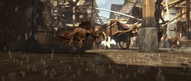} &
    \fcolorbox{gray!50}{white}{\includegraphics[width=19.1mm]{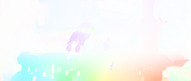}} &
    \darkleftbox{19.1mm}{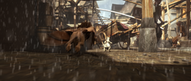}{\scriptsize Adversarial} &
    \includegraphics[width=19.1mm]{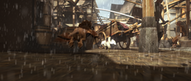} &
    \fcolorbox{gray!50}{white}{\includegraphics[width=19.1mm]{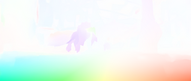}}
    \\
    \darkleftbox{19.1mm}{attacks_half/bandage_1/frame_0003.png}{\scriptsize Original} &
    \includegraphics[width=19.1mm]{attacks_half/bandage_1/frame_0004.png} &
    \fcolorbox{gray!50}{white}{\includegraphics[width=19.1mm]{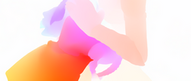}} &
    \darkleftbox{19.1mm}{attacks_half/bandage_1/frame_0003.png}{\scriptsize Original} &
    \includegraphics[width=19.1mm]{attacks_half/bandage_1/frame_0004.png} &
    \fcolorbox{gray!50}{white}{\includegraphics[width=19.1mm]{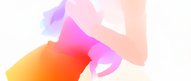}} &
    \darkleftbox{19.1mm}{attacks_half/bandage_1/frame_0003.png}{\scriptsize Original} &
    \includegraphics[width=19.1mm]{attacks_half/bandage_1/frame_0004.png} &
    \fcolorbox{gray!50}{white}{\includegraphics[width=19.1mm]{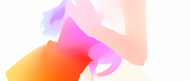}}
    \\[-3.5pt]
    \darkleftbox{19.1mm}{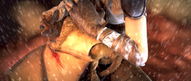}{\scriptsize Random} &
    \includegraphics[width=19.1mm]{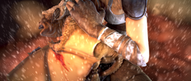} &
    \fcolorbox{gray!50}{white}{\includegraphics[width=19.1mm]{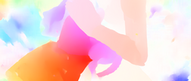}} &
    \darkleftbox{19.1mm}{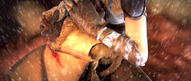}{\scriptsize Random} &
    \includegraphics[width=19.1mm]{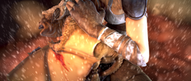} &
    \fcolorbox{gray!50}{white}{\includegraphics[width=19.1mm]{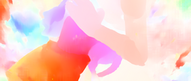}} &
    \darkleftbox{19.1mm}{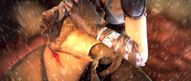}{\scriptsize Random} &
    \includegraphics[width=19.1mm]{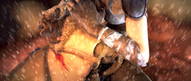} &
    \fcolorbox{gray!50}{white}{\includegraphics[width=19.1mm]{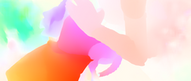}}
    \\[-3.5pt]
    \darkleftbox{19.1mm}{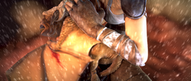}{\scriptsize Adversarial} &
    \includegraphics[width=19.1mm]{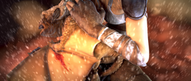} &
    \fcolorbox{gray!50}{white}{\includegraphics[width=19.1mm]{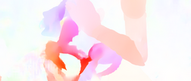}} &
    \darkleftbox{19.1mm}{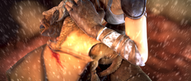}{\scriptsize Adversarial} &
    \includegraphics[width=19.1mm]{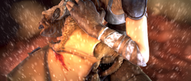} &
    \fcolorbox{gray!50}{white}{\includegraphics[width=19.1mm]{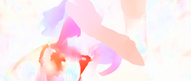}} &
    \darkleftbox{19.1mm}{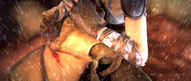}{\scriptsize Adversarial} &
    \includegraphics[width=19.1mm]{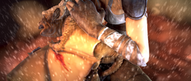} &
    \fcolorbox{gray!50}{white}{\includegraphics[width=19.1mm]{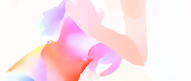}}
    \\
\end{tabular}
\vspace{1mm}
\caption{\emph{Rain}. Qualitative results for 3000 \emph{rain streaks} on images from the Sintel final dataset with \emph{random} initialization and \emph{adversarial} optimization with optical flow predictions for FlowNet2~\cite{Ilg2017Flownet2Evolution}, FlowNetCRobust~\cite{Schrodi2022TowardsUnderstandingAdversarial}, SpyNet~\cite{Ranjan2017OpticalFlowEstimation}, RAFT~\cite{Teed2020RaftRecurrentAll}, GMA~\cite{Jiang2021LearningEstimateHidden} and FlowFormer~\cite{Huang2022FlowformerTransformerArchitecture} as extension to Main Fig.~\ref{fig:strongattack}. See also Figs.~\ref{fig:strongattacksnow}, \ref{fig:strongattacksparks} and~\ref{fig:strongattackfog}.}
\label{fig:strongattackrain}
\end{figure*}

\begin{figure*}
\centering
\setlength{\fboxrule}{0.1pt}%
\setlength{\fboxsep}{0pt}%
\begin{tabular}{@{}M{19.1mm}@{}M{19.1mm}@{}M{19.1mm}@{\ }M{19.1mm}@{}M{19.1mm}@{}M{19.1mm}@{\ }M{19.1mm}@{}M{19.1mm}@{}M{19.1mm}@{}}
    \multicolumn{3}{c}{FlowNet2} & \multicolumn{3}{c}{FlowNetCRobust} & \multicolumn{3}{c}{SpyNet}
    \\
    \darkleftbox{19.1mm}{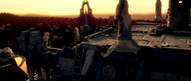}{\scriptsize Original} &
    \includegraphics[width=19.1mm]{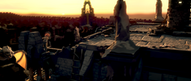} &
    \fcolorbox{gray!50}{white}{\includegraphics[width=19.1mm]{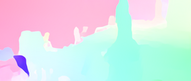}} &
    \darkleftbox{19.1mm}{attacks_half/temple_2/frame_0005.png}{\scriptsize Original} &
    \includegraphics[width=19.1mm]{attacks_half/temple_2/frame_0006.png} &
    \fcolorbox{gray!50}{white}{\includegraphics[width=19.1mm]{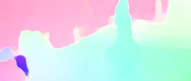}} &
    \darkleftbox{19.1mm}{attacks_half/temple_2/frame_0005.png}{\scriptsize Original} &
    \includegraphics[width=19.1mm]{attacks_half/temple_2/frame_0006.png} &
    \fcolorbox{gray!50}{white}{\includegraphics[width=19.1mm]{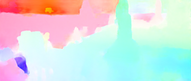}}
    \\[-3.5pt]
    \darkleftbox{19.1mm}{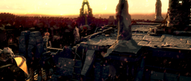}{\scriptsize Random} &
    \includegraphics[width=19.1mm]{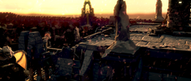} &
    \fcolorbox{gray!50}{white}{\includegraphics[width=19.1mm]{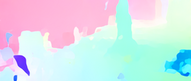}} &
    \darkleftbox{19.1mm}{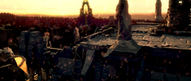}{\scriptsize Random} &
    \includegraphics[width=19.1mm]{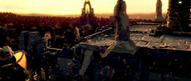} &
    \fcolorbox{gray!50}{white}{\includegraphics[width=19.1mm]{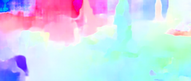}} &
    \darkleftbox{19.1mm}{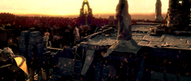}{\scriptsize Random} &
    \includegraphics[width=19.1mm]{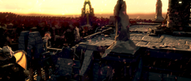} &
    \fcolorbox{gray!50}{white}{\includegraphics[width=19.1mm]{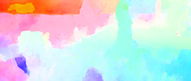}}
    \\[-3.5pt]
    \darkleftbox{19.1mm}{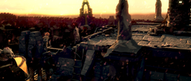}{\scriptsize Adversarial} &
    \includegraphics[width=19.1mm]{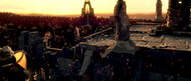} &
    \fcolorbox{gray!50}{white}{\includegraphics[width=19.1mm]{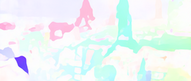}} &
    \darkleftbox{19.1mm}{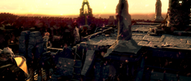}{\scriptsize Adversarial} &
    \includegraphics[width=19.1mm]{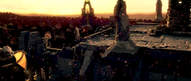} &
    \fcolorbox{gray!50}{white}{\includegraphics[width=19.1mm]{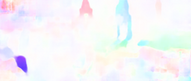}} &
    \darkleftbox{19.1mm}{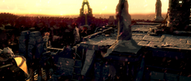}{\scriptsize Adversarial} &
    \includegraphics[width=19.1mm]{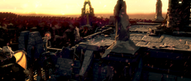} &
    \fcolorbox{gray!50}{white}{\includegraphics[width=19.1mm]{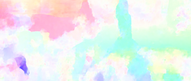}}
    \\
    \darkleftbox{19.1mm}{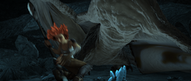}{\scriptsize Original} &
    \includegraphics[width=19.1mm]{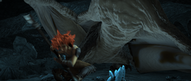} &
    \fcolorbox{gray!50}{white}{\includegraphics[width=19.1mm]{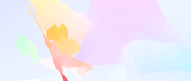}} &
    \darkleftbox{19.1mm}{attacks_half/cave_4/frame_0003.png}{\scriptsize Original} &
    \includegraphics[width=19.1mm]{attacks_half/cave_4/frame_0004.png} &
    \fcolorbox{gray!50}{white}{\includegraphics[width=19.1mm]{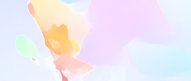}} &
    \darkleftbox{19.1mm}{attacks_half/cave_4/frame_0003.png}{\scriptsize Original} &
    \includegraphics[width=19.1mm]{attacks_half/cave_4/frame_0004.png} &
    \fcolorbox{gray!50}{white}{\includegraphics[width=19.1mm]{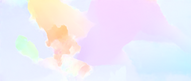}}
    \\[-3.5pt]
    \darkleftbox{19.1mm}{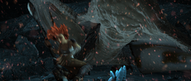}{\scriptsize Random} &
    \includegraphics[width=19.1mm]{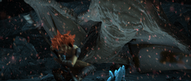} &
    \fcolorbox{gray!50}{white}{\includegraphics[width=19.1mm]{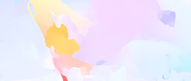}} &
    \darkleftbox{19.1mm}{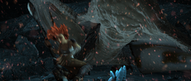}{\scriptsize Random} &
    \includegraphics[width=19.1mm]{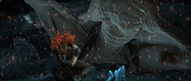} &
    \fcolorbox{gray!50}{white}{\includegraphics[width=19.1mm]{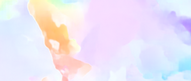}} &
    \darkleftbox{19.1mm}{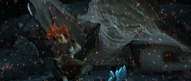}{\scriptsize Random} &
    \includegraphics[width=19.1mm]{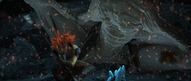} &
    \fcolorbox{gray!50}{white}{\includegraphics[width=19.1mm]{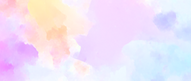}}
    \\[-3.5pt]
    \darkleftbox{19.1mm}{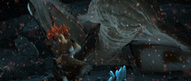}{\scriptsize Adversarial} &
    \includegraphics[width=19.1mm]{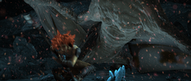} &
    \fcolorbox{gray!50}{white}{\includegraphics[width=19.1mm]{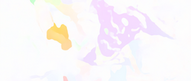}} &
    \darkleftbox{19.1mm}{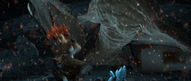}{\scriptsize Adversarial} &
    \includegraphics[width=19.1mm]{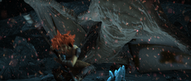} &
    \fcolorbox{gray!50}{white}{\includegraphics[width=19.1mm]{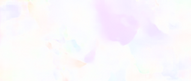}} &
    \darkleftbox{19.1mm}{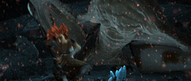}{\scriptsize Adversarial} &
    \includegraphics[width=19.1mm]{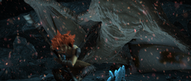} &
    \fcolorbox{gray!50}{white}{\includegraphics[width=19.1mm]{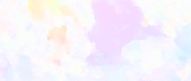}}
    \\
    \darkleftbox{19.1mm}{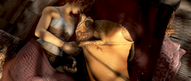}{\scriptsize Original} &
    \includegraphics[width=19.1mm]{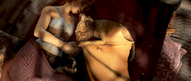} &
    \fcolorbox{gray!50}{white}{\includegraphics[width=19.1mm]{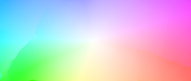}} &
    \darkleftbox{19.1mm}{attacks_half/sleeping_1/frame_0003.png}{\scriptsize Original} &
    \includegraphics[width=19.1mm]{attacks_half/sleeping_1/frame_0004.png} &
    \fcolorbox{gray!50}{white}{\includegraphics[width=19.1mm]{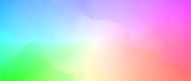}} &
    \darkleftbox{19.1mm}{attacks_half/sleeping_1/frame_0003.png}{\scriptsize Original} &
    \includegraphics[width=19.1mm]{attacks_half/sleeping_1/frame_0004.png} &
    \fcolorbox{gray!50}{white}{\includegraphics[width=19.1mm]{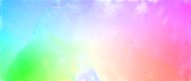}}
    \\[-3.5pt]
    \darkleftbox{19.1mm}{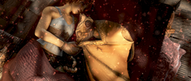}{\scriptsize Random} &
    \includegraphics[width=19.1mm]{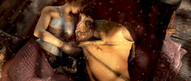} &
    \fcolorbox{gray!50}{white}{\includegraphics[width=19.1mm]{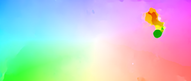}} &
    \darkleftbox{19.1mm}{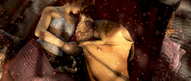}{\scriptsize Random} &
    \includegraphics[width=19.1mm]{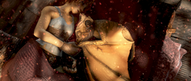} &
    \fcolorbox{gray!50}{white}{\includegraphics[width=19.1mm]{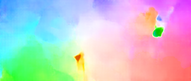}} &
    \darkleftbox{19.1mm}{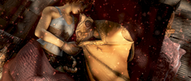}{\scriptsize Random} &
    \includegraphics[width=19.1mm]{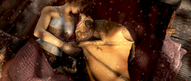} &
    \fcolorbox{gray!50}{white}{\includegraphics[width=19.1mm]{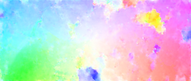}}
    \\[-3.5pt]
    \darkleftbox{19.1mm}{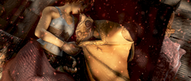}{\scriptsize Adversarial} &
    \includegraphics[width=19.1mm]{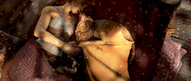} &
    \fcolorbox{gray!50}{white}{\includegraphics[width=19.1mm]{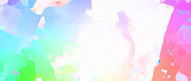}} &
    \darkleftbox{19.1mm}{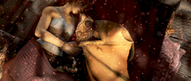}{\scriptsize Adversarial} &
    \includegraphics[width=19.1mm]{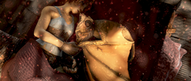} &
    \fcolorbox{gray!50}{white}{\includegraphics[width=19.1mm]{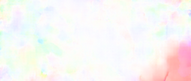}} &
    \darkleftbox{19.1mm}{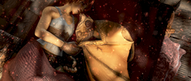}{\scriptsize Adversarial} &
    \includegraphics[width=19.1mm]{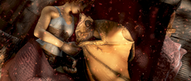} &
    \fcolorbox{gray!50}{white}{\includegraphics[width=19.1mm]{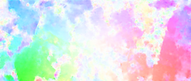}}
    \\
    \multicolumn{3}{c}{RAFT} & \multicolumn{3}{c}{GMA} & \multicolumn{3}{c}{FlowFormer}
    \\
    \darkleftbox{19.1mm}{attacks_half/temple_2/frame_0005.png}{\scriptsize Original} &
    \includegraphics[width=19.1mm]{attacks_half/temple_2/frame_0006.png} &
    \fcolorbox{gray!50}{white}{\includegraphics[width=19.1mm]{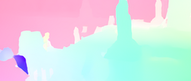}} &
    \darkleftbox{19.1mm}{attacks_half/temple_2/frame_0005.png}{\scriptsize Original} &
    \includegraphics[width=19.1mm]{attacks_half/temple_2/frame_0006.png} &
    \fcolorbox{gray!50}{white}{\includegraphics[width=19.1mm]{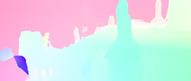}} &
    \darkleftbox{19.1mm}{attacks_half/temple_2/frame_0005.png}{\scriptsize Original} &
    \includegraphics[width=19.1mm]{attacks_half/temple_2/frame_0006.png} &
    \fcolorbox{gray!50}{white}{\includegraphics[width=19.1mm]{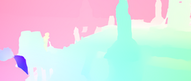}}
    \\[-3.5pt]
    \darkleftbox{19.1mm}{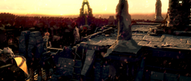}{\scriptsize Random} &
    \includegraphics[width=19.1mm]{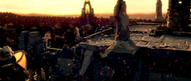} &
    \fcolorbox{gray!50}{white}{\includegraphics[width=19.1mm]{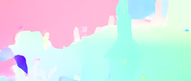}} &
    \darkleftbox{19.1mm}{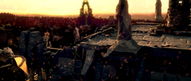}{\scriptsize Random} &
    \includegraphics[width=19.1mm]{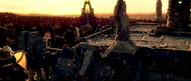} &
    \fcolorbox{gray!50}{white}{\includegraphics[width=19.1mm]{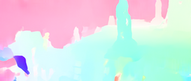}} &
    \darkleftbox{19.1mm}{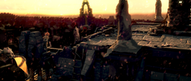}{\scriptsize Random} &
    \includegraphics[width=19.1mm]{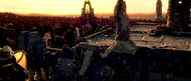} &
    \fcolorbox{gray!50}{white}{\includegraphics[width=19.1mm]{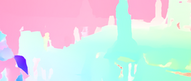}}
    \\[-3.5pt]
    \darkleftbox{19.1mm}{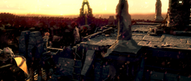}{\scriptsize Adversarial} &
    \includegraphics[width=19.1mm]{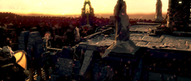} &
    \fcolorbox{gray!50}{white}{\includegraphics[width=19.1mm]{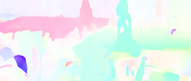}} &
    \darkleftbox{19.1mm}{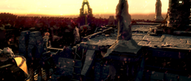}{\scriptsize Adversarial} &
    \includegraphics[width=19.1mm]{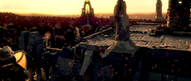} &
    \fcolorbox{gray!50}{white}{\includegraphics[width=19.1mm]{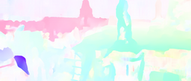}} &
    \darkleftbox{19.1mm}{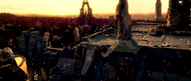}{\scriptsize Adversarial} &
    \includegraphics[width=19.1mm]{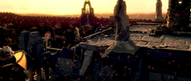} &
    \fcolorbox{gray!50}{white}{\includegraphics[width=19.1mm]{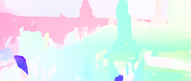}}
    \\
    \darkleftbox{19.1mm}{attacks_half/cave_4/frame_0003.png}{\scriptsize Original} &
    \includegraphics[width=19.1mm]{attacks_half/cave_4/frame_0004.png} &
    \fcolorbox{gray!50}{white}{\includegraphics[width=19.1mm]{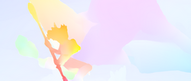}} &
    \darkleftbox{19.1mm}{attacks_half/cave_4/frame_0003.png}{\scriptsize Original} &
    \includegraphics[width=19.1mm]{attacks_half/cave_4/frame_0004.png} &
    \fcolorbox{gray!50}{white}{\includegraphics[width=19.1mm]{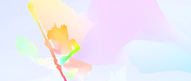}} &
    \darkleftbox{19.1mm}{attacks_half/cave_4/frame_0003.png}{\scriptsize Original} &
    \includegraphics[width=19.1mm]{attacks_half/cave_4/frame_0004.png} &
    \fcolorbox{gray!50}{white}{\includegraphics[width=19.1mm]{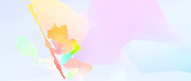}}
    \\[-3.5pt]
    \darkleftbox{19.1mm}{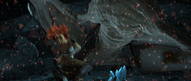}{\scriptsize Random} &
    \includegraphics[width=19.1mm]{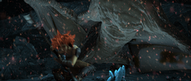} &
    \fcolorbox{gray!50}{white}{\includegraphics[width=19.1mm]{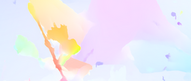}} &
    \darkleftbox{19.1mm}{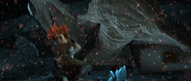}{\scriptsize Random} &
    \includegraphics[width=19.1mm]{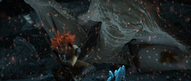} &
    \fcolorbox{gray!50}{white}{\includegraphics[width=19.1mm]{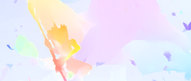}} &
    \darkleftbox{19.1mm}{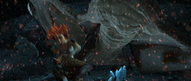}{\scriptsize Random} &
    \includegraphics[width=19.1mm]{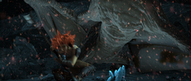} &
    \fcolorbox{gray!50}{white}{\includegraphics[width=19.1mm]{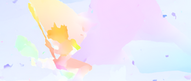}}
    \\[-3.5pt]
    \darkleftbox{19.1mm}{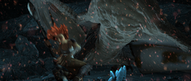}{\scriptsize Adversarial} &
    \includegraphics[width=19.1mm]{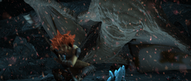} &
    \fcolorbox{gray!50}{white}{\includegraphics[width=19.1mm]{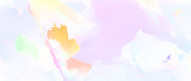}} &
    \darkleftbox{19.1mm}{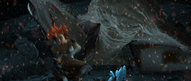}{\scriptsize Adversarial} &
    \includegraphics[width=19.1mm]{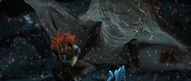} &
    \fcolorbox{gray!50}{white}{\includegraphics[width=19.1mm]{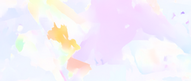}} &
    \darkleftbox{19.1mm}{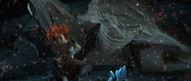}{\scriptsize Adversarial} &
    \includegraphics[width=19.1mm]{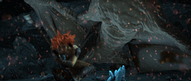} &
    \fcolorbox{gray!50}{white}{\includegraphics[width=19.1mm]{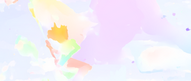}}
    \\
    \darkleftbox{19.1mm}{attacks_half/sleeping_1/frame_0003.png}{\scriptsize Original} &
    \includegraphics[width=19.1mm]{attacks_half/sleeping_1/frame_0004.png} &
    \fcolorbox{gray!50}{white}{\includegraphics[width=19.1mm]{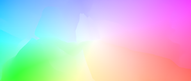}} &
    \darkleftbox{19.1mm}{attacks_half/sleeping_1/frame_0003.png}{\scriptsize Original} &
    \includegraphics[width=19.1mm]{attacks_half/sleeping_1/frame_0004.png} &
    \fcolorbox{gray!50}{white}{\includegraphics[width=19.1mm]{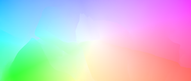}} &
    \darkleftbox{19.1mm}{attacks_half/sleeping_1/frame_0003.png}{\scriptsize Original} &
    \includegraphics[width=19.1mm]{attacks_half/sleeping_1/frame_0004.png} &
    \fcolorbox{gray!50}{white}{\includegraphics[width=19.1mm]{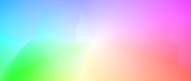}}
    \\[-3.5pt]
    \darkleftbox{19.1mm}{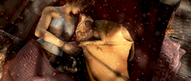}{\scriptsize Random} &
    \includegraphics[width=19.1mm]{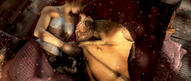} &
    \fcolorbox{gray!50}{white}{\includegraphics[width=19.1mm]{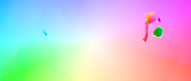}} &
    \darkleftbox{19.1mm}{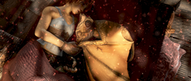}{\scriptsize Random} &
    \includegraphics[width=19.1mm]{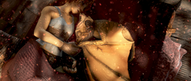} &
    \fcolorbox{gray!50}{white}{\includegraphics[width=19.1mm]{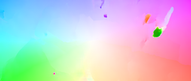}} &
    \darkleftbox{19.1mm}{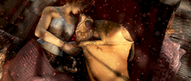}{\scriptsize Random} &
    \includegraphics[width=19.1mm]{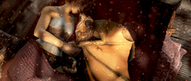} &
    \fcolorbox{gray!50}{white}{\includegraphics[width=19.1mm]{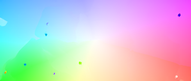}}
    \\[-3.5pt]
    \darkleftbox{19.1mm}{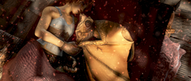}{\scriptsize Adversarial} &
    \includegraphics[width=19.1mm]{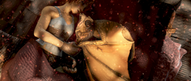} &
    \fcolorbox{gray!50}{white}{\includegraphics[width=19.1mm]{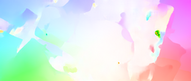}} &
    \darkleftbox{19.1mm}{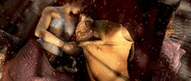}{\scriptsize Adversarial} &
    \includegraphics[width=19.1mm]{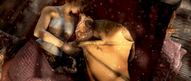} &
    \fcolorbox{gray!50}{white}{\includegraphics[width=19.1mm]{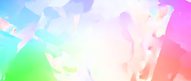}} &
    \darkleftbox{19.1mm}{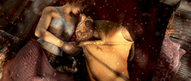}{\scriptsize Adversarial} &
    \includegraphics[width=19.1mm]{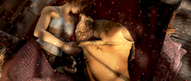} &
    \fcolorbox{gray!50}{white}{\includegraphics[width=19.1mm]{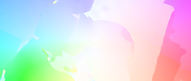}}
    \\
\end{tabular}
\vspace{1mm}
\caption{\emph{Sparks}. Qualitative results for 3000 \emph{fire sparks} on images from the Sintel final dataset with \emph{random} initialization and \emph{adversarial} optimization with optical flow predictions for FlowNet2~\cite{Ilg2017Flownet2Evolution}, FlowNetCRobust~\cite{Schrodi2022TowardsUnderstandingAdversarial}, SpyNet~\cite{Ranjan2017OpticalFlowEstimation}, RAFT~\cite{Teed2020RaftRecurrentAll}, GMA~\cite{Jiang2021LearningEstimateHidden} and FlowFormer~\cite{Huang2022FlowformerTransformerArchitecture} as extension to Main Fig.~\ref{fig:strongattack} and visualization of exemplary results from Main Tab.~4. See also Figs.~\ref{fig:strongattacksnow}, \ref{fig:strongattackrain} and~\ref{fig:strongattackfog}.}
\label{fig:strongattacksparks}
\end{figure*}

\begin{figure*}
\centering
\setlength{\fboxrule}{0.1pt}%
\setlength{\fboxsep}{0pt}%
\begin{tabular}{@{}M{19.1mm}@{}M{19.1mm}@{}M{19.1mm}@{\ }M{19.1mm}@{}M{19.1mm}@{}M{19.1mm}@{\ }M{19.1mm}@{}M{19.1mm}@{}M{19.1mm}@{}}
    \multicolumn{3}{c}{FlowNet2} & \multicolumn{3}{c}{FlowNetCRobust} & \multicolumn{3}{c}{SpyNet}
    \\
    \darkleftbox{19.1mm}{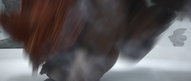}{\scriptsize Original} &
    \includegraphics[width=19.1mm]{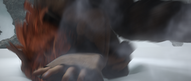} &
    \fcolorbox{gray!50}{white}{\includegraphics[width=19.1mm]{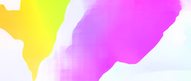}} &
    \darkleftbox{19.1mm}{attacks_half/ambush_4/frame_0005.png}{\scriptsize Original} &
    \includegraphics[width=19.1mm]{attacks_half/ambush_4/frame_0006.png} &
    \fcolorbox{gray!50}{white}{\includegraphics[width=19.1mm]{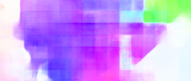}} &
    \darkleftbox{19.1mm}{attacks_half/ambush_4/frame_0005.png}{\scriptsize Original} &
    \includegraphics[width=19.1mm]{attacks_half/ambush_4/frame_0006.png} &
    \fcolorbox{gray!50}{white}{\includegraphics[width=19.1mm]{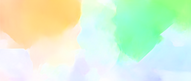}}
    \\[-3.5pt]
    \darkleftbox{19.1mm}{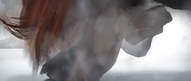}{\scriptsize Random} &
    \includegraphics[width=19.1mm]{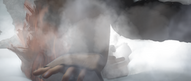} &
    \fcolorbox{gray!50}{white}{\includegraphics[width=19.1mm]{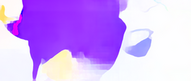}} &
    \darkleftbox{19.1mm}{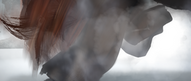}{\scriptsize Random} &
    \includegraphics[width=19.1mm]{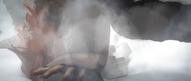} &
    \fcolorbox{gray!50}{white}{\includegraphics[width=19.1mm]{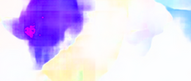}} &
    \darkleftbox{19.1mm}{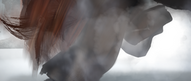}{\scriptsize Random} &
    \includegraphics[width=19.1mm]{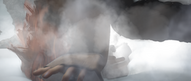} &
    \fcolorbox{gray!50}{white}{\includegraphics[width=19.1mm]{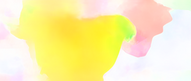}}
    \\[-3.5pt]
    \darkleftbox{19.1mm}{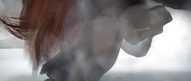}{\scriptsize Adversarial} &
    \includegraphics[width=19.1mm]{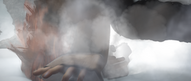} &
    \fcolorbox{gray!50}{white}{\includegraphics[width=19.1mm]{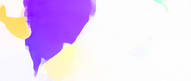}} &
    \darkleftbox{19.1mm}{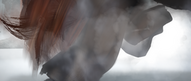}{\scriptsize Adversarial} &
    \includegraphics[width=19.1mm]{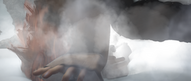} &
    \fcolorbox{gray!50}{white}{\includegraphics[width=19.1mm]{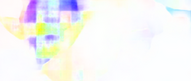}} &
    \darkleftbox{19.1mm}{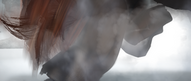}{\scriptsize Adversarial} &
    \includegraphics[width=19.1mm]{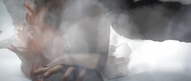} &
    \fcolorbox{gray!50}{white}{\includegraphics[width=19.1mm]{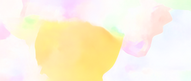}}
    \\
    \darkleftbox{19.1mm}{attacks_half/market_6/frame_0003.png}{\scriptsize Original} &
    \includegraphics[width=19.1mm]{attacks_half/market_6/frame_0004.png} &
    \fcolorbox{gray!50}{white}{\includegraphics[width=19.1mm]{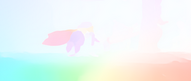}} &
    \darkleftbox{19.1mm}{attacks_half/market_6/frame_0003.png}{\scriptsize Original} &
    \includegraphics[width=19.1mm]{attacks_half/market_6/frame_0004.png} &
    \fcolorbox{gray!50}{white}{\includegraphics[width=19.1mm]{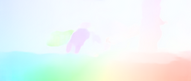}} &
    \darkleftbox{19.1mm}{attacks_half/market_6/frame_0003.png}{\scriptsize Original} &
    \includegraphics[width=19.1mm]{attacks_half/market_6/frame_0004.png} &
    \fcolorbox{gray!50}{white}{\includegraphics[width=19.1mm]{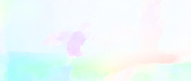}}
    \\[-3.5pt]
    \darkleftbox{19.1mm}{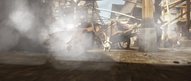}{\scriptsize Random} &
    \includegraphics[width=19.1mm]{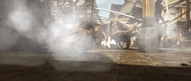} &
    \fcolorbox{gray!50}{white}{\includegraphics[width=19.1mm]{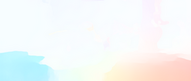}} &
    \darkleftbox{19.1mm}{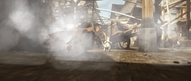}{\scriptsize Random} &
    \includegraphics[width=19.1mm]{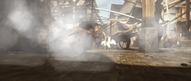} &
    \fcolorbox{gray!50}{white}{\includegraphics[width=19.1mm]{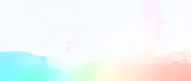}} &
    \darkleftbox{19.1mm}{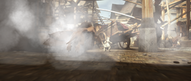}{\scriptsize Random} &
    \includegraphics[width=19.1mm]{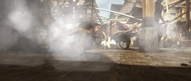} &
    \fcolorbox{gray!50}{white}{\includegraphics[width=19.1mm]{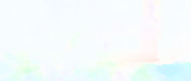}}
    \\[-3.5pt]
    \darkleftbox{19.1mm}{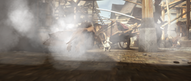}{\scriptsize Adversarial} &
    \includegraphics[width=19.1mm]{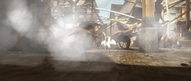} &
    \fcolorbox{gray!50}{white}{\includegraphics[width=19.1mm]{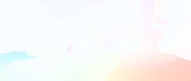}} &
    \darkleftbox{19.1mm}{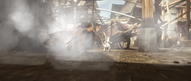}{\scriptsize Adversarial} &
    \includegraphics[width=19.1mm]{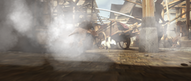} &
    \fcolorbox{gray!50}{white}{\includegraphics[width=19.1mm]{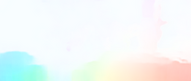}} &
    \darkleftbox{19.1mm}{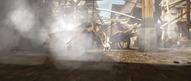}{\scriptsize Adversarial} &
    \includegraphics[width=19.1mm]{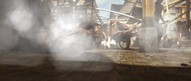} &
    \fcolorbox{gray!50}{white}{\includegraphics[width=19.1mm]{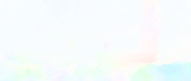}}
    \\
    \darkleftbox{19.1mm}{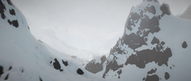}{\scriptsize Original} &
    \includegraphics[width=19.1mm]{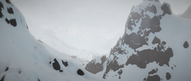} &
    \fcolorbox{gray!50}{white}{\includegraphics[width=19.1mm]{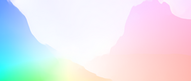}} &
    \darkleftbox{19.1mm}{attacks_half/mountain_1/frame_0003.png}{\scriptsize Original} &
    \includegraphics[width=19.1mm]{attacks_half/mountain_1/frame_0004.png} &
    \fcolorbox{gray!50}{white}{\includegraphics[width=19.1mm]{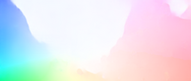}} &
    \darkleftbox{19.1mm}{attacks_half/mountain_1/frame_0003.png}{\scriptsize Original} &
    \includegraphics[width=19.1mm]{attacks_half/mountain_1/frame_0004.png} &
    \fcolorbox{gray!50}{white}{\includegraphics[width=19.1mm]{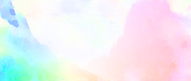}}
    \\[-3.5pt]
    \darkleftbox{19.1mm}{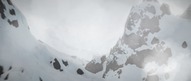}{\scriptsize Random} &
    \includegraphics[width=19.1mm]{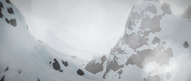} &
    \fcolorbox{gray!50}{white}{\includegraphics[width=19.1mm]{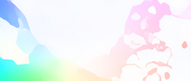}} &
    \darkleftbox{19.1mm}{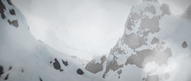}{\scriptsize Random} &
    \includegraphics[width=19.1mm]{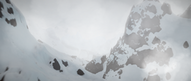} &
    \fcolorbox{gray!50}{white}{\includegraphics[width=19.1mm]{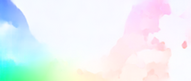}} &
    \darkleftbox{19.1mm}{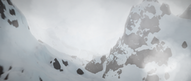}{\scriptsize Random} &
    \includegraphics[width=19.1mm]{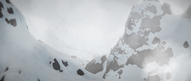} &
    \fcolorbox{gray!50}{white}{\includegraphics[width=19.1mm]{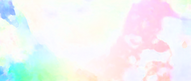}}
    \\[-3.5pt]
    \darkleftbox{19.1mm}{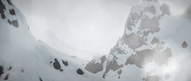}{\scriptsize Adversarial} &
    \includegraphics[width=19.1mm]{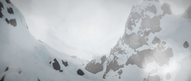} &
    \fcolorbox{gray!50}{white}{\includegraphics[width=19.1mm]{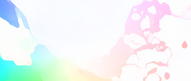}} &
    \darkleftbox{19.1mm}{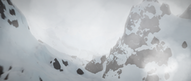}{\scriptsize Adversarial} &
    \includegraphics[width=19.1mm]{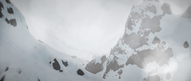} &
    \fcolorbox{gray!50}{white}{\includegraphics[width=19.1mm]{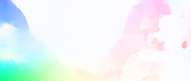}} &
    \darkleftbox{19.1mm}{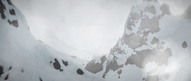}{\scriptsize Adversarial} &
    \includegraphics[width=19.1mm]{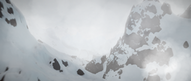} &
    \fcolorbox{gray!50}{white}{\includegraphics[width=19.1mm]{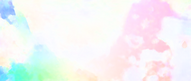}}
    \\
    \multicolumn{3}{c}{RAFT} & \multicolumn{3}{c}{GMA} & \multicolumn{3}{c}{FlowFormer}
    \\
    \darkleftbox{19.1mm}{attacks_half/ambush_4/frame_0005.png}{\scriptsize Original} &
    \includegraphics[width=19.1mm]{attacks_half/ambush_4/frame_0006.png} &
    \fcolorbox{gray!50}{white}{\includegraphics[width=19.1mm]{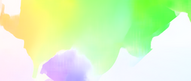}} &
    \darkleftbox{19.1mm}{attacks_half/ambush_4/frame_0005.png}{\scriptsize Original} &
    \includegraphics[width=19.1mm]{attacks_half/ambush_4/frame_0006.png} &
    \fcolorbox{gray!50}{white}{\includegraphics[width=19.1mm]{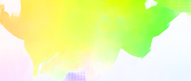}} &
    \darkleftbox{19.1mm}{attacks_half/ambush_4/frame_0005.png}{\scriptsize Original} &
    \includegraphics[width=19.1mm]{attacks_half/ambush_4/frame_0006.png} &
    \fcolorbox{gray!50}{white}{\includegraphics[width=19.1mm]{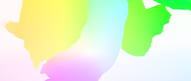}}
    \\[-3.5pt]
    \darkleftbox{19.1mm}{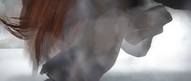}{\scriptsize Random} &
    \includegraphics[width=19.1mm]{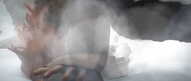} &
    \fcolorbox{gray!50}{white}{\includegraphics[width=19.1mm]{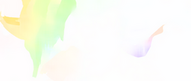}} &
    \darkleftbox{19.1mm}{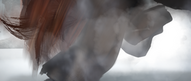}{\scriptsize Random} &
    \includegraphics[width=19.1mm]{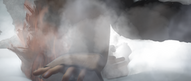} &
    \fcolorbox{gray!50}{white}{\includegraphics[width=19.1mm]{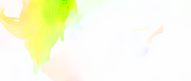}} &
    \darkleftbox{19.1mm}{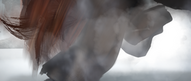}{\scriptsize Random} &
    \includegraphics[width=19.1mm]{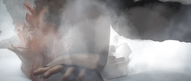} &
    \fcolorbox{gray!50}{white}{\includegraphics[width=19.1mm]{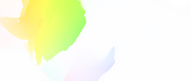}}
    \\[-3.5pt]
    \darkleftbox{19.1mm}{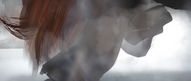}{\scriptsize Adversarial} &
    \includegraphics[width=19.1mm]{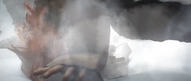} &
    \fcolorbox{gray!50}{white}{\includegraphics[width=19.1mm]{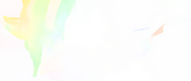}} &
    \darkleftbox{19.1mm}{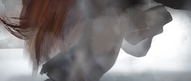}{\scriptsize Adversarial} &
    \includegraphics[width=19.1mm]{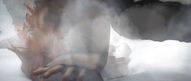} &
    \fcolorbox{gray!50}{white}{\includegraphics[width=19.1mm]{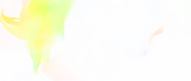}} &
    \darkleftbox{19.1mm}{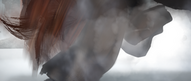}{\scriptsize Adversarial} &
    \includegraphics[width=19.1mm]{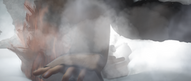} &
    \fcolorbox{gray!50}{white}{\includegraphics[width=19.1mm]{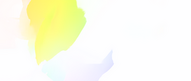}}
    \\
    \darkleftbox{19.1mm}{attacks_half/market_6/frame_0003.png}{\scriptsize Original} &
    \includegraphics[width=19.1mm]{attacks_half/market_6/frame_0004.png} &
    \fcolorbox{gray!50}{white}{\includegraphics[width=19.1mm]{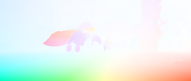}} &
    \darkleftbox{19.1mm}{attacks_half/market_6/frame_0003.png}{\scriptsize Original} &
    \includegraphics[width=19.1mm]{attacks_half/market_6/frame_0004.png} &
    \fcolorbox{gray!50}{white}{\includegraphics[width=19.1mm]{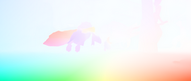}} &
    \darkleftbox{19.1mm}{attacks_half/market_6/frame_0003.png}{\scriptsize Original} &
    \includegraphics[width=19.1mm]{attacks_half/market_6/frame_0004.png} &
    \fcolorbox{gray!50}{white}{\includegraphics[width=19.1mm]{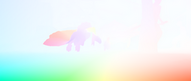}}
    \\[-3.5pt]
    \darkleftbox{19.1mm}{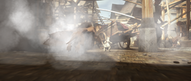}{\scriptsize Random} &
    \includegraphics[width=19.1mm]{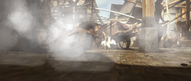} &
    \fcolorbox{gray!50}{white}{\includegraphics[width=19.1mm]{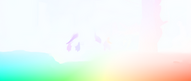}} &
    \darkleftbox{19.1mm}{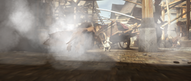}{\scriptsize Random} &
    \includegraphics[width=19.1mm]{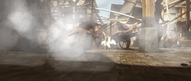} &
    \fcolorbox{gray!50}{white}{\includegraphics[width=19.1mm]{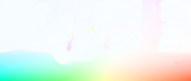}} &
    \darkleftbox{19.1mm}{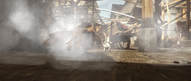}{\scriptsize Random} &
    \includegraphics[width=19.1mm]{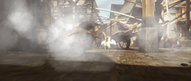} &
    \fcolorbox{gray!50}{white}{\includegraphics[width=19.1mm]{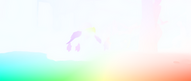}}
    \\[-3.5pt]
    \darkleftbox{19.1mm}{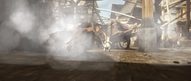}{\scriptsize Adversarial} &
    \includegraphics[width=19.1mm]{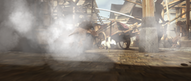} &
    \fcolorbox{gray!50}{white}{\includegraphics[width=19.1mm]{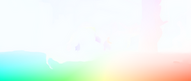}} &
    \darkleftbox{19.1mm}{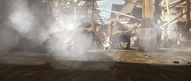}{\scriptsize Adversarial} &
    \includegraphics[width=19.1mm]{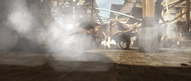} &
    \fcolorbox{gray!50}{white}{\includegraphics[width=19.1mm]{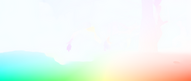}} &
    \darkleftbox{19.1mm}{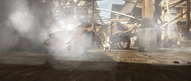}{\scriptsize Adversarial} &
    \includegraphics[width=19.1mm]{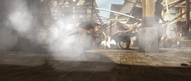} &
    \fcolorbox{gray!50}{white}{\includegraphics[width=19.1mm]{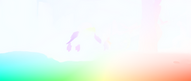}}
    \\
    \darkleftbox{19.1mm}{attacks_half/mountain_1/frame_0003.png}{\scriptsize Original} &
    \includegraphics[width=19.1mm]{attacks_half/mountain_1/frame_0004.png} &
    \fcolorbox{gray!50}{white}{\includegraphics[width=19.1mm]{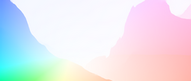}} &
    \darkleftbox{19.1mm}{attacks_half/mountain_1/frame_0003.png}{\scriptsize Original} &
    \includegraphics[width=19.1mm]{attacks_half/mountain_1/frame_0004.png} &
    \fcolorbox{gray!50}{white}{\includegraphics[width=19.1mm]{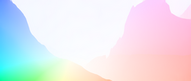}} &
    \darkleftbox{19.1mm}{attacks_half/mountain_1/frame_0003.png}{\scriptsize Original} &
    \includegraphics[width=19.1mm]{attacks_half/mountain_1/frame_0004.png} &
    \fcolorbox{gray!50}{white}{\includegraphics[width=19.1mm]{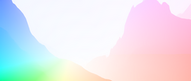}}
    \\[-3.5pt]
    \darkleftbox{19.1mm}{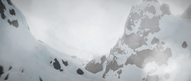}{\scriptsize Random} &
    \includegraphics[width=19.1mm]{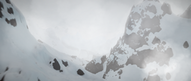} &
    \fcolorbox{gray!50}{white}{\includegraphics[width=19.1mm]{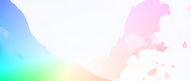}} &
    \darkleftbox{19.1mm}{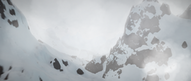}{\scriptsize Random} &
    \includegraphics[width=19.1mm]{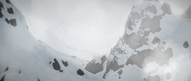} &
    \fcolorbox{gray!50}{white}{\includegraphics[width=19.1mm]{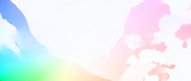}} &
    \darkleftbox{19.1mm}{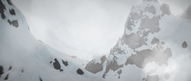}{\scriptsize Random} &
    \includegraphics[width=19.1mm]{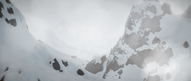} &
    \fcolorbox{gray!50}{white}{\includegraphics[width=19.1mm]{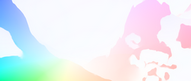}}
    \\[-3.5pt]
    \darkleftbox{19.1mm}{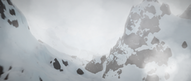}{\scriptsize Adversarial} &
    \includegraphics[width=19.1mm]{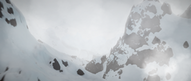} &
    \fcolorbox{gray!50}{white}{\includegraphics[width=19.1mm]{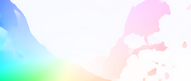}} &
    \darkleftbox{19.1mm}{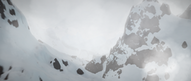}{\scriptsize Adversarial} &
    \includegraphics[width=19.1mm]{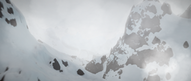} &
    \fcolorbox{gray!50}{white}{\includegraphics[width=19.1mm]{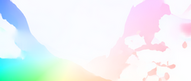}} &
    \darkleftbox{19.1mm}{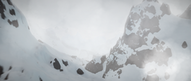}{\scriptsize Adversarial} &
    \includegraphics[width=19.1mm]{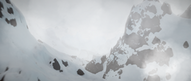} &
    \fcolorbox{gray!50}{white}{\includegraphics[width=19.1mm]{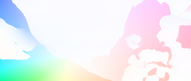}}
    \\
\end{tabular}
\vspace{1mm}
\caption{\emph{Fog.} Qualitative results for 60 \emph{fog clouds} on images from the Sintel final dataset with \emph{random} initialization and \emph{adversarial} optimization with optical flow predictions for FlowNet2~\cite{Ilg2017Flownet2Evolution}, FlowNetCRobust~\cite{Schrodi2022TowardsUnderstandingAdversarial}, SpyNet~\cite{Ranjan2017OpticalFlowEstimation}, RAFT~\cite{Teed2020RaftRecurrentAll}, GMA~\cite{Jiang2021LearningEstimateHidden} and FlowFormer~\cite{Huang2022FlowformerTransformerArchitecture}, as extension to Main Fig.~\ref{fig:strongattack}. See also Figs.~\ref{fig:strongattacksnow}, \ref{fig:strongattackrain} and~\ref{fig:strongattacksparks}.}
\label{fig:strongattackfog}
\end{figure*}

Finally, in Figures~\ref{fig:strongattacksnow}, \ref{fig:strongattackrain}, \ref{fig:strongattacksparks} and~\ref{fig:strongattackfog} we provide additional visualizations for attacks with snow, rain, sparks and fog.
They complement Main Fig.~\ref{fig:strongattack}, and provide visualizations for sample images from the attack runs in Main Tab.~\ref{table:weatherattacks}.


\end{document}